\documentclass{ieeeaccess}
\usepackage{float}
\usepackage{cite}
\usepackage{pgf-pie}
\usepackage{amsmath,amssymb,amsfonts}
\usepackage{algorithmic}
\usepackage{graphicx}
\usepackage{textcomp}
\usepackage{color,soul}
\usepackage{booktabs}
\usepackage{vcell}
\usepackage{tabularx}
\usepackage{graphicx}
\usepackage{adjustbox}
\usepackage{longtable}
\usepackage{array}
\usepackage{pdflscape}
\usepackage{tabu}
\usepackage{supertabular}
\newcommand{\etal}{et al.}
\def\BibTeX{{\rm B\kern-.05em{\sc i\kern-.025em b}\kern-.08em
    T\kern-.1667em\lower.7ex\hbox{E}\kern-.125emX}}
\usepackage{xcolor}
\usepackage{pgfplots}
\pgfplotsset{compat=newest}
\usepackage{pgfplotstable}
\usepackage{tikz}
\usetikzlibrary{svg.path}
\NewSpotColorSpace{PANTONE}
\AddSpotColor{PANTONE} {PANTONE3015C} {PANTONE\SpotSpace 3015\SpotSpace C} {1 0.3 0 0.2}
\SetPageColorSpace{PANTONE}

\definecolor{orcidlogocol}{HTML}{A6CE39}
\tikzset{
    orcidlogo/.pic={
        \fill[orcidlogocol] svg{M256,128c0,70.7-57.3,128-128,128C57.3,256,0,198.7,0,128C0,57.3,57.3,0,128,0C198.7,0,256,57.3,256,128z};
        \fill[white] svg{M86.3,186.2H70.9V79.1h15.4v48.4V186.2z}
        svg{M108.9,79.1h41.6c39.6,0,57,28.3,57,53.6c0,27.5-21.5,53.6-56.8,53.6h-41.8V79.1z M124.3,172.4h24.5c34.9,0,42.9-26.5,42.9-39.7c0-21.5-13.7-39.7-43.7-39.7h-23.7V172.4z}
        svg{M88.7,56.8c0,5.5-4.5,10.1-10.1,10.1c-5.6,0-10.1-4.6-10.1-10.1c0-5.6,4.5-10.1,10.1-10.1C84.2,46.7,88.7,51.3,88.7,56.8z};
    }
}

\newcommand\orcidicon[1]{\href{https://orcid.org/#1}{
                \begin{tikzpicture}[xscale=.03,yscale=.03, transform shape]
                \pic{orcidlogo};
                \end{tikzpicture}
            }}
\usepackage{hyperref}
\renewcommand\hl[1]{#1}

\begin{document}
\setcounter{page}{1} 
\history{Date of publication xxxx 00, 0000, date of current version xxxx 00, 0000.}
\doi{10.1109/ACCESS.2017.DOI}

\title{Bangla Natural Language Processing: A Comprehensive Analysis of Classical, Machine Learning, and Deep Learning Based Methods}
\author{\uppercase{Ovishake Sen$^{\textsuperscript{\orcidicon{0000-0001-7727-9503}}}$\authorrefmark{1}, Mohtasim Fuad\authorrefmark{1}, MD. NAZRUL ISLAM\authorrefmark{1}, Jakaria Rabbi$^{\textsuperscript{\orcidicon{0000-0001-9572-9010}}}$\authorrefmark{1}, Mehedi Masud\authorrefmark{2}, MD. Kamrul Hasan$^{\textsuperscript{\orcidicon{0000-0003-1292-4350}}}$\authorrefmark{3}},
\uppercase{Md. Abdul Awal\authorrefmark{4}, Awal Ahmed Fime\authorrefmark{1}, {Md. Tahmid Hasan Fuad}\authorrefmark{1}, Delowar Sikder\authorrefmark{1}, and MD. Akil Raihan Iftee\authorrefmark{1}}}
\address[1]{Department of Computer Science and Engineering, Khulna University of Engineering \& Technology, Khulna-9203, Bangladesh}
\address[2]{Department of Computer Science, College of Computers and Information Technology, Taif University, P. O. Box 11099, Taif 21944, Saudi Arabia}
\address[3]{Department of Electrical and Electronic Engineering, Khulna University of Engineering \& Technology, Khulna-9203, Bangladesh}
\address[4]{Electronics and Communication Engineering Discipline, Khulna University, Khulna-9208, Bangladesh}

\markboth
{Ovishake \headeretal: BNLP: A Comprehensive Analysis of Classical, ML, and DL Based Methods}
{Ovishake \headeretal: BNLP: A Comprehensive Analysis of Classical, ML, and DL Based Methods}

\corresp{Corresponding author: Jakaria Rabbi (jakaria\_rabbi@cse.kuet.ac.bd) }

\begin{abstract}
The Bangla language is the seventh most spoken language, with 265 million native and non-native speakers worldwide. However, English is the predominant language for online resources and technical knowledge, journals, and documentation. Consequently, many Bangla-speaking people, who have limited command of English, face hurdles to utilize English resources. To bridge the gap between limited support and increasing demand, researchers conducted many experiments and developed valuable tools and techniques to create and process Bangla language materials. Many efforts are also ongoing to make it easy to use the Bangla language in the online and technical domains. There are some review papers to understand the past, previous, and future Bangla Natural Language Processing (BNLP) trends. The studies are mainly concentrated on the specific domains of BNLP, such as sentiment analysis, speech recognition, optical character recognition, and text summarization. There is an apparent scarcity of resources that contain a comprehensive review of the recent BNLP tools and methods. Therefore, in this paper, we present a thorough analysis of 75 BNLP research papers and categorize them into 11 categories, namely Information Extraction, Machine Translation, Named Entity Recognition, Parsing, Parts of Speech Tagging, Question Answering System, Sentiment Analysis, Spam and Fake Detection, Text Summarization, Word Sense Disambiguation, and Speech Processing and Recognition. We study articles published between 1999 to 2021, and 50\% of the papers were published after 2015. \hl{Furthermore,} we discuss Classical, Machine Learning and Deep Learning approaches with different datasets while addressing the limitations and current and future trends of the BNLP.      
\end{abstract}


\titlepgskip=-15pt

\maketitle

\section{Introduction}
\label{sec:introduction}
\PARstart{L}{anguage} is a medium for living beings to communicate with each other. In everyday life, communication is a necessary activity for everyone which enables us to express our feelings, judgment, necessity and language made all this possible. As humans evolved through civilization the format of language has been changed drastically. \hl{Obviously,} the invention of the computer made a far-reaching impact in our modern time as it is helping everything towards making our life easier. The computer communicates and understands everything through 0 and 1 which is called machine language. But machine language is very complex to understand for humans as well as human language is not understandable to the computer. To reduce this language barrier, computer scientists came up with methods to enable computers to understand and process human language. The area of computer science that deals with understanding and processing human language through computers is known as NLP (Natural Language Processing). In recent years, NLP has become a popular topic in computer science as it is solving various real-life problems like auto-correction, text to speech processing, machine translation, sentiment analysis etc. As computers are getting more capable of computational power, the applications of NLP is also increasing. As different languages are used in different regions of the world, it has become a necessity to increase the domain of NLP to enable various language to be used in NLP applications. There are many languages used in the Indian sub-continent and Bangla is one of them. It is widely used in the region of Bangladesh and the West Bengal part of India. The part of NLP that deals with understanding and processing tasks related to the Bangla language are known as Bangla Natural Language Processing (BNLP). \par 

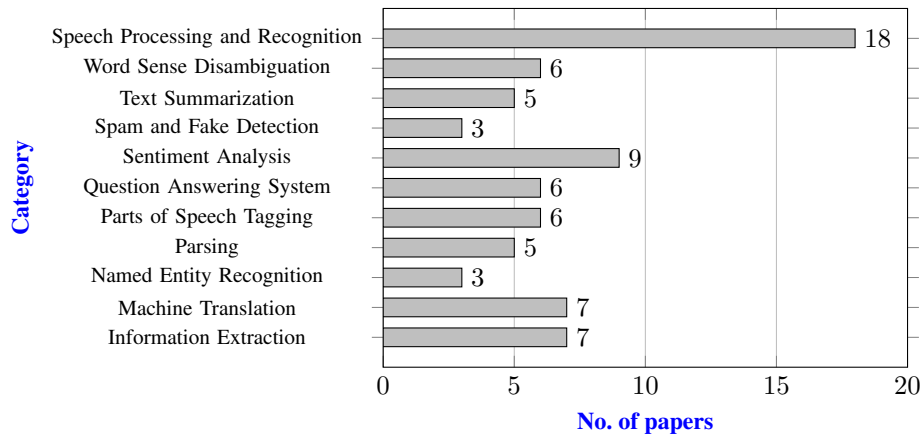
\begin{figure*}
\centering
    \begin{tikzpicture}
        \begin{axis}[
            xbar,
            xlabel={\textbf{\small \textcolor{blue}{No. of papers}}},
            ylabel={\textbf{\small \textcolor{blue}{Category}}},
            height= 6.3cm,
            width= 8.5cm,
            xmin                = 0,
            xmax                = 20,
            nodes near coords,
            xmajorgrids=true,
            y tick label style  = {text width=4.1cm,align=center,font=\footnotesize},
            x tick label style  = {align=center,font=\small,font=\bfseries},
            symbolic y coords={Information Extraction,Machine Translation,Named Entity Recognition,Parsing,Parts of Speech Tagging,Question Answering System,Sentiment Analysis,Spam and Fake Detection,Text Summarization,Word Sense Disambiguation,Speech Processing and Recognition},
            ytick={Information Extraction,Machine Translation,Named Entity Recognition,Parsing,Parts of Speech Tagging,Question Answering System,Sentiment Analysis,Spam and Fake Detection,Text Summarization,Word Sense Disambiguation,Speech Processing and Recognition},
            yticklabels   = {{Information Extraction}, {Machine  Translation}, {Named Entity Recognition},{Parsing},{Parts of Speech Tagging},{Question Answering System},{Sentiment Analysis},{Spam and Fake Detection},{Text Summarization},{Word Sense Disambiguation},{Speech Processing and Recognition}},
            y tick label style={rotate=0},
            every axis plot/.append style={
            xbar,
            bar width=7pt,
            bar shift=0pt
            }
            ]
            \addplot [fill=lightgray] coordinates {(7,Information Extraction)(7,Machine Translation)(3,Named Entity Recognition)(5,Parsing)(6,Parts of Speech Tagging)(6,Question Answering System)(9,Sentiment Analysis)(3,Spam and Fake Detection)(5,Text Summarization)(6,Word Sense Disambiguation)(18,Speech Processing and Recognition)};
        \end{axis}
    \end{tikzpicture}
   \caption  {\centering {Number of Papers Per Category.} }
   \label{fig:figure2}
\end{figure*}

We identified various components and methods of BNLP and Bangla speech processing to present them in this paper. Figure \ref{fig:figure1} shows the components of Natural Language Processing and Figure \ref{fig:figure2} shows the number of papers reviewed as per category. Figure \ref{fig:figure3} shows the methods used in Bangla speech processing and recognition and Figure \ref{fig:figure4} shows the methods used in Bangla natural language processing of text data.

\Figure[h!](topskip=0pt, botskip=0pt, midskip=0pt)[width=8.3cm, height=4.5cm]{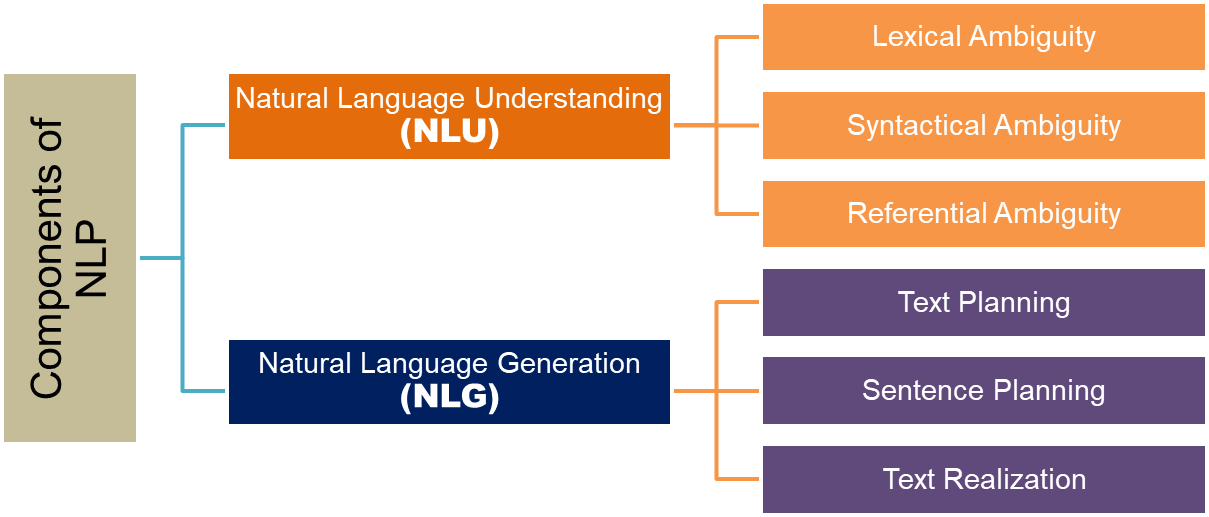}
{Natural Language Processing Components \cite{nlpweb} \label{fig:figure1}. }

\par
\begin{figure*}
\centering
\begin{minipage}{.51\textwidth}
  \centering
  \begin{tikzpicture}
\begin{scope}[scale=.75,xshift=2cm]
\pie[
    text = legend
]
{31.77/CNN,
    15.39/RNN,
    15.39/HMM,
    23.08/LPC,
    15.39/Hybrid methods,
    6.69/Other methods}
\end{scope}
\end{tikzpicture}
\caption  {\centering {Methods used in Bangla Speech Processing and Recognition.}}
\label{fig:figure3}
\end{minipage}%
\begin{minipage}{.5\textwidth}
  \centering
  \begin{tikzpicture}
\begin{scope}[scale=.75,xshift=2cm]
\pie[
    text = legend
]
{7.02/Statistical models,
14.04/ANN,
    8.77/Naive Bayes,
    17.54/Rule-based,
    5.26/N-gram,
    8.77/Clustering algorithms,
    12.28/Hybrid methods,
    26.32/Other methods}
\end{scope}
\end{tikzpicture}

\caption  {\centering {Methods used in Text Data of Bangla Natural Language Processing.}}
\label{fig:figure4}
\end{minipage}
\end{figure*}
\subsection{Preprocessing techniques}
The main idea behind data preprocessing is to transform the raw data into a form so that the data becomes usable and analyzable for the target task and the computer can understand that collected data in the desired format. For every natural language processing related problem, data preprocessing of the collected raw data is one of the most crucial parts. The preprocessed data makes the NLP applications faster and more accurate for the intended task.

\subsection{Classical methods}
The most commonly used classical methods in speech processing and recognition system are Dynamic Time Warping (DTW) \cite{ali2013automatic}, Hidden Markov Model (HMM) \cite{ververidis2006emotional}, Linear Predictive Coding (LPC) \cite{paul2009bangla}, Gaussian Mixture Model (GMM) \cite{ali2013automatic}, Template Matching \cite{ali2013automatic}, Autocorrelation \cite{aadit2016pitch}, Cepstrum \cite{aadit2016pitch}, Speech Application
Program Interface (SAPI) \cite{sultana2012bangla}, Factorial Hidden Markov Model \cite{virtanen2006speech}, Minimum Classification Error \cite{juang1997minimum}, Knowledge Based Approaches \cite{gaikwad2010review}, Template Based Approaches \cite{gaikwad2010review}, and Perceptual Linear Prediction\cite{dave2013feature}. \par
The classical methods used in question answering system are  Cosine similarity \cite{kowsher2019bangla}, Jaccard similarity \cite{kowsher2019bangla}, Anaphora-Cataphora Resolution \cite{khan2018improving}, Vector Space Model \cite{sarker2019bengali}, Semantic Web Technologies \cite{abacha2015means}, Inductive Rule Learning and Reasoning \cite{mitra2016addressing}, Logic Prover \cite{moldovan2005natural} and many more.\par
Spam and fake detection systems mainly use the following classical methods: Traditional Linguistic Features \cite{hossain2020banfakenews}, Text Mining and Probabilistic Language Model \cite{lau2012text}, Review Processing Method \cite{ghai2019spam}, Time Series \cite{heydari2016detection}, and Active Learning \cite{ahsan2016review}. \par
Most of the sentiment analysis systems adopted machine learning (ML) approaches. There are very few classical approaches taken in sentiment analysis. One of the classical methods is the rule-based method\cite{mandal2018preparing}. The classical approaches on machine translation are: Verb based approach\cite{rabbani2014new}, Rule-based approach using parts of speech tagging\cite{chowdhury2013developing}, Fuzzy rules \cite{francisca2011adapting},  Rule based  method\cite{anwar2009syntax} and many more methods. Parts of speech tagging uses the following classical approaches: Brill’s Tagger\cite{hasan2007comparison}, Rule-based approach\cite{hoque2015bangla}, \cite{chowdhury2004parts}, \cite{chakrabarti2011layered}, and Morphological Analysis\cite{chowdhury2004parts}, \cite{chakrabarti2011layered}. \par
The classical approaches on text summarization are Heuristic approach\cite{abujar2017heuristic} and for parsing, mostly used methods are Simple Suffix Stripping algorithm\cite{das2010morphological}, Score based Clustering algorithm\cite{das2010morphological}, and Feature Unification based Morphological Parsing\cite{dasgupta2005morphological}.
In the case of the information extraction for the Bangla languages, many statistical approaches \cite{chandra2013hunting} \cite{kundu2012automatic} have been used to construct different systems. Also, to make a robust system, the use of the classical methods to add more feature sets to their datasets is observed. Like Hough Transform-based Fuzzy Feature \cite{sural1999mlp}, Gradient Feature, and Haar Wavelet Configuration \cite{mandal2011handwritten} are used to add more features to a dataset for extraction of the information. 

\par
In NER (Named Entity Recognition), most of the research works use Dictionary-based, Rule-based, and Statistical-based approaches \cite{chaudhuri2008experiment}. There are also different statistical models such as Conditional Random Fields (CRF)\cite{chowdhury2018towards} is used for NER. Most of the papers in the parsing are based on the Ruled-based method. Different Lexical Analysis, Semantic Analysis \cite{mehedy2003bangla},  \cite{saha2006parsing}
methods have been used for parsing. The researchers also have taken the help of different context-free grammar \cite{mehedy2003bangla} for creating new rules for the parsing. \hl{Additionaly, }the researchers have also used different open-source tools like PC-KIMMO \cite{dasgupta2004morphological} to construct a morphological parser. Analyzing different semantic information \cite{saha2006parsing} can be utilized to create a parser.

\par 

In the text summarization task, various information retrieval techniques for summarization with relational graphs are used. The topic sentiment recognition and aggregation are done through the Topic-sentiment model \cite{lin2009joint}, and Theme Clustering \cite{das2010topic}. Frequency, position value, cue words, and skeleton of the documents are also used \cite{efat2013automated} for text summarization.
The classical approaches that used in word sense disambiguation are Rule-based methods \cite{alam2008text}, \cite{pal2018word}. The rules to mark the semiotic classes and the regular expressions are used frequently \cite{alam2008text}. Also, the use of different Context-free Grammars are seen in the word sense disambiguation\cite{pal2018word}. 

\subsection{Machine learning and deep learning methods}
The most commonly used machine learning and deep learning methods in speech processing and recognition are Convolutional Neural Network (CNN) \cite{sumon2018bangla}, \cite{palaz2015analysis}, Backpropagation Neural Network (BPNN) \cite{ahammad2016connected}, Transfer Learning\cite{sumon2018bangla}, Gated Recurrent Unit (GRU) \cite{sumit2018noise}, Recurrent  Neural Network (RNN) \cite{islam2019speech}, Long Short Term Memory (LSTM) \cite{nahid2017bengali}, Artificial Neural Network (ANN) \cite{neto1995speaker}, Deep Generative Model \cite{nugraha2019deep}, and Deep Neural Network (DNN) \cite{wu2017end}.\par
The machine learning and deep learning methods used in question answering system are Naive Bayes algorithm \cite{kowsher2019bangla}, Support Vector Machine (SVM) \cite{kowsher2019bangla}, Stochastic Gradient Descent (SGD) \cite{islam2016word}, Decision Tree \cite{sarker2019bengali}, N-grams Formation \cite{islam2019design}, and Convolutional Neural Network \cite{sarker2019bengali}.\par
The machine learning and deep learning methods used in spam and fake detection system are Multinomial Naive Bayes classifier \cite{islam2019using}, Support Vector Machine \cite{hussain2020detection}, Neural Network Models \cite{hossain2020banfakenews}, 
Lagrangian Support Vector Machine \cite{ahmed2018detecting}, Logistic regression \cite{ahmed2018detecting}, and Anomalous Rating Deviation \cite{savage2015detection}.
\par
Many machine learning approaches were used for building sentiment analysis systems. The most commonly used machine learning approaches on sentiment analysis are MaxEnt (Maximum Entropy)\cite{chowdhury2014performing}, CNN \cite{tripto2018detecting}, \cite{sarkar2019sentiment}, LSTM \cite{hassan2016sentiment}, \cite{tripto2018detecting}, SVM \cite{chowdhury2014performing}, \cite{tripto2018detecting}, \cite{mahtab2018sentiment}, \cite{sarkar2017sentiment}, \cite{ghosal2015sentiment}, NB (Naive Bayes)\cite{tripto2018detecting}, \cite{islam2016supervised}, \cite{mahtab2018sentiment}, \cite{ghosal2015sentiment}, Decision Tree\cite{mahtab2018sentiment}, \cite{ghosal2015sentiment}, DBN (Deep Belief Network) \cite{sarkar2019sentiment}, SGDC (Stochastic Gradient Descent Classifier)\cite{mandal2018preparing}, Multinomial Naive Bayes\cite{sarkar2017sentiment}, K- Nearest Neighbour (KNN)\cite{ghosal2015sentiment}, and Random Forest classifier (RF)\cite{ghosal2015sentiment}.\par

Machine translation techniques mostly uses classical approaches. The machine learning approach that was used on machine translation is N-gram\cite{islam2010english}
There are many machine learning approaches used in parts of speech tagging, such as Semi-supervised method using Hash Mapping\cite{ismail2014developing}, Baum-Welch algorithm\cite{ali2010unsupervised}, and HMM \cite{ali2010unsupervised},\cite{hasan2007comparison}. Text summarization tasks mostly use K-means clustering\cite{akter2017extractive}, and Unsupervised method using ULMFiT model\cite{rayan2021unsupervised}. \hl{Moreover, }parsing uses machine learning approaches like K-means clustering\cite{das2010morphological}.

\par
Machine learning approaches like segmentation using the Multilayer Perceptron model
\cite{rahman2009real} has been used in retrieving the textual information from the images. LSTM \cite{uddin2019extracting} is usually used for manipulating Bangla sentences. Hough transform method \cite{sural1999mlp}, KNN classifier  \cite{mandal2011handwritten}, CNN-based models with a Histogram of Oriented Gradient (HOG) feature\cite{7853957} are used in retrieving Bangla text information, detecting Bangla scripts. In the NER task, most of the research works have been done around the Ruled-based models. However, some researchers have used modern machine learning algorithms for the NER task. The machine learning algorithm that has been used in the NER is the RNN model \cite{banik2018gru}. In case of word sense disambiguation different machine learning techniques have been used like KNN-based algorithm, Pal  \etal \cite{pal2018word}, \cite{pandit2015memory},  Naive Bayes classifier and ANN \cite{nazah2017word}.

\subsection{Challenges and future research in BNLP}
\subsubsection{\textbf{Challenges}}
\begin{itemize}
\item The researcher faces several preprocessing challenges at the time of working with Bangla text and speech data, including preprocessing of Bangla grammatically wrong words, ambiguous words, erroneous words, falsely interpreted Bangla speeches, preprocessing of very noisy or loud Bangla speeches, and reverberations and overlapping Bangla speeches \cite{indurkhya2010handbook}, \cite{kuligowska2018speech}.\par 

\item The research works of Bangla Natural Language Processing have occasionally faced lackings of rigorous and comprehensive public corpus availability. \hl{That's why} the researchers need to work on a self-built corpus. Moreover, Bangla language does not have modern linguistic tools. \par

\item The complexity of Bangla grammar and structure is one of the most challenging constraints for the researchers for working with BNLP. \hl{Furthermore, }the amount of significant and comprehensive research on BNLP is not enough compared to the other languages, and most are not publicly available. \par

\item There are many non-native Bangla speakers, and they also have variations in their speaking attitude. The researchers have been facing a common difficulty when working with the development of a Bangla speech processing and recognition system that can accurately work in natural, freestyle, noisy, and all possible environments \cite{deng2004challenges}. \par 
\end{itemize} \par
\subsubsection{\textbf{Future Research}}
Despite being the 7th largest spoken language globally, only small amount of significant research works have been done on Bangla language. The research projects and works on BNLP were started in the late 1980s in Bangladesh, and some tangible results have already been produced on BNLP researches \cite{islam2009research}. However, very little research has been done on Bangla language processing compared to English or other more spoken languages. But, there is considerable scope for working with Bangla natural language processing. Sufficient linguistic tools and working prototypes for BNLP can be developed for the Bangla language to create an easier path for the BNLP researchers. 

\par 
Quality research on romanized Bangla can be done as most people communicate through the romanized Bangla \cite{ahmed2021deployment}. Moreover, romanized Bangla language processing research has become one of the most trending challenges on Bangla natural language processing. Quality and significant public corpora on Bangla language can be created as most of the research works on BNLP face the public corpus unavailability. \hl{Furthermore,} more contemporary and suitable deep learning-based models can be developed in the research fields of BNLP as upgraded deep learning models have shown better performance for working with other languages. Therefore, a vast amount of significant research works can be done on Bangla language. \par

\subsection{contributions}
The highlighted points of this paper are noted below:
\begin{itemize}
\item Recent works on BNLP based on classical, machine learning and deep learning have been discussed thoroughly in this article \hl{with insightful analysis. Currently, there is no published research that studies the whole BNLP research area with a thorough investigation.}
\item  We have categorized all the papers into 11 categories (both textual and visual representation), namely Information Extraction, Machine Translation, Named Entity Recognition, Parsing, Parts of Speech Tagging, Question Answering System, Sentiment Analysis, Spam and Fake Detection, Text Summarization, Word Sense Disambiguation, and Speech Processing and Recognition. \hl{To the best of our knowledge, this is the first work that classifies BNLP in those categories.}
\item We have discussed the limitations \hl{and challenges} of the studies with improvement ideas, various datasets and current and future trends of BNLP research.  
\end{itemize}

\subsection{LITERATURE SEARCH PROCEDURE AND CRITERIA}
\hl{As this is a review analysis, it is necessary to describe the literature search criteria as well as the underlying process of our study }\cite{dutt2017systematic}\hl{. The methodological steps and guidelines of Kitchenham et al. }\cite{kitchenham2009systematic}\hl{ for performing a comprehensive literature review were followed in this study. The research purpose of this analysis is to make a comprehensive review study of classical, machine learning, and deep learning-based methods of the existing Bangla Natural Language Processing research works. The primary steps for performing the literature search areas are as follows:
}
\subsubsection{CONSTRUCTING SEARCH TERMS}
\hl{The following information will assist in identifying the search terms that we utilized for our research topic: speech recognition, speech processing, word sense disambiguation, text summarization, spam and fake detection, sentiment analysis, question answering system, parts of speech tagging, parsing, machine translation, named entity recognition, information, extraction, machine learning, deep learning, Bangla natural language processing, text dataset, speech dataset, CNN, RNN, HMM, LPC, ANN, Naive Bayes, clustering, N-gram, etc.}

\begin{table*}
\centering
\caption{Data sources and selections for literature search.}
\label{tab:sourcedata}
\resizebox{\linewidth}{!}{%
\begin{tabu}{>{\hspace{0pt}}p{0.083\linewidth}>{\hspace{0pt}}p{0.414\linewidth}>{\centering\arraybackslash\hspace{0pt}}p{0.242\linewidth}} 
\toprule
\multicolumn{1}{>{\hspace{0pt}}m{0.083\linewidth}}{} & \multicolumn{1}{>{\hspace{0pt}}m{0.414\linewidth}}{} & \multicolumn{1}{>{\centering\arraybackslash\hspace{0pt}}m{0.242\linewidth}}{} \\
\vcell{\textbf{SL. No}} & \multicolumn{1}{>{\centering\hspace{0pt}}p{0.414\linewidth}}{\vcell{\textbf{Data Sources}}} & \vcell{\textbf{No of selected articles}} \\[-\rowheight]
\printcelltop & \multicolumn{1}{>{\centering\hspace{0pt}}m{0.414\linewidth}}{\printcelltop} & \printcelltop \\ \\
\hline\hline
\multicolumn{1}{>{\hspace{0pt}}m{0.083\linewidth}}{} & \multicolumn{1}{>{\hspace{0pt}}m{0.414\linewidth}}{} & \multicolumn{1}{>{\centering\arraybackslash\hspace{0pt}}m{0.242\linewidth}}{} \\ 
\vcell{1.\newline} & \vcell{IEEEXplore} & \vcell{40} \\[-\rowheight]
\printcelltop & \printcelltop & \printcelltop \\
\vcell{2.\newline} & \vcell{Springer} & \vcell{4} \\[-\rowheight]
\printcelltop & \printcelltop & \printcelltop \\
\vcell{3.\newline} & \vcell{Elsevier} & \vcell{2\par{}} \\[-\rowheight]
\printcelltop & \printcelltop & \printcelltop \\
\vcell{4.\newline} & \vcell{Brac University} & \vcell{2\par{}} \\[-\rowheight]
\printcelltop & \printcelltop & \printcelltop \\
\vcell{5.\newline} & \vcell{International Joint Conference on Natural Language Processing\par{}} & \vcell{1} \\[-\rowheight]
\printcelltop & \printcelltop & \printcelltop \\
\vcell{6.} & \vcell{Others} & \vcell{26} \\[-\rowheight]
\printcelltop & \printcelltop & \printcelltop \\
\bottomrule
\end{tabu}
}
\end{table*}
\subsubsection{SEARCH STRATEGY}
\hl{The search terms that have been used in our study are mainly classified by algorithms of natural language processing as well as its used cases. We also looked for alternate synonyms and keywords as well. In our search strings, we included Boolean operators such as AND, OR, etc. To find and filter out relevant publications, multiple data sources are utilized. Table} \ref{tab:sourcedata} \hl{covers the data sources and selected article numbers for our literature search.}
\subsubsection{PUBLICATION SELECTION}
\begin{enumerate}
    \item \textbf{INCLUSION CRITERIA: }\hl{The criteria we used to determine the relevant literature such as journals, articles, conference papers, books, technical reports etc are as follows:}
    \begin{itemize}
        \item Studies that were related to NLP focused on Bangla language.
        \item Studies that contained relevant researches on the subcategories of NLP.
    \end{itemize}
    \item \textbf{EXCLUSION CRITERIA: }\hl{The exclusion of literature that was not relevant to our analysis was determined by the following criteria:}
    \begin{itemize}
        \item Studies that do not provide answers to the relevant NLP research questions.
        \item Studies that were outside the domain of the Bangla language.
    \end{itemize}
    \item \textbf{SELECTING PRIMARY SOURCES: }\hl{The selection process of the papers was initialized by selecting papers that satisfy the search strings or selection criteria which was based on their respective title,  abstract and keywords. After the primary selection was completed the final selection was proceeded by reading the full text of the papers and selecting those that did not fall under the exclusion criteria.  With this process we have identified our final selection of papers. }
    \item \textbf{RANGE OF RESEARCH PAPERS: }\hl{The literature review performed in our study covers researches published from the year 1999 to 2021.}
\end{enumerate}

\subsection{PAPER SUMMARY}
We have discussed various deep learning approaches and the ongoing challenges and advances in Bangla natural language processing. We have shown various methods and pre-processing techniques that are being used in information extraction, machine translation, named entity recognition, parsing, parts of speech tagging, question answering system, sentiment analysis, spam and fake detection, text summarization, word sense disambiguation and speech processing and recognition. Figure \ref{fig:num_paper_yearwise} illustrates the year-wise number of paper we have reviewed in this paper. \hl{Additionaly, }we have discussed the characteristics and complexity of BNLP in the beginning. We have discussed various preprocessing techniques used in various papers. After that, we discussed various approaches, models, results and their limitations in their respective topics such as Information Extraction, Named entity recognition, Sentiment analysis etc. Figure \ref{fig:figure_2} illustrates various categories and techniques used in their respective sections. We have categorized each section into classical, machine learning and hybrid approaches. Figure \ref{fig:bnlp-trend} illustrates the year wise linear trends of using various approaches in BNLP. \hl{Furthermore, }we have briefly discussed classical, machine learning and hybrid approach in the respective sections. Table \ref{tab:CL-BNLP} shows the classical methods and Table \ref{tab:ML-BNLP} shows the machine learning methods used in BNLP. Bangla natural language processing has various challenges to overcome and we have discussed challenges in various sections of BNLP in the challenges section. Datasets are one of the crucial factors for the advancement of BNLP research and therefore, in the last section, we have discussed various text datasets and speech datasets. In conclusion, various advances, challenges and methods used in BNLP have been shown in our paper.

\Figure[t!](topskip=0pt, botskip=0pt, midskip=0pt)[width=8.52cm,height=6.5cm]{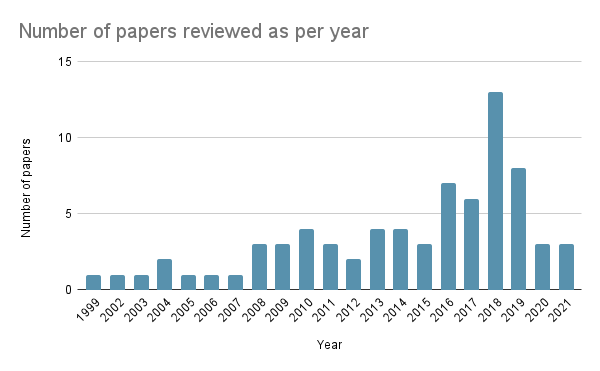}
    {Number of Papers Reviewed as per Year.
    \label{fig:num_paper_yearwise}}

\Figure[t!](topskip=0pt, botskip=0pt, midskip=0pt)[width=8.52cm,height=6.5cm]{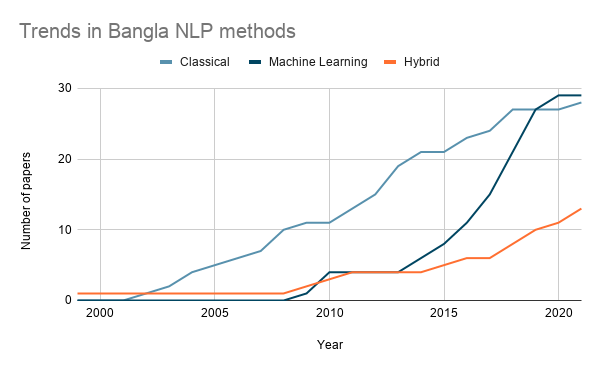}
{Trends in BNLP Methods.
    \label{fig:bnlp-trend}}

\subsection{ORGANIZATION}
We have organized the rest of this paper in the following way.
\begin{itemize}
  \item \ref{sec:BanglaLanguage} {\textbf{Bangla Language:}} The characteristics and complexities of the Bangla language are described briefly. 
  \item \ref{sec:BNLP} {\textbf{Bangla Natural Language Processing:}} The necessity and needs, as well as different categories and methods used in BNLP are described briefly. Articles published on different categories of BNLP till now in both text and speech formats are discussed in this section.
  \item \ref{sec:challenges} {\textbf{Challenges in BNLP:}} The challenges faced in BNLP research areas on both text and speech formats are described in this section.
  \item \ref{sec:datasets} {\textbf{BNLP Datasets:}} The datasets used in BNLP research articles published till now on both text and speech formats are described in this section.
  \item \ref{sec:conclusion} {\textbf{Conclusion:}} Overall summary of our work.
\end{itemize}

\section{Bangla Language}
\label{sec:BanglaLanguage}
\subsection{Characteristics}
Bangla language belongs to the Indo-European language family it is one of the members of the Indo-Aryan group of the Indo-Iranian branch. The phonology of Bangla is similar to the other Indo-Aryan languages. The basic Bangla alphabet consists of 11 vowels, 39 consonants. There are also ten numerical and compound characters. These compound characters are constructed using a combination of consonant and consonant or consonant and vowel. Bangla uses diacritical marks for vowels as well as for some consonants and compound consonants take different shapes. All of these result in a total of almost 300 characters set for Bangla and written from left to right. There is no capitalization or every character is written on upper-lower case letter. Many conjuncts, upstrokes, downstrokes, and other features hanging from a horizontal line define the script. \hl{Furthermore, }a regional variation is seen in spoken Bangla. There are two written forms, namely SADHU and CHALITA. The stress in Bangla is primarily initial and the stress position does not affect the meaning of the words. 
\par
Bangla is an inflectional language which uses various prefixes and suffixes to show the grammatical relations and there is no gender distinctions. There are three persons to represent the pronouns in Bangla. Normally the sentences follow subject-object-verb word patterns. Bangla primarily has two moods known as indicative and imperative. The indicative mood has three different tenses, namely present, past and future. Bangla root verbs are mostly monosyllabic or bisyllabic. Bangla uses left-branching for the order of words which mean adjective places before nouns and adverbs are placed before verbs. Bangla uses a post-verbal negative particle. The position dependency of a headword is comparatively less firm in the sentences. \hl{Moreover, }the Bangla's lexis consists of native Bangla words and borrowings from Sanskrit along with other neighbouring languages. Bangla is a morphologically, rich language.

\subsection{Complexity} 

Bangla is a widely spoken language and around 265 million people speak in Bangla. However, there is a lack of proper datasets for the research and in most cases, the researchers use their own set of corpus for their research. It is also seen that there is less collaborative work among the researchers in this field. Also, the research's corpus often does not get labeled professionally, which has become a primary concern in Bangla NLP works. Lack of standard corpus, labeled data are making things hard for the researchers to study in this area. Also, there is a lack of rigorous morphological analysis of Bangla language. 

\par 

\hl{Furthermore, }there are more Bangla characters than other languages, and many characters are similar. The use of compound characters makes it hard for different OCR \cite{smith2007overview} and other related works. Bangla is also a morphologically rich language, which imposes challenges in research and there is no proper lexicon built for it. Moreover, Bangla is often mixed with English and created the code mixed problem for the research purposes. Code mixing imposes a significant challenge in sentiment analysis or other types of tasks. Additionally, most of the computing tools available for Bangla only work on specific platforms, becoming troublesome for researchers. Also, the lack of adequately labeled datasets often arises problems when the researchers need to annotate the datasets for themselves, which may lead to erroneous results or the impact of the research sometimes becomes incoherent. 

\section{Bangla Natural Language Processing}
\label{sec:BNLP}
\Figure[h!](topskip=0pt, botskip=0pt, midskip=0pt)[width=18cm, height=16.5cm]{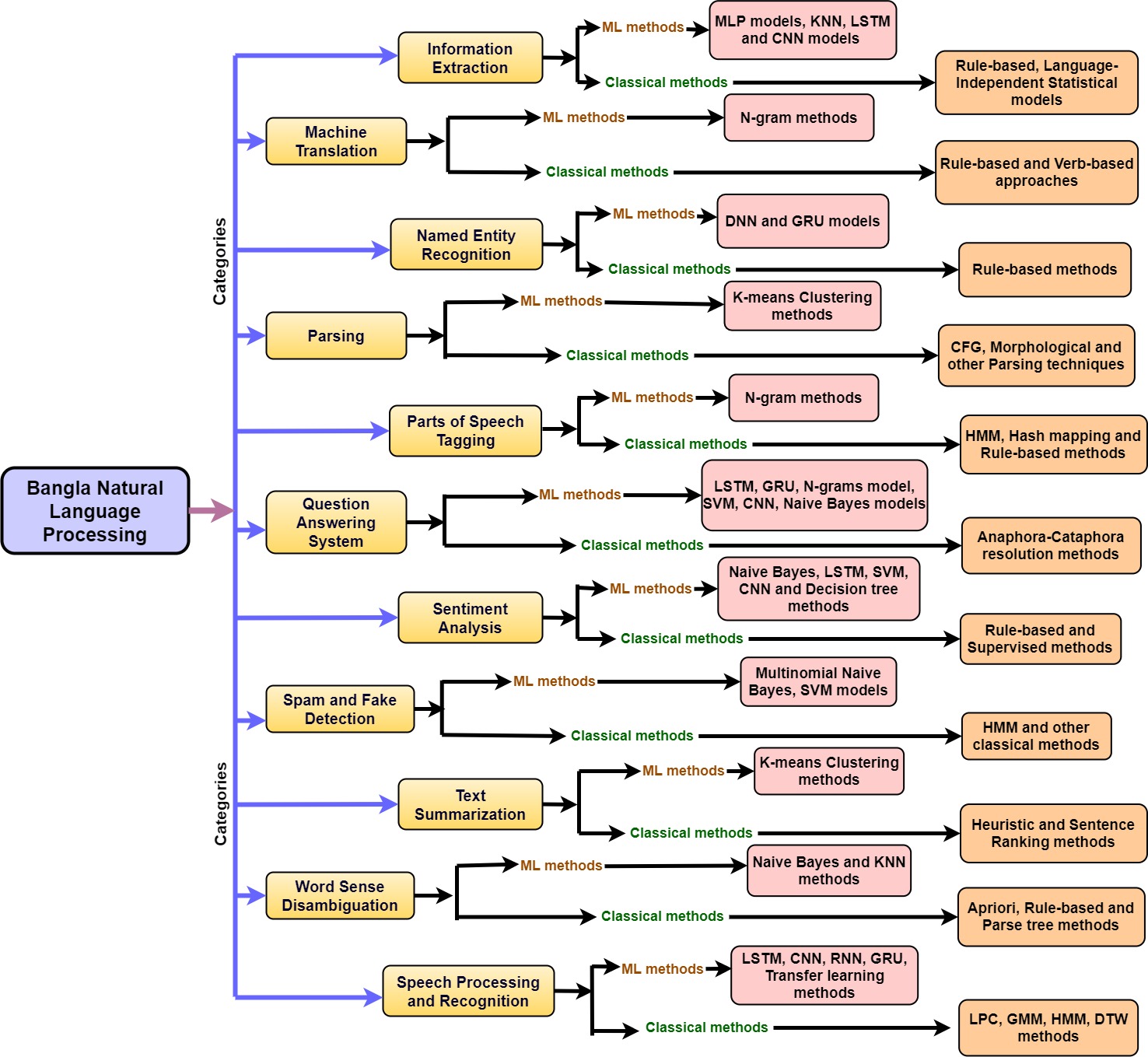}
{\hl{Bangla Natural Language Processing Tasks} \label{fig:figure_2}}

Natural language processing is an emerging field in modern computer science. Language is an important medium of sharing data. It is a way to understand the world that surrounds us. Every living species has its ways to communicate. But humans created languages to communicate and transfer understandable data to one another. Because of region-based culture and environment, different languages have been created throughout centuries. \hl{Additionaly, }some languages evolved, some languages went extinct. But the main object of the language remained the same, communicating with each other. In modern times, the world has become smaller, people can communicate with each other living in one place to a person living in another part of the world. This has introduced a challenge to lessen the language barrier of people’s communication. Also, the increasing use of computer created a necessity of broadening the embedding of human language with the computer.

\par 
We can quote some texts about the aim of a linguistic science from a NLP book \cite{manning1999foundations} written by Christopher D. Manning as follows - "The aim of a linguistic science is to be able to characterize and explain the multitude of linguistic observations circling around us, in conversations, writing, and other media. Part of that has to do with the cognitive side of how humans acquire, produce, and understand language, part of it has to do with understanding the relationship between linguistic utterances and the world, and part of it has to do with understanding the linguistic structures by which language communicates.” Natural language processing enables computers to deal with the linguistic science of human language. Figure \ref{fig:figure_2} shows the categories and techniques used in Bangla natural language processing till now.
\par
Bangla is a culturally enriched language used mostly in the region of South Asia. It is an Indo-Aryan originated language mostly used in the country of Bangladesh as the official and the national language and as the official language in the state of West Bengal Tripura and Assam of India. The language has around 230 million native speakers and around 35 million second-language speakers. \hl{Moreover, }it is 7th among the most spoken language in the world. The part of natural language processing that works on linguistic analysis of Bangla language is known as Bangla natural language processing. The digitalization of modern society has increased the use of computers and other computationally capable devices. Bangla natural language processing is an important part of this transformation to the digital world. As one of the biggest spoken languages in South Asia, the Bangla language is widely used as a script, books, communication, social media posts etc. This wide usage of this language in digital form is creating a massive amount of data in the digital world and by utilizing the resources, it would be beneficial for the field of modern linguistic analysis and data science. Sentiment analysis, named entity recognition, parts of speech tagging, speech processing and recognition, text summarization and other fields of natural language processing is now being able to apply in the Bangla language because of Bangla natural language processing. Bangla is a culturally diverse language with having various accent and forms like mixed and romanized Bangla. Many works are being conducted on Bangla which is enriching the natural language processing part of the Bangla language.

\subsection{Necessity and needs}
Bangla natural language processing systems are essential for making the natural Bangla language understandable and interpretable to computers. BNLP focuses on clipping crucial features from the Bangla language speech and text data using available computing resources \cite{banik2018gru}. The modern world is generating a large amount of unstructured Bangla text and speech data every second. So, BNLP systems are used to structure and process this vast amount of Bangla data and make it compatible with computers. \hl{As a result,} Nowadays BNLP has become highly demandable in many sectors.
\par
Information extraction in BNLP allows machines to decipher and bring out the essential knowledge of words from Bangla documents \cite{karim2019step}.
Machine translation in BNLP is used to translate the Bangla language text data to another language \cite{islam2010english}. Named entity recognition in BNLP aims at identifying and classifying every word of a Bangla document into some predefined named categories \cite{ekbal2007hidden}. The main concern of parsing in BNLP is to identify and classify the part of speech in a sentence of Bangla documents \cite{saha2006parsing}. Parts of speech tagging systems in BNLP are used to tag the parts-of-speech, tense, aspect, case, person, number, gender, etc. of Bangla words in a sentence \cite{chowdhury2004parts}.
\par
The primary purpose of the question answering system in BNLP is to implement a chatbot that helps to create better human-computer interaction system in Bangla language \cite{orin2017implementation}.
The main concern of sentiment analysis in BNLP is to select the opinions, emotions, and sentiments from Bangla texts and speech data \cite{chowdhury2014performing}.
Spam and fake detection systems in BNLP aim to detect fake, fabricated, forged, or suspicious Bangla data produced every day with the help of digital technologies \cite{sharif2020detecting}. Text summarization in BNLP summarizes and extracts necessary information from the Bangla documents \cite{efat2013automated}. The main purpose of word sense disambiguation in BNLP is to extract the exact meaning of a Bangla word with multiple meanings by analyzing the other words used in that specific sentence. \cite{kaysar2018word}.
Speech processing and recognition in BNLP use Bangla speech data as a user interface to interact with computers and make the Bangla speech data understandable to the computers \cite{hossain2013implementation}.
\subsection{Preprocessing} \par
The general and most used preprocessing techniques in text data that are used in Natural language processing are  Tokenization \cite{borocs2018nlp}, \cite{sarker2019bengali}, Removal of Noise and Outliers \cite{perovvsek2016textflows}, Lower Casing \cite{ahsan2016review}, Removal of Stop Words \cite{islam2019design}, Integration of Raw data \cite{chowdhury2003natural}, Text Stemming \cite{ahmed2018detecting}, Dimensionality Reduction\cite{gorrell2006generalized}, Text Lemmatization \cite{kowsher2019bangla}, Parts of Speech Tagging \cite{voutilainen2003part}, Removal of HTML tags \cite{gupta2003dom}, Removal of URLs \cite{kanakaraj2015nlp}, Spelling Correction \cite{etoori2018automatic}, Removal of Emoticons \cite{wang2015sentiment}\cite{hassan2016sentiment}, Removing Punctuations or Special Characters \cite{kowsher2019bangla}\cite{islam2019using}, Removing Frequent words \cite{gonen2019lipstick}, Removing of Rare words \cite{chiu2016train}, Removing Single Characters \cite{kolak2003generative}, Removing Extra Whitespaces \cite{nath2016natural}, Removal of Numerical Values \cite{islam2019using}, Removing Alphabets \cite{hussain2020detection}, Data Compression
 \cite{nadkarni2011natural}, Converting Emojis to Words \cite{wu2019bnu}, Converting Numbers to Words \cite{rychalska2016samsung}, Text Normalization \cite{tursun2017noisy,hossain2020banfakenews},  Text Standardization \cite{kreimeyer2017natural}, Poping Wh-type Words \cite{khan2018improving}, Anaphora \cite{kowsher2019bangla}, Verb Processing \cite{hanks2005pattern}, Synonym Words Processing \cite{jarmasz2012roget}, N-gram
Stemming \cite{urmi2016corpus} and many more. \par
The general and most used preprocessing techniques on speech data in Natural Language processing are Noise Removal     \cite{droppo2002uncertainty}\cite{hasnat2007isolated}, Pre-emphasis \cite{vergin1995pre}\cite{paul2009bangla}, Hamming Window \cite{ali2013automatic}, Segmentation \cite{ahammad2016connected}, Sampling \cite{shannon2017optimizing}\cite{aadit2016pitch}, Phoneme Mapping \cite{nahid2017bengali}, Speech Coding \cite{paul2009bangla}, Stemming without Noise Removal \cite{repp2007segmentation}, Voice Activity Detection \cite{ramirez2007voice}, Framing \cite{lokesh2019speech} and many more.

\subsection{Information Extraction}
\hl{Information extraction systems use natural language text as input and create structured information based on predefined criteria that is relevant to a specific application. The main target of information extraction is to extract salient facts about pre-specified categories of events, entities, or connections to construct more meaningful, rich representations of their semantic content that may be utilized to populate databases that require more organized input. \mbox{\cite{singh2018natural}}}
\subsubsection{Classical Approaches}

In 2012, Kundu \etal \cite{kundu2012automatic} proposed a statistical language-independent approach for identifying the foreign words in the code-mixed language. They mainly implemented the model in Bangla-English code-mixed scenarios. Moreover, for Bangla-English code-mixed language, the model gained a reasonable accuracy. Three cases are seen in Bangla-English code-mixing. Tag-switching, inter-sentential and intrasentential are the everyday use cases for the code-mixed example in Bangla. The authors suggested a statistical model for detecting foreign words in the code-mixed language and here, in this case, to detect English words. They used several steps to detect English words. In their model, the authors first detected the English words in context with roman script. Then they approached the idea that English which appeared in Bangla script, might or might not contain Bangla infection. Also, no Bangla words contain English suffix except some exceptional examples. Furthermore, Bangla words generally appear more than English in code-mixed scenarios. The structure of Bangla grammar remains primary during a code-mixed example. Based on these ideas, the authors suggested their statistical model. They trained their model from a blog article and collected the English words and sentences manually. They got an accuracy of 71.82\%, implementing the model at social media post data.\hl{However, No preprocessing steps were mentioned, and a clear indication of the dataset is absent. This paper can be improved by using the model on sophisticated datasets to identify the model's potential.}
\par
Chandra \etal \cite{chandra2013hunting} suggested a statistical methods combined with a rule-based approach for identifying the English words in Benglish and Hinglish languages. Some basic patterns seen in the code-mixed scenario with English, like English words are written within roman scripts. Some suffixes may be added after the English words. English words may contain a light verb of Bangla, or they are mixed with
the language thoroughly. The authors had Bangla text collected from CIMC \cite{chandra2013hunting} for their work purpose. They implemented a combined rule-based and statistical method for detection of the English text from the Bangla language. They tried to detect the English word based on the rules in three categories, and they used the Graphene Language Model (GLM), and Phoneme Language Model (PLM) based statistical model for their proposed method. The proposed model achieved 95.96\% accuracy in detecting English words from the Bangla-English code mixed language. The model was evaluated with 9,152 sentences, and those were collected from different online newspapers, blogs, social media and journals.
\hl{However, no results for Hinglish code mixed language have been presented in this paper. Combining the method with modern machine learning algorithms can improve this paper.}

\subsubsection{Machine Learning Approaches}
Rahman \etal \cite{rahman2009real} proposed an automated road sign recognition system using ANN to extract information written in Bangla. The authors offered a system that takes photos of the road every 2 seconds to detect any road sign.  Using those images, they extracted the text information and road sign information like a picture of a fixed shape. Related textual information boxes had been traced, and their proposed system provided audio feedback of the recognized sign's to the driver. The authors made an API that collects photos from a video every two seconds. Those images then went through some filters and get resized. Then the image text from the images got extracted, the authors fed that information to a multilayer perceptron model (MLP) to recognize the sign. Then based on the output, audio feedback was given. To achieve that, they used an API built on visual basic 6.0. The authors first normalized the images by turning them into grayscale images and resizing them to 320x240 pixels. Then the images went through a 3x3 median filter and an edge-based detection based on the Sobel edge detection filter used for text detection and extraction. After detecting the text from the image, the authors converted the modified images into binary images. Text segmentation were required for identifying the characters from the binary images. Line segmentation techniques and character segmentation techniques are used to determine the individual characters and convert into a feature vector. Then a multilayer perceptron model was used to recognize the characters. If the identified text was present in the record of the predefined list of the road sign, the system gave audio feedback about the detected sign. The authors were able to recognize the road sign with 91.5\% accuracy. This accuracy is higher than the previously proposed model. The success of their text detection and extraction algorithm was 90.6\%. \hl{However, their proposed model had lower time complexity as well and the authors did not identify the road signs without the textual information. This paper can be improved by adding a method that recognises the model from the fixed pictograms to the system.}

\par
In 2019, Uddin \etal \cite{uddin2019extracting} suggested a neural network model for identifying the negative sentences from Bangla sentences. The authors used the LSTM model for their work and created their own datasets. The authors created a model to detect negative sentence pattern in Bangla language with LSTM based model. They applied two-step LSTM. They did the hyperparameter tuning for their model and implemented 10-fold cross-validation on the best LSTM model they got by tuning the hyperparameters. They also build a preprocessing algorithm for the Bangla data. For the work, the authors created their dataset collected from the Bangla sentences and no translated sentences are used for their model. They collected the data from Twitter and for preprocessing, they implemented a regex rule. The dataset was mainly labelled into two categories, namely positive and negative. They collected 588 samples of positive and negative sentences for each category for training their model. Their model achieved an accuracy of 73.6\% with the LSTM size of 128. They observed that their model achieved the best accuracy during a 1-2 batch size, and it is around 77.1\%. However, they did not implement the model with batch size 1 or 2 because it was computationally far more expensive. Overall, the authors have implemented an RNN based LSTM model to analyze the negative Bangla sentences. They have achieved a good result with their small dataset with some hyperparameter tuning and cross-validation.\hl{However, the dataset used in this study was quite small. This paper can be improved by increasing the size of the dataset and using a larger feature set to make the model more dynamic.}

\begin{table*}
\centering
\caption{Summary Table for Information Extraction.}
\label{tab:information} 
\resizebox{\linewidth}{!}{%
\begin{tabu}{>{\hspace{0pt}}p{0.089\linewidth}>{\hspace{0pt}}p{0.470\linewidth}>{\hspace{0pt}}p{0.400\linewidth}} 
\toprule
\multicolumn{1}{>{\hspace{0pt}}m{0.089\linewidth}}{} & \multicolumn{1}{>{\hspace{0pt}}m{0.470\linewidth}}{} & \multicolumn{1}{>{\hspace{0pt}}m{0.400\linewidth}}{} \\
\multicolumn{1}{>{\centering\hspace{0pt}}p{0.056\linewidth}}{\vcell{\textbf{Article}}} & \multicolumn{1}{>{\centering\hspace{0pt}}p{0.488\linewidth}}{\vcell{\textbf{Datasets and Results}}} & \multicolumn{1}{>{\centering\arraybackslash\hspace{0pt}}p{0.415\linewidth}}{\vcell{\textbf{Preprocessing Steps and Methods}}} \\[-\rowheight]
\multicolumn{1}{>{\centering\hspace{0pt}}m{0.056\linewidth}}{\printcelltop} & \multicolumn{1}{>{\centering\hspace{0pt}}m{0.488\linewidth}}{\printcelltop} & \multicolumn{1}{>{\centering\arraybackslash\hspace{0pt}}m{0.415\linewidth}}{\printcelltop} \\ 
\\
\hline\hline
\multicolumn{1}{>{\hspace{0pt}}m{0.036\linewidth}}{} & \multicolumn{1}{>{\hspace{0pt}}m{0.488\linewidth}}{} & \multicolumn{1}{>{\hspace{0pt}}m{0.415\linewidth}}{} \\
\vcell{Sural \etal \cite{sural1999mlp}} &
  \vcell{A 300 dpi flatbed HP scanner was used to scan the Bangla documents from different sources including newspapers, magazines, and novels and the model achieved 98\% accuracy.} &
  \vcell{Hough transformation on the scanned images was applied and a three MLP model that took a fuzzy feature set obtained from the Hough transformation for Bangla script recognition.\newline} \\[-\rowheight]
\printcelltop & \printcelltop & \printcelltop  \\
\vcell{Rahman \etal \cite{rahman2009real}} &
  \vcell{Captured a photo with a webcam every 2 seconds and the accuracy of the system in recognizing the Text detection and extraction was 90.6\% and the road sign was 91.5\%.} &
  \vcell{ Text segmentation, feature extraction was used as preprocessing and character recognition was achieved using an MLP neural network.\newline} \\[-\rowheight]
\printcelltop & \printcelltop & \printcelltop  \\

  \vcell{Mandal \etal \cite{mandal2011handwritten}} &
  \vcell{25 samples of each character was taken and accuracy of the system in recognizing the using scheme one was 87.65\% and using scheme two was 88.95\%.} &
  \vcell{Segmentation of form, Character segmentation and feature extraction were done with gradient computation. Wavelet transform and the character recognition were done using a KNN algorithm.\newline} \\[-\rowheight]
\printcelltop & \printcelltop & \printcelltop  \\

  \vcell{Kundu \etal \cite{kundu2012automatic}\newline} &
  \vcell{Self-collected data from the IT tutorial blog and Bangla sentences and 71.82\% accuracy.} &
  \vcell{Language independent statistical model.} \\[-\rowheight]
\printcelltop & \printcelltop & \printcelltop  \\

  \vcell{Chandra \etal \cite{chandra2013hunting}} &
  \vcell{CIMC dataset and 95.96\% accuracy.} &
  \vcell{Introspection of the sentences of CMIC to identify certain patterns for the rule and rule-based and statistical mode combined method were applied.\newline} \\[-\rowheight]
\printcelltop & \printcelltop & \printcelltop  \\ 

  \vcell{Sharif \etal \cite{7853957}} &
  \vcell{ISI handwritten character database and CMATERDB3.1.1 collection and the model achieved 99.02\% accuracy with 55 epochs.} &
  \vcell{Smoothing operation for decreasing noise, image converted into binary format. Hybrid model containing HOG feature extraction into a two hidden layer ANN plus a multilayer CNN and ten layers of SoftMax function for the outcome has been used.\newline} \\[-\rowheight]
\printcelltop & \printcelltop & \printcelltop  \\

  \vcell{Uddin \etal \cite{uddin2019extracting}} &
  \vcell{Self-made data, collected from the twitter and 73.6\% accuracy.} &
  \vcell{Regex rules to eliminate any roman or unnecessary scripts and LSTM model with LSTM size of 128 and batch size of 5.} \\[-\rowheight]
\printcelltop & \printcelltop & \printcelltop  \\
\bottomrule
\end{tabu}
}
\end{table*}
\subsubsection{Combination of Classical and ML Approaches}

Sural \etal \cite{sural1999mlp} proposed a character recognition from Bangla scripts using an MLP which uses Hough transform-based fuzzy feature extraction. The model uses an MLP model to recognize the character in three stages, and the Hough transform of character is used for feature extraction. The authors first scanned Bangla scripts with an optical scanner and generated a bit image. Then they skewed the scanned data many times for the alignment of the scanner. The Hough transform method was used to solve the problem. The matra is then deleted from the word, and the authors used a recursive algorithm to find the minimum containing rectangles. Then they used the segmented image in the MLP model to identify the character in three steps. In the first stage, only vowels, consonants, and modifiers were detected. Then, gradually in the second and third stages, the yuktakhars were detected if a specific threshold value is not reached in the first stage. The model was tested with the legal document image where the noise was created artificially. The authors created a noisy image using a two-step Markov chain where a random state produces the error, and the brush stage corrupts the image pixel. They collected Bangla documents from magazines, newspapers, novels using a 300 dpi flatbed hp scanner. They managed to get an overall accuracy of 98\% for their proposed model. The authors found it challenging to get a good outcome because of the paper's low quality texts or defective scan of the documents from their observation.\hl{However, the authors did not give enough information on what type of data had been used. This paper can be improved by introducing a more sophisticated preprocessing for the image for sparsely yuktakhars.}
\par
In 2011, Mandal \etal \cite{mandal2011handwritten} suggested a model to determine the handwritten Bangla characters in printed forms. Their model worked based on the combination of the gradient feature and Haar wavelet \cite{mandal2011handwritten} configuration. Then combining two segments, the authors created a feature set, and they have used a KNN classifier to identify the characters. The authors proposed a hybrid model based on gradient features and coefficients of the wavelet transform to recognize characters inside a printed form. They segmented the machine-printed portion of the form first and then segmented the character using a morphological opening operation. After that, a Sobel gradient operator was used to identify the grayscale image’s gradient. The authors also used the Haar wavelet for finding the coefficient of the wavelet transform of the character. By combining two features, they created a feature vector. Then by using the KNN algorithm, they determined the character. A 1-NN algorithm was used in general. If a tie arises for a similar outcome to break the tie, a 3-NN or 5-NN algorithm was used. For the datasets, the authors used 25 samples of 49 symbols in the Bangla language. They used 4,372 instances for the training of the model and evaluate their model with 46 Bangla forms, and each containing character forms 83 to 112. The authors used two types of methods in making the combined feature vector. In both cases, they managed to gain accuracy of over 87\%. In their first method, the normalized image was decomposed by level three Haar wavelet, and they gained an accuracy of 87.65\%. The second method used the level 2 Haar wavelet transformation directly with the feature set and achieved an accuracy of 88.95\%. \hl{However, in the training phase of this study, the sample size is insufficient. More training samples could be used in the training phase to improve this paper.}
\par

Sharif \etal \cite{7853957} suggested a hybrid CNN model for detecting the Bangla numerical digit recognition. The model combines CNN-based models with a Histogram of Oriented Gradient (HOG) feature. The model was trained with the Bangla numerals dataset, and the authors evaluated untrained datasets for evaluation. The authors proposed a hybrid model consisting of two parts. In the first case, two hidden layers of the respective 32 and 64 sized ANN take feature vectors extracted from an image as input. This feature has been acquired from the HOG. In the second part, a conventional multi-layered CNN has been used, which has a filter of 32x3x3 and a max-pooling layer of 2x2. The authors used the output of these two-parts to make a combined larger feature vector. The model finally goes through a softmax function of ten layers. As for the dataset, the authors used the Bangla numeral datasets from the Indian Statistical Institute (ISI) handwritten character database and CMATERDB3.1.1 collection. The authors used image augmentation for ISI datasets, containing 19,392 training images and 3,986 testing images. However, by applying augmentation to the ISI training dataset images, the authors then created a dataset of 58,176 to train the model. They used this dataset for training the model, and the CMATERDB3.1.1 contained 6,000 images used for the testing of the model. The model gained good accuracy using less computation power, which is the proposed model's key feature. Their model reached an accuracy of 99.02\% in only 55 epochs. There is another work by Sazal et al. \cite{sazal2014bangla} where authors used DBN to recognize Bangla handwritten characters. \hl{However, a limited dataset was used, and the evaluation was done with an excellent noiseless set. This paper can be improved by enlarging the dataset size and by using more handcrafted feature extraction.} \par
Table \ref{tab:information} shows the short description of the articles of Bangla ionformation extraction systems.

\subsection{Machine Translation}
\hl{Throughout civilizations, humans have created language to communicate with each other. The language differs from region to region. To understand a language for a person who does not understand the language we use a translator. Machine Translation is a technique to translate one language to another through computer processing.}
\subsubsection{Classical Approaches}
 Ali et al. \cite{ali2002development} presented a procedure for developing machine translation dictionaries. The authors also provided some morphological rules that helped to develop a Bangla machine translation dictionary. The procedure of developing a Bangla machine translation dictionary by the authors goes as follows:  determining contents of the dictionaries, determining the detail of information, determining the organization of the dictionaries. The authors discussed that the domain of the application affects the contents of the dictionaries. The detailed information of the dictionaries depends on rules and restrictions that are imposed. The morphology plays an important role in structuring the words and forming the sentences. The organization of the dictionaries was signified by the authors as the storing and availability of information and the extendibility also plays an important role in developing the machine translation dictionary. Later in the paper, the authors discussed various contents for Bangla machine translation dictionaries such as grammatical properties and Bangla morphology. The authors of this paper primarily focused the development procedure of a machine translation dictionary and discussed various factors related to the points. \hl{However, no implementation was provided.} The authors denoted that generalized morphological rules can reduce the size of dictionaries. The significance of morphological rule was reflected in the paper as the authors also provided some morphological rules.

\begin{table*}
\centering
\caption{Summary Table for Machine Translation.}
\label{tab:machine-translation}
\begin{tabular}{>{\hspace{0pt}}p{0.221\linewidth}>{\hspace{0pt}}p{0.35\linewidth}>{\hspace{0pt}}p{0.371\linewidth}} 
\toprule
\\
\multicolumn{1}{>{\centering\hspace{0pt}}m{0.221\linewidth}}{\textbf{Article}} & \multicolumn{1}{>{\centering\hspace{0pt}}m{0.35\linewidth}}{\textbf{Datasets and Results}} & \multicolumn{1}{>{\centering\arraybackslash\hspace{0pt}}m{0.371\linewidth}}{\textbf{Preprocessing Steps and Methods}} \\ 
\\
\hline\hline
\vcell{Rabbani \etal \cite{rabbani2014new}} & \vcell{Comparing with Google translation and Anubadok online gave a satisfactory result.} & \vcell{Methods applied: Verb based approach.} \\[-\rowheight]
\\
\printcelltop & \printcelltop & \printcelltop \\
\vcell{Chowdhury \cite{chowdhury2013developing}} & \vcell{Unicode based document corpus was used.} & \vcell{Methods applied: A rule-based approach using parts of speech tagging.} \\[-\rowheight]
\\
\printcelltop & \printcelltop & \printcelltop \\
\vcell{Ali et al. \cite{ali2002development}} & \vcell{Machine translation dictionaries.~} & \vcell{Methods applied: Morphological rules.} \\[-\rowheight]
\\
\printcelltop & \printcelltop & \printcelltop \\
\vcell{Ali \etal \cite{ali2008specific}} & \vcell{Web documents.~} & \vcell{Methods applied: Rule-based approach.} \\[-\rowheight]
\\
\printcelltop & \printcelltop & \printcelltop \\
\vcell{Islam \etal \cite{islam2010english}} & \vcell{EMILLE, KDE4 and Prothom-alo corpus were used. BLEU, NIST and TER score of 11.70, 4.27 and 0.76.} & \vcell{Pre-processing steps: Parallel corpus aligning, manual cleaning of the corpus. Methods applied: N-gram, 5-gram, 8-gram.} \\[-\rowheight]
\\
\printcelltop & \printcelltop & \printcelltop \\
\vcell{Francisca \etal \cite{francisca2011adapting}} & \vcell{The least error rate for adaptive 0.7666\% and non-adaptive 0.8544\% using bootstrap.} & \vcell{Pre-processing steps: Tokenization. Methods applied: Rule-based (fuzzy rules).} \\[-\rowheight]
\\
\printcelltop & \printcelltop & \printcelltop \\
\vcell{Anwar \etal \cite{anwar2009syntax}} & \vcell{The success rate of simple, complex and compound sentences are 93.33\%, 92.6\%, and 91.67\% respectively.} & \vcell{Pre-processing steps: Tokenization. Methods applied: Rule-based approach.~} \\[-\rowheight]
\\
\printcelltop & \printcelltop & \printcelltop \\
\\
\bottomrule
\end{tabular}
\end{table*}

 Ali \etal\cite{ali2008specific} demonstrated ways to link the Bangla language to Universal Networking Language (UNL)\cite{uchida2001universal}. The conversion between Bangla and UNL would allow Bangla documents to be converted into any language and vice versa. The authors provided an outline to make a Bangla-UNL dictionary, annotation editor for Bangla texts, morphological rules etc. The goal of the authors was to eliminate the huge task of translation between Bangla and any other languages by converting the Bangla text to Universal Networking Language. The core structure of UNL contains six elements: universal words, attribute labels, relational labels, UNL expression, hypergraph, knowledge base. To build a Bangla-UNL dictionary the authors analysed various Bangla morphological structures to reduce dictionary entries. The authors suggested the annotation of web documents can be done scientifically by trained people in that subject. They also suggested that the caption of the news can be used as an annotation. The UNL-Bangla dictionary requires morphological analysis and for that the authors provided some structures to get the root words. They also provided UNL structure for the structures. For syntactic analysis and semantic analysis, the authors provided some rules. The paper provided a suitable guideline to start working for tasks related to converting Bangla texts to UNL. The authors did not implement anything but provided an outline so that can be used for future work.
\par 
Anwar \etal\cite{anwar2009syntax} proposed a method that analyzes the Bangla sentences syntactically accepting all types of Bangla sentences and performs machine translation into English using an NLP conversion unit. The rules imposed in this proposed method allows five categories of sentences for parsing. The proposed method performs the first task by tokenizing the sentences. Then the syntax analyzer analyzes Bangla grammatical rules. For representing the grammar, a set of production rules was used. The lexicon of this method contained a priori tag and suffix for words. After that, a parse tree was generated for the given input string. The analysis and conversion into a parse tree were done by an NLP conversion unit. The NLP conversion unit used a corpus containing large amounts of English sentences as training corpora. Then the output sentence was generated. The result of the system showed that the success rate of simple, complex and compound sentences are 93.33\%, 92.6\% and 91.67\% respectively. The authors of this paper proposed a machine translation technique combined with syntax analysis that can translate any type of Bangla sentences into English sentences. \hl{The system generated satisfactory results. However, the system was unable to handle idioms and phrases and mixes Bangla sentences.}Adding the capability of handling mixed sentences and using a bigger dataset can improve the system drastically.

\par
In 2011, Francisca \etal\cite{francisca2011adapting} presented a rule-based adaptive Machine translation system from English to Bangla. The proposed system is based especially on \hl{fuzzy rules}\cite{wang1992generating}.  The translation process works by classifying the English sentences into a particular class and then translating them into Bangla using rules. At the beginning of the methodology, the authors analysed the essential grammatical rules of English grammar. After that, the authors performed a comparative analysis of structures between English and Bangla. After that morphological analysis was done during mapping of English to Bangla. After that, a set of rules was imposed for translation. In this part the English sentence was tokenized by words, then after lexical analysis fuzzy rule was matched for structuring the sentences for translation. Then the dictionary was used to translate the corresponding English words to Bangla. Finally, the translated part was reconstructed into Bangla sentences using corresponding rules. Thus, the final result was provided. The authors used three methods to compare the experiment they conducted. Those methods are: Hold out, Cross-validation and Bootstrap. Among them, the bootstrap had the least error rate for adaptive 0.7666\% and non-adaptive 0.8544\%. The result provided by the authors supported the claim made by them. However, a bigger dataset could clarify their claim robustly. The rule-based approach provided by the authors in this paper was able to translate English sentences into Bangla sentences. Testing in a bigger dataset and reducing complexity can improve the method. 

\par
Chowdhury \cite{chowdhury2013developing} presented an approach to develop Bangla to English Machine translation using parts of speech tagging. The article also shows some methods that can resolve pronouns to summarize the text so that coherence and important information are conserved. The methodology of the proposed method is divided into three parts: parts of speech tagging, Bangla, and English rule implementation, and finally, the English sentence generation. The tag vector for tagging consists of sixteen bits, and three bits out of them were kept for parts of speech. To implement grammatical rules to structure a proper sentence, the rules of Bangla and the English language were compared and implemented by generating an algorithm. The algorithm works by translating individual words using a dictionary, determining the subject of the sentence, and finally, implementing the grammatical suffixes. In the end, the words were rearranged following Bangla to English grammar rules. The author gave an example of the implementation of the method but did not provide any test result. The author expressed a new approach to machine translation from Bangla to English in this paper. The method was described as efficient for simple sentences. The paper provides a unique approach to translate Bangla into English machine translation.\hl{The performance analysis of the system was not provided.} The author’s claim could be robust if any performance analysis were provided and tested, comparing different methods. Providing a performance analysis could provide clarification of the proposed method.

\par
In 2014, Rabbani \etal\cite{rabbani2014new} proposed a verb-based machine translation technology to translate the English language to the Bangla language. The verb-based approach detects the main verb of the sentence given in English, then binds other parts of speeches as subject and object. The authors successfully implemented the technique in assertive, interrogative, imperative, exclamatory, active-passive, simple, complex, and compound English sentences. The methodology of this verb-based approach has several steps. The first step is to translate individual English sentences if there is a paragraph. The second step performs lexical analysis on the texts. In the third step, words are bound and tagged based on the previous step, which is an iterative process. The fourth step is determining the verbs in a sentence using VBMT (Verb Based Machine Translation) \cite{rabbani2014new}. In the next step, VBMT works on defining the Bangla sentence structure in correspondence to the English sentence. In the last step, VBMT is used to generate the final Bangla sentence based on the modified English sentence structure obtained from the previous step. The result was obtained by the authors comparing their machine translation with Google translate and Anubadok online. The authors claimed that the proposed machine translation technique was able to translate and give a satisfactory result. The machine translation technique proposed by the authors in this paper suggests a verb-based approach. But the limitations of preposition binding and phrase binding gives a worse result than the compared two systems while testing. The authors expressed the future intention to make improvements, including the preposition bindings and appropriate tagging of Parts Of Speech of a word within a sentence. The paper expresses a proposition of a machine translation technique based on a verb-based approach. The authors also expressed some limitations regarding the approach, like limitations in preposition and phrase binding. Resolving preposition and phrase binding limitations and improving POS tagging using a rich dataset can improve the paper.

\subsubsection{Machine Learning Approaches}

\begin{figure}
    \centering
    \includegraphics[width=\linewidth,height=6.5cm]{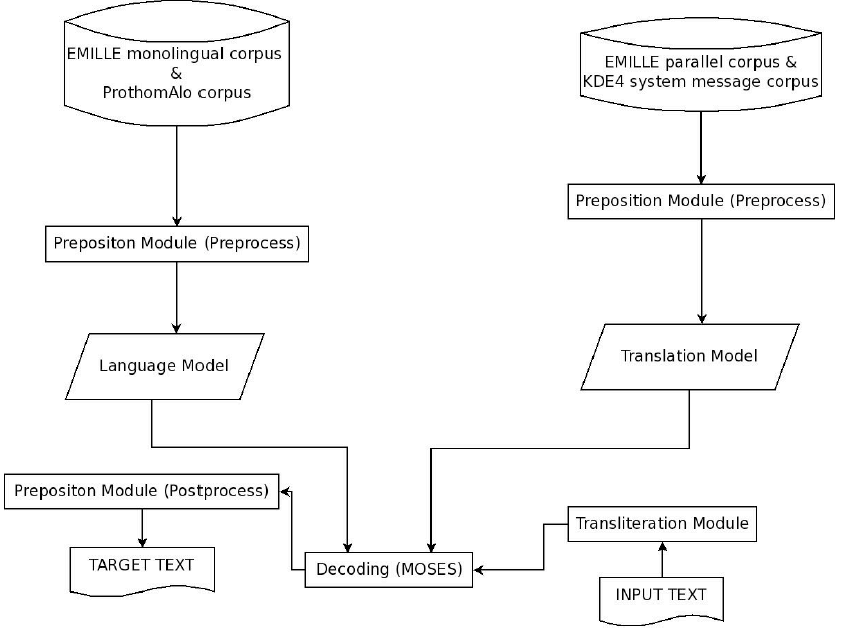}
    \caption{System Architecture of Phrase based Machine Translation \cite{islam2010english}.}
    \label{fig:machine_t_5}
\end{figure}

Islam \etal \cite{islam2010english} proposed a phrase-based machine translation system for translating the English language to Bangla language. Two additional modules were created, for this reason, one is a transliteration module and another is a preposition handling module. The experimental results were obtained in different metrics. The methodology proposed by the authors gives importance to two points: handling preposition and transliteration. That is why two dedicated modules were used in the proposed system. The system architecture is illustrated in Figure \ref{fig:machine_t_5}. For the dataset, a parallel corpus of South Asian languages having 12,654 English and 12,633 Bangla sentences were used. The preposition handling module was divided into pre-processing and post-processing. The training corpus and testing corpus were developed individually. For training, a parallel training corpus containing 10,850 sentence pairs were used. The baseline system was built in a 5-gram language model and tested on a corpus. The corpus was cleaned by an alignment tool and hard manual labour. After that, a new translation system was made with an 8-gram model. The transliteration model was developed to increase the accuracy. A collection of 2,200 unique names from Wikipedia and Geonames was used to make sure nouns are not translated. The implementation of the prepositional module used the intersection of word alignment. These post-translation corpora were pre-processed in two steps: coming up with 19 postpositional words, coming up with a group of 9 suffixes for attaching nouns. The result compared with Anubadok shows the proposed system had \hl{ BLEU} \cite{papineni2002bleu} ,\hl{ NIST}\cite{przybocki2009nist} and TER score of 11.70, 4.27, and 0.76 respectively where the Anubadok scored 1.60, 1.46 and 1.03 respectively. The evaluation of the final combined system shows the KDE4 corpus performs better than EMILLE \cite{islam2010english}. \hl{The system handles prepositions very well.}
\par
The authors expressed that their proposed system produced satisfactory results but has some limitations.\hl{The system handles prepositions very well.} The parallel corpora for English and Bangla were not enough.  The system works well for short sentences. Also, some verb-noun ambiguity occurs which was not resolved. The satisfactory results obtained were good for low-density languages like Bangla.  The experimentation process includes using various N-gram\cite{doddington2002automatic} models for testing. The paper signifies handling prepositions very well. The overall performance of the proposed system provided a satisfactory result. Handling compound words, using a bigger corpus and extension of preposition handling can make the paper better.

Table \ref{tab:machine-translation} covers overall datasets, results, pre-processing steps and methods used in machine translation papers.

\subsection{Named Entity Recognition}
\hl{The task of identifying and categorizing phrases into specific classes of named entities, such as names of people, organizations, and locations, is known as named entity recognition \mbox{\cite{sasano2008japanese}}. Named entity recognition has several applications, including machine translation, text-to-speech synthesis, natural language comprehension, information extraction, information retrieval, question answering, and so on \mbox{\cite{morwal2012named}}.}
\begin{table*}
\centering
\label{tab:name-entity}
\caption{Summary Table for Named Entity Recognition.}
\label{tab:namedentity}
\begin{tabu}{>{\hspace{0pt}}m{0.160\linewidth}>{\hspace{0pt}}m{0.400\linewidth}>{\hspace{0pt}}m{0.336\linewidth}} 
\toprule \\
\multicolumn{1}{>{\centering\hspace{0pt}}m{0.160\linewidth}}{\textbf{Article}} & \multicolumn{1}{>{\centering\hspace{0pt}}m{0.400\linewidth}}{\textbf{Datasets and Results}} & \multicolumn{1}{>{\centering\arraybackslash\hspace{0pt}}m{0.336\linewidth}}{\textbf{Preprocessing Steps and Methods}} \\ \\
\hline
\hline

\vcell{Chaudhuri \etal \cite{chaudhuri2008experiment}} &
  \vcell{E-newspapers of Anandabazar Patrika for the years 2001-2004 were used as the datasets and the result was presented in the form of recall (R), precision (P), and F-measure, respectively 85.5\%, 94.24\%, and 89.51\%.} &
  \vcell{Word level morphological parser, combination of dictionary-based, rule-based, and statistical (N-gram based) approaches.} \\[-\rowheight]
\\
\printcelltop & \printcelltop & \printcelltop \\
\vcell{Chowdhury \etal \cite{chowdhury2018towards}} &
  \vcell{Location specified dataset BCAB with seven entities and created a black list of words for Bangla and the model achieved a F1 score of 0.58 and for the partial match of domain the model achieved 0.72.} &
  \vcell{Created a dataset with seven entities and annotated them manually. Used a CRF’s based model and a DNN model for comparison.} \\[-\rowheight]
\\
\printcelltop & \printcelltop & \printcelltop \\
\vcell{Banik \etal \cite{banik2018gru}} &
  \vcell{Dataset collected from the “Prothom Alo” newspaper and the model achieved F1 score of 69 with 50 epochs.} &
  \vcell{Url remover, punctuation character removal, text segmentation and 3 layer GRU based RNN model are used.}
\\[-\rowheight]
\\
\printcelltop & \printcelltop & \printcelltop \\
\\
\bottomrule
\end{tabu}
\end{table*}
\subsubsection{Classical Approaches}
In 2008, Chaudhuri \etal \cite{chaudhuri2008experiment} proposed a Name Entity (NE) detection process for the Bangla language. The authors described a three-staged approach for name entity recognition, containing dictionary-based, rule-based, and statistical-based approaches. In their proposed model, first, a given word goes through a morphological analyzer for stemming. Then the given word goes through the dictionary which maintains one of the three categories of tags attached to each word. If any match occurs for the word in the dictionary then the algorithm search for the tags otherwise it put a weight on the word and send it to the next phase where rule-based analysis is done. The rule for name entity can be positive or negative and based on the fact the authors suggested 12 different rules for named entity. A Weight is added to the word if it maintains certain rules. To be certain if a word is a name entity, a certain threshold value is set and to be accepted as name entity the weight of the word has to be greater than the threshold weight. If a word is not a name entity another threshold weight is set. If the weight of the test word is less than this threshold weight, then it must be less than the threshold value and it is not a name entity. Otherwise, the word is sent for the next stage, a statistical-based approach where the N-gram method is used. To generate the N-gram model, the NE words have been tagged manually. From these tags, words left and right neighbor words have been checked (for 2-N-gram Model). The frequency of each pair is calculated from the corpus. If a particular pair of neighbors occur about a word, then based on the corpus probability of name entity have been decided. To form the corpus Anadabazar Patrika was used as the resource. The data obtained from the newspaper cannot be used directly as they were in the glyph code. So these texts were transformed into ISCII format and the result of their work has been evaluated based on the recall, precision, and F-measure which is respectively, 85.5\%, 94.24\%, and 89.51\%. \hl{The proposed technique, on the other hand, was trained on tiny datasets and evaluated on controlled datasets.}

\subsubsection{Machine Learning Approaches}

Banik \etal \cite{banik2018gru} suggested a name entity recognition method. The authors used a recurrent neural network for their model, and they have used a gated recurrent unit in their model for the recurrent neural network. They used data collected from the newspaper for their purpose. The  authors presented a simple, RNN based network that has three layers. To solve the vanishing gradient problem in the simple RNN network, they used GRUs. They have used one-hot encoded vocabulary for their system. The hidden state in the RNN would change based on the input size in the network. They used a non-linear activation function \hl{Relu}\cite{hanin2019universal}. In their NER system, the authors focus on four name entity criteria like person, L
location, organization, and day. They created a dataset containing data collected from a newspaper. Then they annotated the data based on those four criteria. They have used some prepossessing for their model such as removal of URL and punctuation character and segmented the text in the meaningful entity. The authors achieved an F1 score of 69 for their model. The result they obtained is from 50 epochs. Their model worked in the predefined form of text documents. Their model has a smaller number of hidden units due to the fewer data. Also, the limited dataset has an impact on the results. \hl{The proposed method, on the other hand, lacked relevant dataset specifics, and the data was also insufficient.}

\subsubsection{Combination of Classical and ML Approaches}

Chowdhury \etal \cite{chowdhury2018towards} presented a NER system that specially focuses on the Bangladeshi Bangla. They used word-level, POS, gazetteers, and contextual characteristics, together with the conditional random fields for the primary model. They also observed the performance of a \hl{ DNN} \cite{canziani2016analysis} to compare their base model. They made their primary dataset based on Bangladesh and created a dataset for the proposed model by following the Automatic Content Extraction (ACE) \cite{chowdhury2018towards} and MUC6 guideline. In their dataset, there were seven entity types. Their contribution becomes a part of Bangla Content Annotation Bank (B-CAB) \cite{chowdhury2018towards}. For collecting data, they selected different newspapers from Bangladesh for the regional difference. Their model explored four different feature sets: word level, which has primarily provided morphological information of the word, POS for making parts of speech taggers, list of lookup features searches for any named entity from a previously created corpus, and finally, a word embedding feature has been used. They also used a DNN model to compare the CRFs model's performance, whether the modern machine learning algorithm exploits the results more or not. The model was run with India's different geographic datasets to show the significance of the NER system's geographic location. They have evaluated the prediction based on the exact and partial match. The CRFs model achieved an F1 score of 0.58 and 0.72 respectively for the domain's exact and partial match and their proposed model performed better than a bidirectional LSTMs-CRFs architecture. \hl{Moreover, further research is ongoing in this field under Bangla Content Annotation Bank (B-CAB).} \par
Table \ref{tab:namedentity} shows the short description of the articles of Bangla named entity recognition system.

\subsection{Parsing}
\hl{A sentence of a language is made up according to the syntax of the grammatical rule provided by the language. Parsing in Natural Language processing works on determining the syntactic structure of the sentence according to the grammatical rules of that language.}
\begin{table*}
\centering
\caption{Summary Table for Parsing.}
\label{tab:parsing-fuad}
\begin{tabular}{>{\hspace{0pt}}p{0.2\linewidth}>{\hspace{0pt}}p{0.441\linewidth}>{\hspace{0pt}}p{0.294\linewidth}} 
\toprule
\\
\multicolumn{1}{>{\centering\hspace{0pt}}m{0.2\linewidth}}{\textbf{Article}} & \multicolumn{1}{>{\centering\hspace{0pt}}m{0.441\linewidth}}{\textbf{Datasets and Results}} & \multicolumn{1}{>{\centering\arraybackslash\hspace{0pt}}m{0.294\linewidth}}{\textbf{Preprocessing Steps and Methods}} \\ 
\\
\hline\hline
\vcell{Das and Bandyopadhyay \cite{das2010morphological}} & \vcell{Indian languages to Indian languages machine translation system’s (IL-ILMT) gold standard morphological dataset. The result had an accuracy of 74.6\%.} & \vcell{Methods Applied: K-means clustering, simple suffix stripping algorithm, score based clustering algorithm.} \\[-\rowheight]
\\
\printcelltop & \printcelltop & \printcelltop \\
\vcell{Dasgupta \etal \cite{dasgupta2005morphological} [2005]} & \vcell{Grammar book examples of compound words were used as the dataset. Result had an accuracy of 100\%.} & \vcell{Methods applied: Feature unification based morphological parsing.} \\[-\rowheight]
\\
\printcelltop & \printcelltop & \printcelltop \\
\vcell{Mehedy \etal \cite{mehedy2003bangla}} & \vcell{No specified dataset. The model can parse all kind of Bangla sentences.} & \vcell{Context free grammar for making parse tree.} \\[-\rowheight]
\\
\printcelltop & \printcelltop & \printcelltop \\ \\
\vcell{Dasgupta \etal \cite{dasgupta2004morphological} [2004]} & \vcell{No specified dataset. The model was used for the morphogolical parsing.} & \vcell{The open-source parser PC-KIMMO was used for the morphological parsing.} \\[-\rowheight]
\printcelltop & \printcelltop & \printcelltop \\
\vcell{Saha \etal \cite{saha2006parsing}} & \vcell{No specifica dataset. The parser can identify parts of speech in Bangla sentences.} & \vcell{The parser was created based on the grammatical rules of Bangla language.} \\[-\rowheight]
\\
\printcelltop & \printcelltop & \printcelltop \\ \\
\bottomrule
\end{tabular}
\end{table*}
\subsubsection{Classical Approaches}

Dasgupta \etal\cite{dasgupta2005morphological} proposed a morphological parser of compound words of Bangla language. The authors used a feature unification based morphological parser for parsing compound words. The authors of this paper used finite-state morphological parsing which is based on Kimmo Koskenniemi’s two-level morphology technique \cite{dasgupta2005morphological}. There are 3 components for this parsing system: lexicon and morphotactic, morphophonology and word-grammar component. The authors prepared finite state machines for compound words. The ambiguity of two parse trees was resolved by defining two new features and implementing feature unification to ensure a single parse tree. The nominal and pronominal inflections were classified into five categories. The features were used to modify the lexicon. After that, the compound words with inflectional suffixes were categorized into four different categories. The final grammar was generated in PC-Kimmo format. The implementation of the morphological analyser was done in PC-KIMMO version 2. The result obtained in the implementation shows a 100\% correct result on compound words found in Bangla grammar books. The paper’s approach to deal with compound words in Bangla sentence was unique. \hl{The evaluation process was not strong enough.}The methodology would be better judged if they used a wide range of test samples to evaluate the result of their proposed method. 
\par 
In 2003, Mehedy \etal \cite{mehedy2003bangla} suggested a context-free grammar for generating a parse tree for the Bangla sentence. The authors have proposed a model that can parse all five categories (assertive, interrogative, imperative, exclamatory, and optative) of Bangla sentences. The parser analysis can be divided into three categories: the lexical analysis phase, the syntax analysis phase, and the semantic analysis phase. In the lexical analysis, the characters of sentences are scanned sequentially and divided into tokens. Then in the syntax analysis, the validity of the sentence is checked according to grammar. Lastly, in the semantic phase ensures that the discrete input components fit together meaningfully. The authors maintained all these analyses during their creation of context-free grammar. Then the authors created context-free grammar for all possible combinations of the sentences such as the structural variations of a sentence. They considered the intonation difference for the sentences. Also, they considered the indicative form of sentences and created different context-free grammar for each case to make a parser. 
\par

Dasgupta \etal \cite{dasgupta2004morphological} suggested a morphological parsing of Bangla words. The authors used PC-KIMMO to establish his model. For constructing morphological parsers, there are three essential equipment like lexicon, morphotactics, and orthographic rules. For the lexicon section, the authors collected a list of stems and affixes, together with necessary information about them. In the morphotactics section, the authors established a rule where it described the order of the morpheme, in other words, which morpheme classes can follow other classes. Then the authors applied orthographic rules for his parser. Then they applied a two-level parsing where the first part was a rules component, and the other part was a lexical component of the lexicon. Finally, the authors implemented the open-source PC-KIMMO for the construction of the parser. For their model, they used basic sets of grammar for Bangla. 
\par
Saha \cite{saha2006parsing} proposed a parser for identifying the POS for the Bangla lexicons. Also, his suggested model annotated the Bangla sentence with semantic information. The model was built on a rule-based method to produce a Bangla parser. Bangla has no small or capital concepts. Thus it does not become easy to find a proper noun based on the characteristics of the characters. Also, in Bangla, a word can be used as different parts of speech. It is also possible for a word to be a proper noun or abstract noun based on a particular sentence. It is observed that common words have different senses based on the use cases. This word disambiguation also becomes an uphill struggle in creating a parser. Thus, the author has suggested an intelligent parser that can deal with these challenges. In the time of creating the parser, the proposed model starts annotating POS to each word. If a particular word is not present in the lexicon, then it is tagged as a proper noun. If, when generating the parse tree, any POS-ambiguity happens, then it has been solved by Bangla grammar rules. In the case of word sense ambiguity, the author used the N-gram model to define the word's correct sense. 

\subsubsection{Combination of Classical and ML Approaches}

In 2010, Das and Bandyopadhyay \cite{das2010morphological} presented a stemming cluster-based morphological parsing technique of Bangla words. Two types of algorithms were experimented for this purpose: simple suffix stripping algorithm and score based stemming cluster identification algorithm. Indian languages to Indian languages machine translation system’s (IL-ILMT) gold standard morphological dataset was used for evaluation. The authors used two types of morphological clustering techniques: agglutinative suffix stripping and minimum edit distance using a suffix list. Four lists were made based on parts of speech on noun, adjective, adverb and verb. The simple suffix stripping algorithm used here works well only for nouns, adverbs and adjectives. A small list consisting of 205 suffixes for Bangla was generated manually. The suffix stripping algorithm checked if a word has suffixes from the list and the cluster is assigned assuming the root word. For verb minimum edit distance resolves the stemming problem for inflected verb words. Total six times the system performed an iteration and generated a finite numbers of stem clusters. A separate list of verb inflections were maintained for validation. The standard K-means clustering technique was used for this purpose. The evaluation of their present system was performed on the IL-ILMT's gold standard Morphological dataset. The accuracy of the system was reported at 74.6\%. The paper proposed a new morphological parsing technique using clustering. \hl{However, the evaluation process was not simplified.} Testing other clustering techniques may improve performance of the system. 

Table \ref{tab:parsing-fuad} covers overall datasets, results, pre-processing steps and method used in Parsing papers.

\subsection{Parts of Speech Tagging}
\hl{Every language has some grammatical rules which determine and tags what type of words are used in a text. Parts of Speech Tagging is a part of Natural Language Processing that classifies and tags different words in a sentence according to that particular human language.}
\begin{table*}
\centering
\caption{Summary Table for Parts of Speech Tagging.}
\label{tab:pos-tagging}
\begin{tabular}{>{\hspace{0pt}}p{0.18\linewidth}>{\hspace{0pt}}p{0.455\linewidth}>{\hspace{0pt}}p{0.301\linewidth}} 
\toprule
\\
\multicolumn{1}{>{\centering\hspace{0pt}}m{0.18\linewidth}}{\textbf{Article}} & \multicolumn{1}{>{\centering\hspace{0pt}}m{0.455\linewidth}}{\textbf{Datasets and Results}} & \multicolumn{1}{>{\centering\arraybackslash\hspace{0pt}}m{0.301\linewidth}}{\textbf{Preprocessing Steps and Methods}} \\ 
\\
\hline\hline
\vcell{Ismail \etal \cite{ismail2014developing}} & \vcell{1,000,000 words from various newspapers, blog posts, and articles were used. Successfully tag 134,749 nouns, 11,067 verbs and 8,435 adjectives out of 320,443.} & \vcell{Pre-processing steps: Filtering unique words. Methods applied:~ Semi-supervised method using hash mapping.} \\[-\rowheight]
\\
\printcelltop & \printcelltop & \printcelltop \\
\vcell{Hammad Ali \cite{ali2010unsupervised}} & \vcell{Prothom-Alo articles total to about 50000 tokens and the tagged subset consists of 18,110 token and 4,760 word types were used. Baum-Welch algorithm failed to obtain HMM parameters.} & \vcell{Pre-processing steps: Cleaning the corpora to remove unnecessary data. Methods applied: Baum-Welch algorithm, HMM.} \\[-\rowheight]
\\
\printcelltop & \printcelltop & \printcelltop \\
\vcell{Hasan \etal \cite{hasan2007comparison}} & \vcell{Corpus of 5,000 words from prothom alo were used. Brill’s tagger performed best among other taggers.} & \vcell{Methods applied: N-gram, Hidden Markov Model, Brill’s Tagger.} \\[-\rowheight]
\\
\printcelltop & \printcelltop & \printcelltop \\
\vcell{Hoque~ and Seddiqui \cite{hoque2015bangla}} & \vcell{Prothom-Alo articles, around 8,155 words, and 585 sentences were used.~93.7\% tagging accuracy on dictionary + stemmer + verb + dataset + rules.} & \vcell{Pre-processing steps: Stemming with custom rules. Methods applied:~ Rule-based approach.} \\[-\rowheight]
\\
\printcelltop & \printcelltop & \printcelltop \\
\vcell{Chowdhury \etal \cite{chowdhury2004parts}} & \vcell{Performance complexity of $n\log(n)$.} & \vcell{Methods applied:~ Rule-based morphological approach.} \\[-\rowheight]
\\
\printcelltop & \printcelltop & \printcelltop \\
\vcell{Chakrabarti~ and Pune CDAC \cite{chakrabarti2011layered}} & \vcell{The tag set is both from the common tag set for Indian Languages and IIIT Tag set guidelines.} & \vcell{Methods applied:~ Rule-based approach, morphological analysis.} \\[-\rowheight]
\\
\printcelltop & \printcelltop & \printcelltop \\
\\
\bottomrule
\end{tabular}
\end{table*}

\subsubsection{Classical Approaches}

Hasan \etal\cite{hasan2007comparison} presented a comparative analysis of different parts of speech tagging techniques (N-gram, HMM \cite{rabiner1986introduction}, and Brill's tagger\cite{brill1992simple}). The authors aimed to maximize the performance on limited resources. The comparison was performed on both English and Bangla to understand which can manage a substantial amount of annotated corpus. At the beginning of methodology first, the tagset was built with 12 tags (noun, adjective, cardinal, ordinal, fractional, pronoun, indeclinable, verb, post positions, quantifiers, adverb, and punctuation). This was the high-level tagset. The second level of tagset was based on 41 tags. The training corpus was built with 5,000 words from daily newspaper prothom alo. The training corpus consisted of 4,484 words and the rest was on the testing set. Then the corpus was trained with various POS taggers. The English POS taggers had high-performance 96\%+ where the Bangla POS taggers did not perform well, some only 90\%. The Brill’s tagger performed best among other methods. The authors compared various parts of speech tagging techniques to figure out which technique performs best on limited resources. The authors figured out that the techniques perform similarly on similar-sized corpus in the English language. So, the authors believe that if a bigger corpus was used Bangla POS taggers would perform the same. This paper answers the question of the best POS tagger available for the Bangla language. Also, the paper concluded the necessary improvement concept as the authors figured out the corpus size makes an effect on POS taggers performance. \hl{The training dataset seems to be limited.} Using a bigger dataset and including other POS taggers for comparison could improve the paper.

\par
In 2011, Debasri and Pune CDAC\cite{chakrabarti2011layered} proposed a rule-based Parts of Speech tagging system for the Bangla language. The POS tagger had included layered tagging. The author proposed 4 levels of tagging which also handles the tagging of multi verb expressions. The methodology of the POS tagger starts with morphological analysis of the words. In the beginning noun, analysis and verb analysis had been done. Then the suffixes were classified based on number, postposition and classifier information. Verbs are classified into 6 paradigms based on the morphosyntactic alternation of the root. The suffixes are further analysed for person and honorific information. The ambiguity between a cardinal and a noun was resolved by a rule. Then the POS tagger was made to go through 3 stages. For multiverb expression, the POS tagger goes through additional steps. There was no result of the analysis given to clarify the claim or performance of the tagger. The author of this paper proposed a POS tagger for the Bangla language which follows a rule-based method. The POS tagger works on multiverb expressions too. The paper discusses a unique rule-based approach for parts of speech tagging of the Bangla language. The drawback of this paper was the lack of ways to measure the performance of the tagger proposed in this method. Including performance analysis or testing on a dataset could clarify the claim of this paper.

\par
Chowdhury \etal\cite{chowdhury2004parts} proposed a rule-based morphological analyser for Bangla parts of speech tagging. The proposed way works from the context-free to the context-bound level. The authors presented the first steps towards an automated morphological analysis of the Bangla language. In the methodology of the proposed method, a tag vector was introduced which was a sixteen-bit tag vector where parts of speech, person, mode, tense number and emotion were put in different lengths. Among the bits, three bits were kept for parts of speech. In POS there are nouns, pronouns, adjectives, verbs, and prepositions. The noun is divided into proper noun and dictionary word whereas the adjective is divided into a proper adjective and modal adjective. Some morphological rules were introduced for regular inflections, derivations and compounding with additional explicit rules for irregular inflection, derivation, and compounding. The authors introduced 21 rules for parts of speech and 4 rules for numbers. The rules were used to tag certain words. The paper provides a complexity analysis of the proposed method which was: $C(n)=N_S*n(\log_2N_r)$ where $N_S$ was the complexity of the suffix, $N_r$ was the complexity of accessing the root and $n$ was the number of characters in the patter to be tagged.\hl{Performance analysis of the system was not provided.} The POS tagger’s performance can be more understandable and clarified if the method is tested practically using any NLP tools.

\subsubsection{Machine Learning Approaches}
 
Hammad Ali\cite{ali2010unsupervised} presented the result of some initial experiment in developing an unsupervised POS tagger for the Bangla language using the Baum-welch algorithm and HMM. The whole methodology can be summarized in four phases: collection of corpus and dataset, search for implementation of Baum-Welch algorithm, performing training and final test against the gold standard for accuracy. The tagset was developed for 54 tags. The corpora were collected from a leading Bangladeshi newspaper called Prothom-Alo totals to about 50,000 tokens and the tagged subset consists of 18,110 token and 4,760 word types. After that, the Baum-Welch algorithm was implemented on the corpora using C++ HMM and NLTK toolkit. The result was stated by the author that the HMM was unable to figure out the Markov parameters from the corpora for the Bangla language in both C++ HMM and NLTK python toolkit. The paper tried to experiment on Bangla POS taggers using an unsupervised approach. The initial experiment was unsuccessful. This paper was an initial presentation of some results on supervised POS tagger of Bangla language. The author experimented with conventional unsupervised methods and figured out that the method doesn’t work in the Bangla language. This paper doesn’t provide any solution but gives some idea to the future work to be done on this topic.

\par
In 2014, Ismail \etal\cite{ismail2014developing} proposed a semi-supervised method to develop an automated Bangla POS tagging dictionary. The dictionary is developed of nouns, verbs, and adjectives. It was evaluated with a paragraph containing 10,000 manually tagged words with 11 tags. To make this automated POS tagger, the authors first created a list of suffixes that contained about 500 suffixes. Then large numbers of words were collected from various resources, which were about 1,000,000 in number. After some filtering, it came into a list of 320,443 unique words. A Hash table was used that maps the keys to values for storing data. The algorithm which was proposed works on three hash tables. The word list words were checked to examine whether it matches any root word stored in the hash map. If the word from the suffix list matches the root word, then this word is added in hashmap-1, hashmap-2, and hashmap-3. If it matches multiple times, then the longest match is considered. Then the stored words in the hash maps were analyzed to figure out the potential words suitable for POS tagging words. In this case, the occurrences of the same tag were eliminated. Finally, the POS tag dictionary was generated using hash maps using the potential candidates. The result of the experiment was evaluated from a dataset of more than 1,000,000 words from online Bangla newspapers, blogs, and other Bangla websites, and from that, 320,443 words were extracted as unique. The proposed algorithm successfully tagged 134,749 nouns, 11,067 verbs, and 8,435 adjectives. The paper shows a unique way to generate automated parts of speech tagging for the Bangla language. The POS tagger generally generates 3 POS tags (noun, verb, and adjectives). It would be better if other POS tags were added. Also, the result shows the biases on the noun tag over other POS tags. Resolving these issues would be able to improve the POS tag’s performance further.

\subsubsection{Combination of Classical and ML Approaches}

Hoque et al. \cite{hoque2015bangla} proposed an automated POS tagging system for the Bangla language based on word suffixes. The proposed method contains a custom stemming technique that is capable of retrieving different forms of suffixes. Also, a vocabulary of 45,000 words with default tags and a pattern-based verb dataset was used. The methodology applies suffix analysis through which the authors developed their desired Bangla POS tagger. Using the suffix-based morphological analysis, a stemmer was developed, which converted the suffix word into a root word. Also, the authors developed a set of rules based on suffixes, rules of Bangla grammar, and some real-time observations. After applying the stemming and the rules, the result was obtained. The dataset used for analysis was obtained from the daily prothom-alo newspaper. The dataset contained around 8,155 words and 585 sentences. The experiment was done in three different contexts: dictionary and stemmer, verb dataset with dictionary and stemmer, and finally, dictionary, stemmer, verb dataset, and set of rules. The POS tagger worked on eight fundamental parts of speech tags. The first case had an accuracy of 47.1\%, the second case had an accuracy of 63.4\%, and the third case had the best accuracy, which was 93.7\%. The method proposed in this paper was very insightful and thoughtful in Bangla POS tagging. The custom stemming and rules were helpful enough to detect the tags as perfectly as possible. Concentrating on all the subcategories of each base tag with punctuation will be able to improve the tagger even more. Accuracy can be increased by providing a small dataset and combining a probabilistic method.

Table \ref{tab:pos-tagging} covers overall datasets, results, pre-processing and methods used in POS Tagging papers.

\subsection{Question Answering System} 
\hl{Question answering is the process of answering a question in natural language and extracting an answer swiftly and simply while validating the response properly. The primary goal of this system is to comprehend and interpret phrases into an internal representation so that it can provide valid responses to user questions \mbox{\cite{mccann2018natural,hirschman2001natural}}. Question answering systems can be classified based on techniques, data resources, domains, responses, question types, and evaluation criteria used to build a question answering system \mbox{\cite{gupta2012survey}}.}

Figure \ref{fig:question} shows the basic system architecture of question answering System.
\Figure[h!](topskip=0pt, botskip=0pt, midskip=0pt)[width=8.52cm, height=6.5cm]{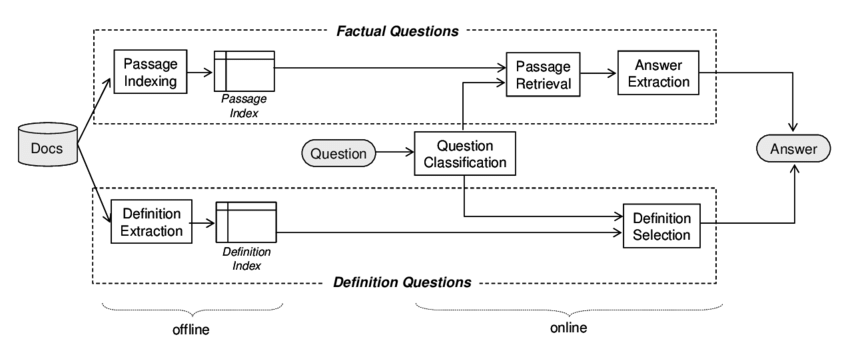}
{Basic System Architecture of Question Answering System \cite{montes2005inaoe} 
\label{fig:question}}
\subsubsection{Classical Approaches}

In 2018, Khan et al.\cite{khan2018improving} proposed a Bangla question answering system by using anaphora-cataphora resolution for simple sentences. Here, both semantic and syntactic analyses were done in this model, and a Bangla word net was built to validate the proposed system process. The authors tokenized the question into words, and the wh-type words were popped. The proposed question answering system yielded an average accuracy of 74\% for giving correct answers for the tested questions.\hl{The authors of this paper focused on using anaphora-cataphora resolution methods to improve answer extraction techniques for a Bangali question answering system. Despite the fact that their proposed system provided satisfactory accuracy, the document corpus used to test the proposed models was small, and there were not enough variations of document data because the proposed corpus lacked complex and descriptive sentences.} This paper can be improved if complex and descriptive sentence documents are collected, and there should be enough variations and contents of factoid questions and descriptive questions in the proposed corpus by enlarging the size of the document corpus.

\begin{table*}
\centering
\caption{Summary Table for Question Answering System.}
\label{tab:questiontab}
\resizebox{\linewidth}{!}{%
\begin{tabu}{>{\hspace{0pt}}p{0.056\linewidth}>{\hspace{0pt}}p{0.488\linewidth}>{\hspace{0pt}}p{0.415\linewidth}} 
\toprule
\multicolumn{1}{>{\hspace{0pt}}m{0.056\linewidth}}{} & \multicolumn{1}{>{\hspace{0pt}}m{0.488\linewidth}}{} & \multicolumn{1}{>{\hspace{0pt}}m{0.415\linewidth}}{} \\
\multicolumn{1}{>{\centering\hspace{0pt}}p{0.056\linewidth}}{\vcell{\textbf{Article}}} & \multicolumn{1}{>{\centering\hspace{0pt}}p{0.488\linewidth}}{\vcell{\textbf{Dataset and Results}}} & \multicolumn{1}{>{\centering\arraybackslash\hspace{0pt}}p{0.415\linewidth}}{\vcell{\textbf{Preprocessing Steps and Methods}}} \\[-\rowheight]
\multicolumn{1}{>{\centering\hspace{0pt}}m{0.056\linewidth}}{\printcelltop} & \multicolumn{1}{>{\centering\hspace{0pt}}m{0.488\linewidth}}{\printcelltop} & \multicolumn{1}{>{\centering\arraybackslash\hspace{0pt}}m{0.415\linewidth}}{\printcelltop} \\ 
\\
\hline\hline
\multicolumn{1}{>{\hspace{0pt}}m{0.036\linewidth}}{} & \multicolumn{1}{>{\hspace{0pt}}m{0.488\linewidth}}{} & \multicolumn{1}{>{\hspace{0pt}}m{0.415\linewidth}}{} \\
\vcell{Khan et al. \cite{khan2018improving}\newline} & \vcell{Fifty documents and ten question queries for each document. 74\% accuracy was obtained.} & \vcell{Tokenizing the question into words and popping the wh-type words. Anaphora- Cataphora resolution method was used.} \\[-\rowheight]
\printcelltop & \printcelltop & \printcelltop \\
\vcell{Islam et al. \cite{islam2016word}\newline} & \vcell{3,334 questions. Precision of 0.95562 for coarse classes and 0.87646 for more adequate classes, and after eliminating stop words, precision of 0.92421 for coarse classes and 0.8412 for finer classes were obtained.} & \vcell{Stochastic Gradient Descent was used.} \\[-\rowheight]
\printcelltop & \printcelltop & \printcelltop \\ \\
\vcell{Islam et al. \cite{islam2019design}\newline} & \vcell{Wikipedia documents and 500 questions. The precision of 0.35 recall of 0.65 and the F score of 0.45 was obtained.} & \vcell{Stop words removing and stemming the questions as well as the documents. Keyword, lexical and semantic feature extraction, and N-grams formation from these Keywords were used.\par{}} \\[-\rowheight]
\printcelltop & \printcelltop & \printcelltop \\ \\
\vcell{Uddin et al. \cite{uddin2020end}\newline} & \vcell{11,500 question-answer pairs based on history domain related questions from Bangla wiki2 corpus. Positioning encoding with LSTM yields 86\%, 97\%, and 82\% accuracy and word embedding with LSTM yields 100\%, 5\% and 5.6\% accuracy, respectively for Softmax, Linear, Relu activation function.\par{}} & \vcell{Unnecessary word or characters and URLs were removed from the raw data, web scraping and tokenization were applied. LSTM and GRU were used with positing encoding and word embedding.} \\[-\rowheight]
\printcelltop & \printcelltop & \printcelltop \\ \\
\vcell{Sarker et al. \cite{sarker2019bengali}} & \vcell{15,355 questions and 220 documents. SVM yielded 90.6\% accuracy with a linear kernel for classifying questions. 75.3\% accuracy was obtained for document categorization. 56.8\% and 66.2\% accuracy obtained without mentioning and with mentioning the object name, respectively, for extracting answers.\newline} & \vcell{Removing stop words and sign characters, tokenizing the words in exceptional cases, checking and making corrections of spelling mistakes of raw data. SGD, Decision Tree, SVM, Naive Bayes, FastText, CNN, Vector Space Model, and Edit distance comparison techniques were used.} \\[-\rowheight]
\printcelltop & \printcelltop & \printcelltop \\
\vcell{Kowsher et al. \cite{kowsher2019bangla}} & \vcell{3,127 questions and 74 topics for training and 2,852 questions from the relevant 74 topics for testing. 93.22\% accuracy using 
Cosine Similarity, 82.64\% accuracy using Jaccard Similarity, and 91.34\% accuracy using the Naive Bayes classifier.} & \vcell{Anaphora, cleaning unique characters and punctuations, stop words removing, verb processing, lemmatization, and synonyms word processing. Cosine similarity, Jaccard similarity, and Naive Bayes algorithm and SVM methods were used.} \\[-\rowheight]
\printcelltop & \printcelltop & \printcelltop \\
\bottomrule
\end{tabu}
}
\end{table*}

\subsubsection{Machine Learning Approaches}
Islam et al.\cite{islam2016word} proposed a machine learning approach to classify the Bangla question-answer types using stochastic gradient descent. The authors used two-layer taxonomy \cite{li2002learning}, which has six coarse classes \cite{gupta2018mmqa}: abbreviation, entity, description, human, location, numeric, and 50 finer classes \cite{aouichat2018arabic}. The authors achieved an average precision of 0.95562 for coarse classes and 0.87646 for more advanced classes, and after eliminating stop words, they achieved an average precision of 0.92421 for coarse classes and 0.8412 for finer classes. \hl{The authors focused on the classification of word or phrase-based answer types in Bangla question answering systems in this paper. Though the authors achieved satisfactory results in both coarse and more refined classes, the corpus size used to test the proposed models was small, and the number of descriptive questions in the proposed corpus was small, so there was not enough variation in the data.} This paper can be improved if more questions of all types are collected, and there should be enough variations and contents of factoid questions and descriptive questions in the proposed corpus.\par

In 2019, Islam et al.\cite{islam2019design} proposed an automatic question answering system for Bangla language from single and multiple documents. The proposed system identifies the question type, and for measurement of time and quantity-related questions, it provides relevant specific answers; otherwise, it retrieves relevant answers based on the questions. The authors removed stop words and performed stemming on the questions as well as the selected documents. Keywords, lexical and semantic features are extracted, and N-grams formation was used from these keywords, lexical and semantic features. Here, an average precision of 0.35, average recall of 0.65, and an F score of 0.45 were obtained.\hl{This paper used multiple documents to create a question answering system. The authors' proposed testing corpus is small and has fewer variations of questions and relevant documents; additionally, the implementation techniques of keyword extraction and n-gram formation from keywords for approximate matching for extracting questions from users' questions are not briefly defined.} This paper can be improved if more questions will be collected and more topics will be selected, and there should be enough variations and contents of factoid questions and descriptive and complex questions in the proposed corpus.\par
Uddin et al.\cite{uddin2020end} proposed a Bangla paraphrased question answering model using a single supporting line. The authors developed question-answer pairs as well as supporting lines from the Bangla wiki2 corpus. The authors gathered about 11,500 question-answer pairs based on history domain-related questions in Bangla language. Firstly, unnecessary words or characters and URLs were removed from the raw data, then web scraping and tokenization were also applied to the modified data. In the proposed system, LSTM and GRU were used with positioning encoding and word embedding. By using positioning encoding with LSTM, the authors got 86\%, 97\%, and 82\% accuracy for softmax \cite{kamath2020selective}, linear \cite{zheng2020predicting}, and relu \cite{hanin2019universal} and by using word2vec word embedding \cite{al2021synoextractor} with LSTM, they got 100\%, 5\% and 5.6\% accuracy, respectively for softmax, linear, and relu activation functions. Again, using positioning encoding with GRU, the authors got 96.59\% and 88.6\%  accuracy, respectively, for softmax and relu, and by using word2vec word embedding with GRU, they got 98.86\% accuracy for softmax activation function. \hl{The accuracy of the authors' proposed system was quite good. However, the document corpus used to test the proposed models was small, and the data did not have enough variation.} This paper can be improved by enlarging the dataset and considering an open domain-based dataset, induction, and deduction methods.\par

\subsubsection{Combination of Classical and ML Approaches}

Sarker et al. \cite{sarker2019bengali} proposed a factoid question answering system on the closed domain \cite{derici2018closed} of Shahjalal University of Science \& Technology (SUST) for helping the admission tests candidates. For question and document categorization, the authors selected five coarse-grained classes. In the preprocessing steps, the author removed stop words, removed sign characters, tokenized the words in exceptional cases, checked and made corrections of spelling mistakes of raw data manually, and rechecked the assigned labels. The authors used stochastic gradient descent, decision tree, support vector machine, and naive Bayes for classifying the questions. For document classification, they used FastText \cite{perez2020unsupervised} as an embedding technique and a convolutional neural network classifier, and also for extracting answers, vector space model and edit distance \cite{kafle2018dvqa} comparison techniques were used. For question classification, support vector machine with linear kernel provided the best accuracy of 90.6\%. For document categorization, 75.3\% accuracy was obtained. Answer extraction technique provided around 56.8\% accuracy without declaring the object name and around 66.2\% with indicating the object name. A document hit of around 72\% was received from the proposed system. \hl{In this paper, a generic factoid question answering system for Bangla was implemented. For classifying questions and extracting answers, the authors used a variety of techniques. The proposed question-answering system, on the other hand, can only respond to factoids about SUST.} This paper can be improved by enlarging this domain and selecting other domains.\par
In 2019, Kowsher et al.\cite{kowsher2019bangla} proposed an informative question answering system in Bangla. The authors selected Noakhali Science and Technology University as the only domain of this proposed chatbot model. The authors used different preprocessing techniques such as anaphora, removal of unique characters and punctuation, removing of stop words, processing verbs, lemmatization, and processing synonyms words. Jaccard similarity, cosine similarity, and naive Bayes algorithm were used to generate answers to users' questions, and support vector machine was also used to reduce space complexity and execution time. Here, 93.22\% accuracy was obtained for cosine similarity, 82.64\%  accuracy was obtained for Jaccard similarity, and 91.34\% accuracy was obtained for the naive Bayes classifier.\hl{This paper used a Bengali intelligence bot to retrieve information. Even though the authors' proposed system provided acceptable accuracy, there was insufficient variation in the questions and relevant documents because the proposed model was developed solely for the NSTU domain.} This paper can be improved by selecting and enlarging other domains and tested them in the proposed system.\par
Table \ref{tab:questiontab} shows the short description of the articles of Bangla Question Answering systems.

\subsection{Sentiment Analysis}
\hl{The sentiment is the state of the human mind. We can determine sentiment through the reactiveness of the person. Humans express their feelings through texts which enables us to figure out the state of mind. Sentiment Analysis is the part of Natural Langue Processing that determines the sentiment through analysing text data.}
\par
Figure \ref{fig:sentiment_1} shows the basic system architecture of Sentiment analysis.\\
\Figure[h!](topskip=0pt, botskip=0pt, midskip=0pt)[width=8.52cm, height=7.7cm]{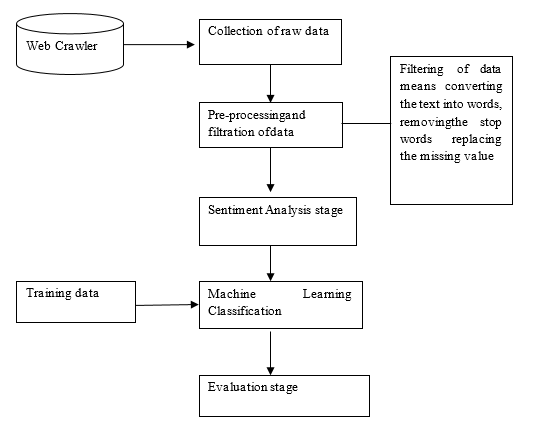}
{Basic System Architecture of Sentiment Analysis \cite{nlpweb2}.
\label{fig:sentiment_1}}

\subsubsection{Machine Learning Approaches}

Chowdhury et al.\cite{chowdhury2014performing} introduced a paper aimed to automatically extract the sentiments and opinions conveyed by users from Bangla microblog posts and identify the overall polarity of texts as either positive or negative. The approach was taken using semi-supervised bootstrapping for training corpus. The classification task was performed using SVM\cite{wang2005support} and \hl{MaxEnt}\cite{phillips2005brief} by experimenting with a combination of various sets of features. Figure \ref{fig:sentiment_2} shows the overall architecture of the system. The dataset contained 1,300 (train: 1,000, test: 300) Bangla tweets downloaded by querying Twitter API, then the tweets were pre-processed using tokenization, normalization, and parts of speech tagging. The pre-processed data were then processed using a semi-supervised bootstrapping method. Self-training bootstrapping works by first labelling a small dataset, then a classifier is trained on that small labelled data, and afterwards, the trained classifier is applied on a set of unlabelled data. The classifier is then retrained on this newly labelled data, and the process was repeated for several iterations. After that SVM and Maxent were used to classify the tweets as positive or negative sentiments. The result was obtained on the \hl{ F- measure}\cite{musicant2003optimizing}. On SVM a score of 0.93 was achieved on both labels using a combination of unigram and emoticons as features. Using only unigrams only or other features the scores were about 0.65-0.71 which increases using emoticons. On the same feature, MaxEnt gives a score of 0.85.\hl{The accuracy of the system was satisfactory. However, the neutral sentiment was absent.} Adding neutral sentiment to the classification can improve the paper.
\Figure[h!][width=8.52cm,height=6.5cm]{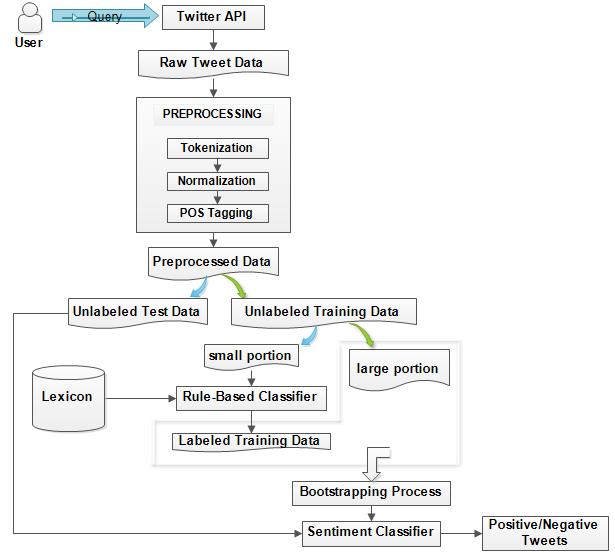}
{System Architecture of Bangla Sentiment Analysis using Microblog Posts \cite{chowdhury2014performing}.
\label{fig:sentiment_2}}

\par
In 2015, Ghosal \etal\cite{ghosal2015sentiment} performed sentiment analysis on Bangla horoscope data collected from daily newspapers using various classification models and evaluated the best model. The sentiment polarity had two classes (positive and negative). The authors’ proposed methodology part starts with collecting a horoscope dataset. The horoscope data was collected from the Bangla newspaper over one year. Then a dataset of 6,000 sentences was created and the preprocessing steps consisted of the removal of punctuation and annotation. Human annotators did annotations and for feature experimentation, unigram, bigram, and trigram features were used. A stop word dataset of 58 words was created from this dataset. Then the data was tested on five classifiers: naive Bayes, KNN, decision tree, and random forest. Then the performance was compared using 10-fold cross-validation. From the results, the best performance obtained from SVM using unigram features having an accuracy of 98.7\%. \hl{The proposed method was compared with different classifiers which show good performance analysis. However, the paper contained a classification of only two sentiment classes.} Addition of more classes can improve the paper.

\par
Hassan \etal \cite{hassan2016sentiment} proposed an approach for sentiment analysis on Bangla and  romanized Bangla \cite{alam2017sentiment} text using recurrent neural network. The emoticons and proper nouns were removed from the dataset, and the data were validated manually by two native speakers as positive, negative, and ambiguous categories. The authors used data for one validation set as pre-training for another validation set. The embedding layer \cite{li2019exploiting}, LSTM layer, and a fully connected layer \cite{abid2019sentiment} with various activations were used to detect positive, negative, and ambiguous categories of Bangla and romanized Bangla sentiments. Authors achieved 70\% accuracy for positive and negative emotions for only Bangla text and 55\% accuracy for positive, negative, and ambiguous \cite{liu2019fuzzy} sentiments for both Bangla and romanized Bangla text. \hl{The authors focused on sentimental analysis on both Bangla and Romanized Bangla Text in this paper. Though they were able to detect positive and negative sentiments with pretty satisfactory accuracy for only Bangla textual, the accuracy for romanized Bangla sentiment detection was poor.} Collection of more data for romanized Bangla sentiments and GRU implementation could give a better result for validating the model.\par

 Islam \etal\cite{islam2016supervised} used the naive Bayes model with a supervised classification method on Bangla Facebook status. The supervised method consists of Bi-gram and linguistic analysis. Their sentiment classes were positive and negative. The authors used Facebook comments for their data domain. They collected Facebook comment data manually and tagged them either positive or negative. They collected 1,000 positive and 1,000 negative comments for training and 500 comments for testing. Their pre-processing steps contained the removal of hashtags, website URLs, stemming, etc. After that, they performed the negation handling on valence shifting words. The normalization process was done using linguistic analysis. It was done according to Bangla grammar. A corpus of 1,200 unique words was used for this purpose. After that, Unigram and Bigram were used as features in the naive Bayes classification model. Laplace smoothing was also used. Then the polarity of the Facebook data was calculated by prior probability and conditional probability. After that naive Bayes classification model was used to obtain the result. The result shows naive Bayes with Unigram had precision, recall, F-score of 0.65, 0.56, and 0.60, respectively. On the other hand, naive Bayes with Bigram had precision, recall, and F-score of 0.77, 0.68, and 0.72, respectively. As people’s social media comments contain sentiment information, this polarization could be beneficial to train a model for sentiment analysis.\hl{The system has some drawbacks due to having only two sentiment classes.} The two-class output in this model is insufficient for some cases. Increasing the classes would bring more fruitful results.

\par
In 2017, Sarkar et al. \cite{sarkar2017sentiment} described their implementation of sentiment polarity detection in Bangla Tweets. They used the Bangla tweet dataset released for SAIL content 2015. Their methodology is based on four steps. (1) data cleaning and preprocessing (2) feature extraction (3) model development (4) classification. The algorithms they used are naive Bayes multinomial classifier and \hl{ SMO (a version of SVM)} \cite{flake2002efficient}. They tested it with various combinations of the N-grams.
The authors claimed that their accuracy on multinomial naive Bayes \cite{kibriya2004multinomial} combined with Unigram, Bigram and SentiWordNet was 44.20\%, and SVM with Unigram and SentiWordNet was 45\%. They claimed that SMO was not giving better results than SMO with the polynomial kernel. The authors observed that SVM classifiers trained with unigram and SentiWordNet features perform the best on the Bangla tweet dataset. The problems authors faced was lack of dataset and incorrect annotation labelled by human annotators. They implied that more training data with proper annotation would help them to develop a more accurate sentiment analysis model. The authors also claimed that their proposed system can be easily extended to other Indian languages like Hindi, Tamil, etc. \hl{The method proposed had good performance analysis. However, the proposed method had low accuracy and also had a limited dataset.}Overcoming the lack of dataset and resolving the human annotation issue can improve the paper. 
\begin{table*}
\centering
\caption{Summary Table for Sentiment Analysis.}
\label{tab:sentiment-analysis}
\begin{tabular}{>{\hspace{0pt}}p{0.123\linewidth}>{\hspace{0pt}}p{0.43\linewidth}>{\hspace{0pt}}p{0.382\linewidth}} 
\toprule
\\
\multicolumn{1}{>{\centering\hspace{0pt}}m{0.123\linewidth}}{\textbf{Article}} & \multicolumn{1}{>{\centering\hspace{0pt}}m{0.43\linewidth}}{\textbf{Datasets and Results}} & \multicolumn{1}{>{\centering\arraybackslash\hspace{0pt}}m{0.382\linewidth}}{\textbf{Preprocessing Steps and Methods}} \\ 
\\
\hline\hline
\vcell{Hassan \etal \cite{hassan2016sentiment}} & \vcell{The dataset contains~ 6,698 entries of Bangla and 2,639 entries of romanized Bangla texts that are collected from social media and other internet resources. Authors achieved 70\% accuracy for positive and negative sentiments for only Bangla text and 55\% accuracy for positive, negative and ambiguous sentiments for both Bangla and romanized Bangla text.} & \vcell{Preprocessing steps: Removal of emoticons, removal of proper nouns, and manual validation (by native speakers). Methods: LSTM and neural network.} \\[-\rowheight]
\\
\printcelltop & \printcelltop & \printcelltop \\
\vcell{Chowdhury et al. \cite{chowdhury2014performing}} & \vcell{The dataset contains 1,300 Twitter posts collected using Twitter API (Train:1000, Test:300). Based on f-score: SVM: 0.93, MaxEnt: 0.85.} & \vcell{Pre-processing steps: Tokenization, normalization, POS tagging, and semi-supervised bootstrapping. Methods: SVM and MaxEnt.} \\[-\rowheight]
\\
\printcelltop & \printcelltop & \printcelltop \\
\vcell{Tripto and  Ali \cite{tripto2018detecting}} & \vcell{The dataset contains YouTube comments collected using YouTube API 3.0 and dataset was divided into 10\% test data and from rest of the data 80\% training data, 20\% validation data were created. The highest result obtained from LSTM: Accuracy: 0.6596, F1 score: 0.63532. The highest result obtained from CNN: Accuracy: 0.6089, F1 score: 0.6052.} & \vcell{Pre-processing steps: Tokenization, normalization, stop words removal, removal of URLs, user tags, and mentions. Methods: for the proposed method: CNN, LSTM and for the baseline method: SVM, naive Bayes.} \\[-\rowheight]
\\
\printcelltop & \printcelltop & \printcelltop \\
\vcell{Islam \etal \cite{islam2016supervised}} & \vcell{The dataset contains Facebook comments (2,500 comments) manually collected. Naive Bayes with Unigram, F-score: 0.65 and Naive Bayes with Bigram, F-score: 0.77.} & \vcell{Pre-processing steps: removal of symbols, URLs, stemming, negation handling, normalization. Method: Naive Bayes.} \\[-\rowheight]
\\
\printcelltop & \printcelltop & \printcelltop \\
\vcell{Mahtab \etal \cite{mahtab2018sentiment}} & \vcell{ABSA Bangla dataset was used. Accuracy: 73.490\% using SVM on ABSA dataset.} & \vcell{Pre-processing steps: Tokenization, stop words removal, vectorization. Methods: for the proposed model: SVM and for comparison: Naive Bayes, Decision Tree.} \\[-\rowheight]
\\
\printcelltop & \printcelltop & \printcelltop \\
\vcell{Kamal Sarkar \cite{sarkar2019sentiment}} & \vcell{SAIL dataset of Bangla tweets (1,500 data) was used. Result obtained from CNN: Accuracy: 46.80\%. Result obtained from DBN: Accuracy: 43\%.} & \vcell{Pre-processing steps: Unusual sign removal, SentiWordNet polarity detection, vectorization. Methods: for the proposed method: CNN and for the baseline method: DBN.} \\[-\rowheight]
\\
\printcelltop & \printcelltop & \printcelltop \\
\vcell{Mandal \etal \cite{mandal2018preparing}} & \vcell{Raw data were collected using Twitter API. Language tagging: accuracy: 81\%. Sentiment tagging: accuracy: 80.97\%, F-score: 81.2\%.} & \vcell{Pre-processing steps: Removal of incomplete tweets, spams and contradicting sentiment tweets, filtering to preserve code-mixed property, data annotation. Methods: A hybrid model combining SGDC and rule-based method.} \\[-\rowheight]
\\
\printcelltop & \printcelltop & \printcelltop \\
\vcell{Sarkar and Bhowmick \cite{sarkar2017sentiment}} & \vcell{Bangla tweet dataset released for SAIL content 2015 was used. Based on accuracy: multinomial naive Bayes: 44.20\% and SVM: 45\%.~} & \vcell{Pre-processing steps: Removing irrelevant characters, obtain a vector representation. Methods: multinomial naive Bayes and SVM.} \\[-\rowheight]
\\
\printcelltop & \printcelltop & \printcelltop \\
\vcell{Ghosal \etal \cite{ghosal2015sentiment}} & \vcell{Bangla horoscope data from the daily newspaper over a year was used. The highest result obtained from SVM: accuracy of 98.7\%.} & \vcell{Pre-processing steps: removing punctuation, annotation. Methods: naive Bayes, SVM, KNN, Decision Tree, Random Forest.} \\[-\rowheight]
\\
\printcelltop & \printcelltop & \printcelltop \\
\bottomrule
\end{tabular}
\end{table*}

\par
 Mahtab \etal\cite{mahtab2018sentiment} implemented sentiment analysis on Bangladesh cricket Bangla text dataset using TF-IDF vectorization and SVM as classifiers. The dataset was polarized into three emotions: praise, criticism, and sadness. The dataset used here was the ABSA Bangla dataset containing 2,979 data and manually collected data containing 1,601 text samples. The manual dataset was collected from the Prothom Alo newspaper. They labelled the opinions on the dataset as praise, criticism, and sadness. There were 513, 604, and 484 labelled data on praise,  criticism, and sadness, respectively. The authors used python NLTK for tokenizing. After that, Bangla stopwords were taken into account. The feature extraction was done using the Bag of words model of the Scikit-learn library. Then TF-IDF was used to vectorize as the authors claim to be one of the most powerful ways to vectorize at that time. The processed data was used on the SVM to get the result and decision tree, multinomial naive Bayes classifiers were used to compare the results. The authors used 10\% of the data for testing and the rest for training. The result obtained shows an accuracy of 73.49\% using SVM on the ABSA dataset. On the other hand, naive Bayes and decision tree had an accuracy of 70.47\% and 64.765\%. \hl{The system can have some drawbacks due to the narrow scope of the dataset.} The authors expressed that the result was better in the ABSA dataset rather than their manually created one. Improving the size and domain of the dataset and the inclusion of more classes can improve the paper. There is another work by Awal \etal \cite{awal2018detecting} where naive Bayes is used to detect abusive comments from social networks.

Tripto and Ali\cite{tripto2018detecting} implemented multilabel sentiment and emotion analysis on Bangla YouTube comments using CNN\cite{lecun1995convolutional} and LSTM \cite{hochreiter1997long}. The labels were as follows: three classes (positive, neutral, and negative), five classes (strongly positive, positive, neutral, negative, and strongly negative), and six emotions (anger, disgust, fear, joy, sadness, and surprise). The dataset was collected from YouTube comments using YouTube API 3.0. Those are collected from videos dating from 2013 to early 2018. Google translation was used to detect the language of each comment. Then the comments were labelled by human annotators. The amount of comments are as follows: three-class (positive: 3,104, neutral: 2,805, negative: 3,001), five-class (strongly positive: 416, positive: 843, neutral: 1,222, negative: 1,064, strongly negative: 341), emotions (anger/disgust: 823, joy: 762, sadness: 272, fear/surprise: 294, none: 739). The collected data were then pre-processed using tokenization, stop words removal, removing URLs, user tags, and mentions. The lemmatization was not applied to preserve sentiment information. After that, the pre-processed data was fed into two separate models of LSTM and CNN. The classification was then compared to baseline SVM and naive Bayes\cite{rish2001empirical} methods. The authors claimed that the model outperformed the baseline models. The results obtained on three-class, five-class and emotions was as follows: LSTM (accuracy: 0.6596, 0.542, and 0.5923. F1 score: 0.63532, 0.5320, and 0.5290), CNN (accuracy: 0.6089, 0.521, and 0.5403. F1 score: 0.6052, 0.52086, and 0.53465). The baseline models scored around 0.44-0.60 on accuracy and 0.46-0.59 on F1 score. The model achieved at least 10\% more accuracy than the baseline and the existing approaches. This paper had unique characteristics of including extra sentiment classes and emotions, which was uncommon compared to other related works on Bangla sentiment analysis.\hl{However, there was some imbalance in the amount of data of various classes in the dataset.} The accuracy could be improved if a bigger dataset was provided.

\par
Kamal Sarkar\cite{sarkar2019sentiment} analysed sentiments on a Bangla Tweet dataset using deep convolutional neural networks. The classes containing sentiment analysis were three (positive, negative, and neutral). The proposed method was then compared with a deep belief neural network\cite{hinton2006fast} for evaluation. The methodology proposed by this paper was divided into four sections: pre-processing, data representation, training CNN and saving model for future prediction. In the pre-processing step, the unusual symbol removal and use of SentiWordNet to detect polarity were done. For data representation, the tweet corpus was turned into word vectors. A vocabulary of 8,483 was created. After that, the CNN model was trained which was composed of one convolutional layer, one hidden layer, and one output layer of softmax activation. For training and testing, the SAIL tweets dataset was used. A rectified linear unit was used in a convolutional and dense layer. After the training and process were completed, the proposed CNN-based model had an accuracy of 46.80\%, where the DBN-based model had an accuracy of 43\%. The author expressed concern about scarcity of benchmark datasets for sentiment analysis in Indian languages. Also, they claimed that the dataset was insufficient and noisy. The author had a plan to perform exploitation of unlabelled data in the training process of DBN and CNN based models. A proposal of using a recursive neural network was also brought. \hl{Though the methodology was simple the accuracy was poor in this proposed system.} A bigger dataset can improve the overall performance of the model. Increasing layers in the model can also be considered.

\subsubsection{Combination of Classical and ML Approaches}

Mandal \etal\cite{mandal2018preparing} collected raw twitter data, created  Bengali-English code-mixed corpus for sentiment analysis, and applied various classification methods to measure the best performance. The data processing work was done manually, and they followed very complicated steps for providing a corpus, which the authors claimed to be the gold standard. The polarity of sentiment analysis had three polarities: positive, negative, and neutral. In methodology, in the beginning, the authors collected raw twitter data using Twitter API. After that, the data was filtered and maintained the code-mixed property. In the clearing process, spam, incomplete tweets, ones with contradicting sentiment were removed. Url and hashtags were preserved as they used to possess some sentiment information. By this, out of 89,000 data, 5,000 data were taken. The annotation had two phases, language tagging, and sentiment tagging. The annotated data were reviewed by human annotators. Language tagging process contained LBM (Lexicon based module) and SLM (supervised learning module). Sentiment tagging was generally based on rule-based and supervised methods. The training data consisted of 1,500 training instances and 600 testing instances that were equally distributed in both processes. Then naive Bayes and linear models were used to evaluate the performance. Among them, the performance of SGDC had the best performance with an F-score of 78.70\%. After that, the SGDC was used on a hybrid classifier built with rule-based and supervised methods. The final result can be divided into two sections, language tagging, and sentiment tagging. The language tagging had a performance of 81\% accuracy, and the sentiment tagging had an accuracy of 80.97\% and an F1-score of 81.2\%. \hl{The performance of the proposed system provided satisfactory results. But the classifier had only 3 sentiment classes and the latest methods were not applied or tested in the system.}Bigger dataset and the latest classification models can improve the overall performance of the model. Also, sentiment classes provided in this paper can be increased, providing a bigger dataset.

Table \ref{tab:sentiment-analysis} covers overall datasets, results, pre-processing and methods used in Sentiment Analysis.

\subsection{Spam and Fake Detection} 
\hl{Deceptive contents, such as fake news and phony reviews, spam emails, offensive tweets, text, or comments, sometimes known as opinion spam, have become a growing threat to online customers and users in recent years. It is feasible to undertake spam and fake detection in online news, emails, tweets, reviews, and other text data using various natural language processing approaches by extracting significant features from the text using various natural language processing techniques \mbox{\cite{crawford2015survey}}.}
\subsubsection{Machine Learning Approaches}

In 2019, Islam et al.\cite{islam2019using} proposed an approach for detecting spam from malicious Bangla text using multinomial naïve Bayes (MNB). Punctuation marks, numerical values, and emoticons were extracted, and a TF-IDF vectorizer \cite{bhutani2019fake} was used to extract the features. Here, the multinomial naive Bayes yielded an overall accuracy of 82.44\% with precision: 0.825, recall: 0.824, F-score: 0.808, and error: 17.56. Here, the authors could not provide enough data for developing the proposed model, especially since the spam data was tiny and no processed data were chosen for the validation set, and the feature extraction mechanism was not good enough. This paper can be improved by training and testing the proposed model on a more extensive dataset.
\par
\begin{table*}
\centering
\caption{Summary Table for Spam and Fake Detection.}
\label{tab:spamtab}
\resizebox{\linewidth}{!}{%
\begin{tabu}{>{\hspace{0pt}}p{0.054\linewidth}>{\hspace{0pt}}p{0.56\linewidth}>{\hspace{0pt}}p{0.334\linewidth}} 
\toprule
\multicolumn{1}{>{\hspace{0pt}}m{0.054\linewidth}}{} & \multicolumn{1}{>{\hspace{0pt}}m{0.56\linewidth}}{} & \multicolumn{1}{>{\hspace{0pt}}m{0.334\linewidth}}{} \\
\multicolumn{1}{>{\centering\hspace{0pt}}p{0.054\linewidth}}{\vcell{\textbf{Article}}} & \multicolumn{1}{>{\centering\hspace{0pt}}p{0.56\linewidth}}{\vcell{\textbf{Datasets and Results}}} & \multicolumn{1}{>{\centering\arraybackslash\hspace{0pt}}p{0.334\linewidth}}{\vcell{\textbf{Preprocessing Steps and Methods}}} \\[-\rowheight]
\multicolumn{1}{>{\centering\hspace{0pt}}m{0.044\linewidth}}{\printcelltop} & \multicolumn{1}{>{\centering\hspace{0pt}}m{0.56\linewidth}}{\printcelltop} & \multicolumn{1}{>{\centering\arraybackslash\hspace{0pt}}m{0.334\linewidth}}{\printcelltop} \\ 
\\
\hline\hline
\multicolumn{1}{>{\hspace{0pt}}m{0.044\linewidth}}{} & \multicolumn{1}{>{\hspace{0pt}}m{0.56\linewidth}}{} & \multicolumn{1}{>{\hspace{0pt}}m{0.334\linewidth}}{} \\
\vcell{Islam \etal \cite{islam2019using}} & \vcell{1,965 variational instances, including 646 spam and 1,319 ham (non-spam). 82.44\% accuracy was obtained with precision: 0.825, Recall: 0.824, F-score: 0.808, and Error: 17.56.} & \vcell{Punctuation marks, numerical values, and emoticons were removed from the raw data. For classification, multinomial naive Bayes and TF-IDF vectorizer were used. \newline} \\[-\rowheight]
\printcelltop & \printcelltop & \printcelltop \\
\vcell{Hussain \etal \cite{hussain2020detection}\newline} & \vcell{2,500 public articles. 93.32\% and 96.64\% accuracy were obtained respectively for MNB and SVM.} & \vcell{Special Characters, English digits, English alphabets, and emoticons were removed from the raw data. For classification, multinomial naive Bayes and support vector machine were used.\newline } \\[-\rowheight]
\printcelltop & \printcelltop & \printcelltop \\
\vcell{Hossain et al. \cite{hossain2020banfakenews}} & \vcell{50,000 news, and among them 8,500 news were manually annonated. Linguistic features with SVM produced a 91\% F1-score. SVM and LR had F1 scores of 46\% and 53\%, respectively. In CNN, average pooling and the global max technique had F1 scores of 59\% and 54\%, respectively. In the BERT model, F1-Score was 68\%.} & \vcell{The raw text data were normalized, and stop words and punctuation were also removed. Traditional linguistic features and neural network models were used for classification purposes.} \\[-\rowheight]
\printcelltop & \printcelltop & \printcelltop \\
\bottomrule
\end{tabu}
}
\end{table*}

In 2020, Hussain et al.\cite{hussain2020detection} proposed a technique for investigating Bangla fake news from social media sites. The authors assumed the articles from very renowned portals as actual news, and reports from satire news sites are considered fake news. Various preprocessing techniques, such as removing special Characters, removing Bangla \& English digits, removing English alphabets, and removing emoticons, were applied to the raw text data. Count vectorizer \cite{kaur2020automating} and TF-IDF vectorizer were used to extract features. For classifying the fake news, the authors used multinomial naive Bayes classifier and support vector machine classifier with a linear kernel. Here, 93.32\% accuracy was obtained for multinomial naive Bayes, and 96.64\% accuracy was gained using support vector machine. Here, the author’s assumption of classifying the fake and real news is controversial because some popular news portals occasionally publish fake news. This paper can be improved by enlarging the size of the proposed news article corpus. Also, the classification technique for fake and real information should be made unambiguous.

\subsubsection{Combination of Classical and ML Approaches}

Hossain et al.\cite{hossain2020banfakenews} proposed an annotated dataset containing  50K news for developing automated phony news detection systems in Bangla language and evaluated the dataset using NLP techniques by classifying Bangla fake news. To classify fake news, the authors used standard linguistic features including lexical features \cite{asghar2020opinion}, syntactic features \cite{choudhary2021linguistic}, semantic features \cite{jiang2019neural}, metadata \cite{rastogi2020effective} and punctuation, and neural network models including CNN and LSTM. Also, a multilingual BERT model \cite{aluru2020deep} was used to classify the news documents. The authors performed several preprocessing techniques like normalizing the text and stop words, removing punctuations from the raw data. The proposed system yielded a 91\% F1-score when using linguistic features with SVM. For news embedding, RF \cite{alom2018detecting} yielded a 55\% of F1-score and SVM, LR \cite{murugan2019feature} yielded 46\% and 53\% of F1-scores respectively. In CNN, 59\% and 54\% F1-scores were gained respectively by using average pooling \cite{kaur2019review}, global max technique \cite{masood2019spammer}. In the BERT model, F1-Score was 68\%. Here the authors manually annotated only around 8.5K news. This paper can be improved by annotating more data.\par
Table \ref{tab:spamtab} shows the short description of the articles of Bangla spam and fake detection systems.

\subsection{Text Summarization} 
\hl{When we need to understand a long text it is time-consuming to read through the whole text and also hard for us to consume the learning. Text Summarization is a section of Natural Language Processing which deals with the summarization of longer texts with computer processing.}
\subsubsection{Classical Approaches}
In 2017, Abujar \etal\cite{abujar2017heuristic} introduced a Bangls text summarization technique using the extractive method along with some heuristic rules. The technique uses sentence scoring to evaluate the key sentences to focus on summarization. In the proposed method, the process starts from pre-processing, which consists of tokenization, stop words removal, and stemming. The prime sentences are then identified by word analysis and sentence analysis based on length, distance, values, etc. Finally, the prime sentences are evaluated using sentence scoring methods. The final process includes aggregate similarities, final gist analysis, and sentence ranking. The result obtained by the authors was counted out of five. The result was compared with human-generated summaries. The proposed method scored around 4.3, where the human-generated result scored 4.6. The authors expressed that the abstractive method gave a better summary than the extractive one, but it requires many development phases. The technique proposed was able to perform the text summarization in the Bangla language successfully. \hl{No standard way of evaluation was followed in the proposed system. The performance can be compared to a standard approach to get further clarification.}

\par
In 2013, Efat \etal \cite{efat2013automated} suggested a text summarization technique on Bangla text document. Their model is extraction-based and summarizes a single document at a time. The authors have proposed a classical method for summarizing the model, and they used newspaper articles for their model. 
The authors collected the data from the newspaper from Bangladesh, then they have used some preprocessing techniques on those Bangla textual documents like tokenization, stop word removal, and stemming. Then they ranked the sentences of the document. They counted on the frequency, position value, cue words, and skeleton of the documents for this purpose. Frequency is the appearance of a word, and the position of sentence influences the summary of a document. Also, cue words or connecting words play essential roles. Skeleton consists of the document header, and title. Based on these features, a summary marking was allocated to each sentence. The authors collected 45 articles from different newspapers, and those articles are stored in UTF-8 file format. They gave a sentence marking by tuning the parameters of their model. Based on the sentence rank, the summary for the article was generated. The authors claimed that their model achieved an F1 score of 83.57\%. They evaluated the model’s performance by comparing the summary that was generated by their proposed method and human summaries. \hl{However, this study was limited to a single themed document. This paper could be improved by using statistical models like CRFs and machine learning algorithms like SVM and LSTM to create a model that can generate a summary of any single document.} 
\begin{table*}
\centering
\caption{Summary Table for Text Summarization.}
\label{tab:text-summarization-fuad}
\begin{tabular}{>{\hspace{0pt}}p{0.086\linewidth}>{\hspace{0pt}}p{0.559\linewidth}>{\hspace{0pt}}p{0.29\linewidth}} 
\hline
\\
\multicolumn{1}{>{\centering\hspace{0pt}}m{0.086\linewidth}}{\textbf{Article}} & \multicolumn{1}{>{\centering\hspace{0pt}}m{0.559\linewidth}}{\textbf{Datasets and Results}} & \multicolumn{1}{>{\centering\arraybackslash\hspace{0pt}}m{0.29\linewidth}}{\textbf{Preprocessing Steps and Methods}} \\ 
\\
\hline\hline
\vcell{Akter \etal \cite{akter2017extractive}} & \vcell{Satisfactory summary generation consisting 30\% of the merged sentences.} & \vcell{Pre-processing steps: Noise removal, tokenization, stop-word removal, stemming, sentence scoring. Methods applied: K-Means clustering.} \\[-\rowheight]
\\
\printcelltop & \printcelltop & \printcelltop \\
\vcell{Abujar \etal \cite{abujar2017heuristic}} & \vcell{Scored 4.3 out of 5 where the human score was 4.6.} & \vcell{Pre-processing steps: Tokenization, stop words removal, stemming. Methods applied: Heuristic approach.} \\[-\rowheight]
\\
\printcelltop & \printcelltop & \printcelltop \\
\vcell{Rayan \etal. \cite{rayan2021unsupervised}} & \vcell{139 samples of human-written abstractive summary from NCTB books and extractive dataset BNLPC were used. The Rouge-1, Rogue-2 and Rouge-L scores for abstractive summary on NCTB dataset: 12.17, 1.92, and 11.35 respectively. The same scores of extractive summary on BNLPC were 61.62, 55.97, and 61.09 respectively. The scores from the human evaluation were 4.41, 3.95, and 4.2 for evaluating the content, readability, and quality.} & \vcell{Pre-processing steps: Tokenization, removal of stopwords, POS tagging, filtering of punctuation marks. Methods applied: Sentence clustering using ULMFiT pre-trained model and word graph using POS tagging.}\\[-\rowheight]\\
 \printcelltop & \printcelltop & \printcelltop \\
\vcell{Das \etal \cite{das2010topic}} & \vcell{Data collected from AnandaBazar newspaper for creating the corpus. The F1 score obtained by theme detection was 79.85\% and theme clustering was 69.65\%.} & \vcell{Manually the data were annotated and stored in XML file. For theme detection, CRFs model and for theme clustering, K-cluster algorithm were used. \newline} \\[-\rowheight]\\
 \printcelltop & \printcelltop & \printcelltop \\
\vcell{Efat \etal \cite{efat2013automated}} & \vcell{Data collected from the newspaper for creating the corpus and The F1 score obtained by the model was 83.57\%.} & \vcell{Tokenization, stop words removal and stemming were used and for the summarization, a sentence ranking system was created.} 
\\[-\rowheight]
\printcelltop & \printcelltop & \printcelltop \\
\bottomrule
\end{tabular}
\end{table*}

\subsubsection{Machine Learning Approaches}

Das \etal \cite{das2010topic} suggested an opinion based summarization of Bangla text in one document. Topic sentiment recognition and aggregation are done through the topic-sentiment model and theme clustering. These also create the document Level theme relational graph from which the summary is generated by Information Retrieval (IR). The authors first created a corpus for their model as Bangla has no established corpus yet. They collected the data from the AnandaBazar newspaper. Then they annotated the collected data and stored the annotated data in an XML file for future use. They chose three main criteria for the feature set creation: lexico-syntactic, syntactic, and discourse-level features. In lexico- syntactic, there was POS tagger, SentiWordNet, frequency, and stemming. Under the syntactic features, there were chunked levels and dependency parsers. Finally, under the discourse level feature, there was the document's title, first paragraph, term distribution, and collocation. For the term distribution, the TF-IDF method was used. The authors then chose to use a conditional random field model to detect the document's theme. For the theme clustering, they used a clustering algorithm. Then they presented the document and their finding into a graph. Then based on the IR score, the summary was generated. The authors achieved accuracy for the theme detection technique is 83.60\% (precision), 76.44\% (recall), and 79.85\% (F-measure). The summarization system achieved the precision of 72.15\%, recall of 67.32\%, and an F1 score of 69.65\%. The authors presented an opinion based on summarization techniques. They were looking forward to generating a hierarchical cluster of theme words with time-frame relations in the future, which may be useful in the further progress of Bangla NLP. \hl{However, there was less data, and the annotators were not trained linguists. This paper can be improved by adding more data and developing linguistic tools to make the model more robust.}
\par

In 2017, Akter \etal\cite{akter2017extractive} proposed a text summarization technique using the K-means clustering algorithm in the extractive method. In the extractive method, sentence scoring was used in this paper to improve summarization. The proposed method can work on single and multiple Bangla documents. The proposed method introduced by the authors is the extractive based text summarization method using the sentence clustering approach. The pre-processing steps included noise removal, tokenization, stop word removal, stemming. The sentence scoring method was used for extraction-based methods. TF-IDF was used to find the word scoring. After that, sentence scoring was obtained by cue words or skeleton words and sorted in decreasing order. This same process was performed for multi documents. After that, based on scoring, the K-means clustering was performed. From that, top K sentences were extracted from each cluster, and the summary was generated from 30\% sentences of the original merged document. The proposed method produced expected output in linear complexity. In the final summary from single or, multiple documents, 30\% of the merged documents were represented. \hl{The authors provided a well-made approach however, the drawback of the proposed method was summarized sentences were not synchronized. Also the system performance was not given using any ideal standard of comparison.} The authors implied that this technique had an improved performance based on linear time complexity. The performance can be further improved by enhancing the clustering technique.  
\begin{figure}
    \centering
    \includegraphics[width=\linewidth,height=6.5cm]{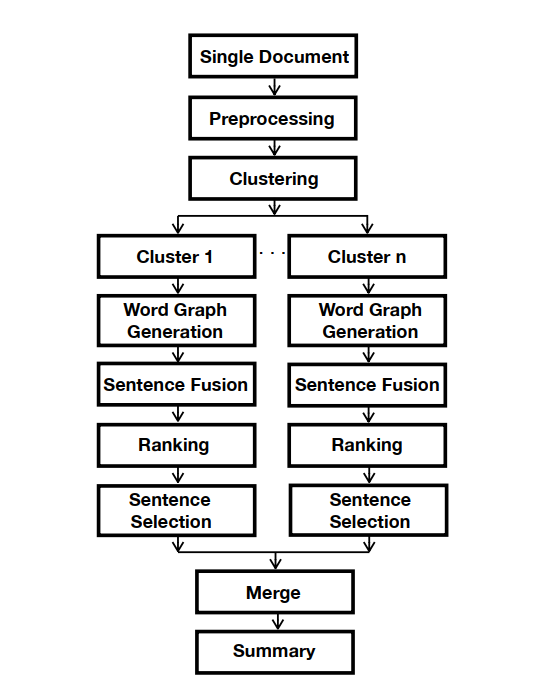}
    \caption{Proposed BenSumm Model for Bangla Abstractive Text Summarization\cite{rayan2021unsupervised}.}
    \label{fig:bensumm}
\end{figure}
\subsubsection{Combination of Classical and ML Approaches}

In 2021, Rayan \etal\cite{rayan2021unsupervised} proposed a graph-based unsupervised abstractive method of summarization. The proposed method required only a POS tagger and a language model which was pre-trained. At first, the sentences were preprocessed by tokenization, removal of stopwords, parts of speech tagging and filtering of punctuation marks. Natural language toolkit was used for preprocessing tasks. After that, the sentences were clustered to group similar sentences. For clustering, ULMFiT pre-trained language model was used as TF-IDF did’t work well. The clusters were minimum 2 and maximum $n-1$. The silhouette score was also used to determine the similarity between the sentences. Next, a word graph was created to obtain abstractive fusions from the sentences which are related. The top-ranked sentences were used to represent the summary. The merging of the sentences provided the summary.
\par 
Figure \ref{fig:bensumm} shows the overall architecture of the model. The proposed model was termed as BenSumm model. The evaluations were done using Rogue scores and human scores. The Rouge-1, Rogue-2, and Rouge-L scores for the abstractive summary on the NCTB dataset were 12.17, 1.92 and 11.35 respectively and for the extractive summary on BNLPC were 61.62, 55.97 and 61.09 respectively. From human evaluation, the scores were 4.41, 3.95, and 4.2 for evaluating content, readability, and quality. The overall paper provided an unsupervised approach for text summarization that outperformed the previous baseline methods. One of the drawbacks was that the given model cannot generate new words. Overcoming this problem will make this model more robust. 

\par

Table \ref{tab:text-summarization-fuad} covers the overall datasets, results, pre-processing steps and methods used in text summarization papers.

\subsection{Word Sense Disambiguation}
\hl{In natural language, the majority of regularly used words have numerous meanings. Word Sense Disambiguation is the process of automatically assigning preset meanings to words in definite contexts and reducing the problem of ambiguous words in natural language. The four primary approaches for constructing a word sense disambiguation system are supervised, unsupervised, semi-supervised, and knowledge-based methods \mbox{\cite{iacobacci2016embeddings}}.}

\subsubsection{Classical Approaches}

Kaysar et al.\cite{kaysar2018word} proposed a technique to determine the meaning of a Bangla word with multiple meanings using the apriori algorithm \cite{kaysar2019word}. For testing purposes, a new sentence with an ambiguous term was tokenized. The tokens were compared with the learned knowledge to determine the intended meaning of the test sentence’s ambiguous word. The authors separated the rows for each sense and applied the apriori algorithm to each set separately to solve the ambiguity problem. A database was formed with nine data samples compared with different consensus in different length of words, and word samples were converted to binary flags. The authors found out frequent itemset using the apriori algorithm, and then association rules were generated. Then, the database attributes were converted into binary flags, and if word sets were present in the sample, then one otherwise zero was generated. For 1,600 sentence patterns, the differences between apriori and grammar-based approaches \cite{gonzales2017improving} were about 26\%. The apriori method had a 46\% prediction rate and the processing time of apriori was 1,604 nanosecond faster than grammar-based processing time.\hl{In this paper, the authors mainly focused on extracting the meaning of  Bengali words with multiple meanings using the apriori algorithm. However, the proposed dataset size for testing the model is small, and the implementation steps of using the apriori algorithm and association rules are not defined briefly in this paper.} This paper can be improved by enlarging the proposed sentence corpus’s size with more ambiguous words and developing a combined dictionary-based system with machine learning.

\par
Alam \etal \cite{alam2008text} suggested a model for text normalization for Bangla. The process is mainly constructed with tokenization, token classification, token sense disambiguation and word representation. The authors defined some semiotic classes for their work. They used two steps procedures to achieve their semiotic classes for Bangla. They used a python script in the newspaper corpus, utilized the corpus for their model and set up some rules to mark the semiotic classes for the Bangla language. Then the authors have specified a particular tag to each of the semiotic classes, and then the corpus undergoes tokenization. There were three levels in the tokenization such as tokenizer, spiller, and classifier. For each type of token, regular expressions was used. Then by using a token expander, the authors verbalized the tokens and disambiguated the ambiguous tokens. The authors achieved a performance of 99\% for the ambiguous tokens such as float, time and currency. Their rule-based model successfully determined the flotation point and currency but lacked much in determining the time where performance was 62\%. The authors used newspaper, blog, and forums data for their model. This model did not deal with the misspelling problem, which can be achieved by introducing a separate module.
\par

\begin{table*}
\centering
\caption{Summary Table for Word Sense Disambiguation.}
\label{tab:wordsense}
\resizebox{\linewidth}{!}{%
\begin{tabu}{>{\hspace{0pt}}p{0.056\linewidth}>{\hspace{0pt}}p{0.488\linewidth}>{\hspace{0pt}}p{0.415\linewidth}} 
\toprule
\multicolumn{1}{>{\hspace{0pt}}m{0.056\linewidth}}{} & \multicolumn{1}{>{\hspace{0pt}}m{0.488\linewidth}}{} & \multicolumn{1}{>{\hspace{0pt}}m{0.415\linewidth}}{} \\
\multicolumn{1}{>{\centering\hspace{0pt}}p{0.056\linewidth}}{\vcell{\textbf{Article}}} & \multicolumn{1}{>{\centering\hspace{0pt}}p{0.488\linewidth}}{\vcell{\textbf{Dataset and Results}}} & \multicolumn{1}{>{\centering\arraybackslash\hspace{0pt}}p{0.415\linewidth}}{\vcell{\textbf{Preprocessing Steps and Methods}}} \\[-\rowheight]
\multicolumn{1}{>{\centering\hspace{0pt}}m{0.056\linewidth}}{\printcelltop} & \multicolumn{1}{>{\centering\hspace{0pt}}m{0.488\linewidth}}{\printcelltop} & \multicolumn{1}{>{\centering\arraybackslash\hspace{0pt}}m{0.415\linewidth}}{\printcelltop} \\ 
\\
\hline\hline
\multicolumn{1}{>{\hspace{0pt}}m{0.036\linewidth}}{} & \multicolumn{1}{>{\hspace{0pt}}m{0.488\linewidth}}{} & \multicolumn{1}{>{\hspace{0pt}}m{0.415\linewidth}}{} \\
\vcell{Kaysar et al.\cite{kaysar2018word}}&
   \vcell{100 to 1,600 sentence patterns for testing. The apriori approach has 46\% prediction rate and 1,604 nanosecond faster processing time than grammar-based method.\newline} &
  \vcell{Tokenizing the ambiguous words and apriori algorithm was used.}\\[-\rowheight]
\printcelltop & \printcelltop & \printcelltop \\  
  \vcell{Alam \etal \cite{alam2008text}} &
  \vcell{Self created datasets from Prothom Alo newspaper, amader projukti forum, and blog and overall accuracy obtained 71\%. \newline} &
  \vcell{Use of Semiotic classes for Bangla language and rule based model were used.} \\[-\rowheight]
\printcelltop & \printcelltop & \printcelltop \\
  \vcell{Haque  \etal \cite{haque2016bangla}} &
  \vcell{500 bangla sentence and overall accuracy obtained 82.40\%. \newline \newline}&
  \vcell{Used a parse tree and a bangla dictionary to detect the ambiguity.\newline} \\[-\rowheight]
\printcelltop & \printcelltop & \printcelltop \\
  \vcell{Pal  \etal \cite{pal2018word}}&
  \vcell{The datasets was used in the models are Bangla POS tagged corpus from Indian Languages corpora initiative and Bangla WordNet developed by Indian Statistical Institute and the average precision value obtained was 84\%. \newline}&
  \vcell{Text normalization, removal of non-functional words, selection of ambiguous words and naive Bayes method.} \\[-\rowheight]
\printcelltop & \printcelltop & \printcelltop \\
  \vcell{Pandit \etal \cite{pandit2015memory}} &
  \vcell{Self created dataset and overall accuracy obtained 71\%. \newline} &
  \vcell{Use of POS tagger, removal of stop words and KNN based algorithms. \newline} \\[-\rowheight]
\printcelltop & \printcelltop & \printcelltop \\
  \vcell{Nazah  \etal \cite{nazah2017word}} &
  \vcell{Self created datasets and overall accuracy obtained 77.5\%. \newline} &
  \vcell{Use of POS tagger, removal of stop words and naive Bayes probabilistic model and GLNN.} \\[-\rowheight]
\printcelltop & \printcelltop & \printcelltop \\
\bottomrule
\end{tabu}
}
\end{table*}

\par
In 2016, Haque  \etal \cite{haque2016bangla} proposed a word sense disambiguation system where they used two primary steps and removed the ambiguity in sentences. First, They took a Bangla sentence as their input which was tokenized at the beginning. They used a Bangla dictionary for storing Bangla words and the corresponding meanings, which was connected to parsers. This dictionary was used for the verification of the parse tree. Tokenized words are then passed to a parser to create a parse tree. The authors created CSG rules for making the Bangla parser. The parser mainly took tokens and checked their lexicon for validity, and if a valid token was detected, then the tokens were compared to the CSG rules to generate the parse tree. From the parse tree, the ambiguity of the word was determined with the help of the Bangla dictionary. The authors used 500 sentences to understand the disambiguation, and among them, they managed to divide 412 sentences correctly. Thus, the overall accuracy of the system was around 82.40\%. The authors also shown concerns about the token they generated as there was no machine-readable dictionaries to validate the tokens. Moreover, the model did not consider the semantic meaning of a sentence. The model also was not tested with a larger sample of the sentence to evaluate the performance. 

\subsubsection{Machine Learning Approaches}

Pandit \etal \cite{pandit2015memory}  suggested a KNN-based algorithm to resolve the word sense disambiguation problem. To implement a KNN-based algorithm, they removed all the stop words. Then they used a Bengali stemmer on the entire dataset. To resolve the POS ambiguity, they used the Stanford POS tagger. They assigned a weight between 0 to 1 to each neighbour selected by the overlap metric for the test example. This measure was used to compare the test sets with the training set. The training example which was nearest to the observation set was given the highest weight. When selecting the value of k, they avoided using even numbers as it might arise the situation of tied votes. The authors introduced a majority voting to determine the classes of unknown test examples. In the question of a tie in majority voting, the test cases were determined based on a priority to select the first class. The experiment was done with a small amount of dataset, which the authors created. The dataset only contains 25 target words to be disambiguated. They used 250 self-annotated sentences for their training set and evaluated the model with 100 sentences containing those target words. The model had an overall accuracy of 71\% and the model was not consistent in detecting various types of parts of speech. Their model gained an accuracy of 76.2\% in detecting the nouns, but for determining the adverb, their model achieved an accuracy of only 53.8\%. The possible reason for this might be the selection of the target words where there were large amount of imbalance between the numbers of adverbs and nouns. There also a question remains how well the dataset was annotated to prepare for the training and testing. Overall, this has been a sound approach to sense the disambiguation for the words. The approach can be further enhanced by introducing more training and testing examples with standard labelling on them. 
\par

In 2017, Nazah  \etal \cite{nazah2017word} created a dataset for developing a word sense disambiguation system for Bangla sentence and got overall accuracy of 77.5\%. The authors used POS tagger, removed stop words and applied naive Bayes probabilistic model and GLNN.
\par

In 2018, Pal \etal \cite{pal2018word} suggested a word sense disambiguation system for Bangla, which uses a naive Bayes probabilistic model for their baseline strategy. The authors also showed two extensions of their naive Bayes probabilistic model. The authors used the Bangla POS tagged corpus of the Indian language corpora Initiative and the Bangla WordNet developed by the Indian Statistical Institute. The authors normalized the text by removing unnecessary spaces, punctuations and delimiters. They converted the whole text into a unicode format. Then they removed the non-functional words from the sentences. Then they tried to mark the unambiguous word. They generated reference output files with the help of a Bangla dictionary. Then they applied a baseline method on 900 sentences. These sentences consist of 19 ambiguous words mostly used in Bangla. The authors established an algorithm to sense ambiguous words. Their model achieved the precision and recall value of 81\% on average for 19 mostly used Bangla ambiguous words. The authors have suggested two more extensions of their baseline model by adopting lemmatization and bootstrapping. The use of lemmatization achieved an improved average precision and recall value of 84\%, and with bootstrapping, they have achieved 83\% of average precision and recall value. 
\par
Table \ref{tab:wordsense} shows the short description of the articles of Bangla word sense disambiguation system.

\subsection{Speech Processing and Recognition} 
\hl{Speech recognition is the process of converting a natural speech signal into a sequence of words using computer algorithms. Isolated, connected, continuous, and spontaneous speech recognition systems are the different types of speech recognition systems \mbox{\cite{gaikwad2010review}}. Signal processing and feature extraction, acoustic model, language model, and hypothesis search are the main components of an automatic speech recognition system. Traditional acoustic models, deep neural networks, DNN-HMM hybrid systems, representation learning in deep neural networks, and advanced deep models can all be used to create speech recognition systems \mbox{\cite{yu2016automatic}}.}
 
\begin{figure}[H]
\centering
    \includegraphics[width=8.7cm, height=4.5cm]{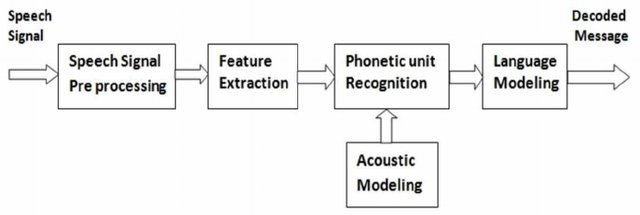}
     \caption  {Basic system architecture of Speech Processing and Recognition System \cite{sarma2015acoustic}.
     \label{fig:speech}
     }
\end{figure}

Figure \ref{fig:speech} shows the basic system architecture Speech Processing and Recognition system.

\subsubsection{Classical Approaches}

Sultana et al.\cite{sultana2012bangla} proposed an approach for Speech-to-Text conversion using Speech Application Programming Interface (SAPI) \cite{chung2020end} for the Bangla language. The authors managed SAPI to combine pronunciation from the spoken continuous Bangla speech with a precompiled grammar file, and then SAPI returned Bangla words in English character if matches occur. For recognizing the Bangla accent, an XML grammar file for SAPI was generated with English character combinations for each Bangla word. \hl{The main goal of this paper is of converting Bangla speech to text data. Though the overall recognition rate for repeated and different words was relatively good, SAPI is slow, and its sequential operation is challenging to work.} Some extra tools were needed to perform this experiment, including Microsoft Visual Studio 2010, Speech Application Programming Interface or SAPI 5.4, Microsoft SQL Server 2008, and Avro software.

The average recognition rate was 78.54\% for repeated words, and the average recognition rate was 74.81\% for different names. However, SAPI is slow, and its sequential operation is problematic, and again an XML grammar was required to generate manually for testing data. This paper can be improved by using the parallel process of SAPI or a faster speech processing engine. \par

Ali et al.\cite{ali2013automatic} proposed a technique for recognizing spoken words in Bangla where Mel-frequency cepstral coefficients (MFCC), LPC, and GMM were used for feature extraction. Then, template matching and DTW were used for matching the speech signal. Preprocessing steps including analog to digital conversion, pre-emphasis filter, hamming window, fast Fourier transform were performed to the raw speech data. MFCC and DTW yielded a recognition rate of 78\%. LPC and DTW achieved a recognition rate of 60\% and using MFCC, GMM, and posterior probability functions \cite{miyoshi2017voice} produced a recognition rate of 84\%. Also, MFCC, LPC, and DTW gained a recognition rate of 50\%. \hl{The principal purpose of this research is of establishing an automatic speech recognition Technique for Bangla isolated voice. Though the authors used numerous strategies for executing their suggested work, they did not supply any test data information for detecting the Bangla voice signal and dealt with isolated Bangla speech signals only.} This paper can be improved by enlarging the corpus size and dealing with connected speech data.  \par
In 2014, Hasnat et al.\cite{hasnat2007isolated} proposed a technique for creating an isolated and continuous Bangla speech recognition system using the Hidden Markov Modeling Toolkit (HTK) \cite{gales2008application}. For recognizing isolated speech word-based HMM model, a simple dictionary containing only the input-output HMM model name were created, and for labeling the dictionary words, the HSLab tool \cite{zarrouk2014hybrid} was used. For recognizing continuous speech, phoneme-based HMM model, language model and pronunciation dictionary were created containing the input-output pronunciation for each word entry. The authors eliminated the detected noise and end points from the raw speech data. The proposed isolated speech recognition model had 70\% speaker-dependent accuracy and 90\% speaker-independent accuracy. The proposed continuous speech recognition model had 80\% speaker-dependent accuracy and 60\% speaker-independent accuracy. \hl{Though the proposed system works well on isolated speech data, the amount of training and testing data were small, with very few variations.} This paper can be improved by testing the model on a bigger dataset. \par

Aadit et al.\cite{aadit2016pitch} proposed a technique for finding the Bangla alphabet's oral characteristics in terms of pitch and formant. The authors downsampled the speech signal and read small segments of 40 milliseconds in MATLAB. Autocorrelation was used for pitch estimation in time domain, and Cepstrum were used for pitch estimation in frequency domain, and LPC was used for formant estimation \cite{welling1998formant}. The authors worked with both Bangla vowels and consonants speech. The authors claimed that similar pitch frequencies were found for the same vowels and consonants in both time and frequency domain, and pitch and formant frequencies of vowels and consonants were higher for female speakers than male speakers. Again, they found that the difference between formant frequency in male and female voice increased from the first formant to the third formant. \hl{The principal purpose of this research is of reporting pitch and formant estimation of Bangla vowels and consonants. The authors only dealt with noise-free speech signals, and only a few tests were done for estimating the formants and pitch of speech signals.} This paper can be improved by testing the model on variational speech data and performing more experiments. \par

Ahmed et al.\cite{ahmed2018implementation} proposed a voice input speech output calculator that can recognize isolated and continuous speech in Bangla language and derive mathematical expressions. Hidden Markov models in CMU Sphinx performed the speech-to-text conversion \cite{nasib2018real}, while the result was converted to speech using Android Text-to-Speech (TTS) API. The acoustic model's training process requires a Bangla phonetic dictionary, a language model, and Bangla acoustic data. The Google Translate application was used to generate the phonetic transcription of Bangla words. The proposed Bangla speech recognizer model achieved 86.7\% word recognition accuracy. \hl{Even though the authors gathered a pretty good accuracy, the amount of voice data utilized to develop the model was fairly little, with fewer variances.} This paper can be improved if more speech data of Bangla speech can be collected from different age groups and various locations' speakers.\par

In 2021, Paul et al.\cite{paul2021bengali} proposed a Bangla numeral recognition system from the speech signal using MFCC and GMM. A dataset containing 1,000 audio samples for ten classes (0-9) was created for the experiment. MFCC features were extracted from the audio files that the authors collected. The authors developed a GMM based Bengali isolated spoken numerals recognition system for their proposed task. The authors claimed that 91.7\% correct prediction was obtained for their self-built Bangla numeral data set. \hl{Despite the fact that the authors achieved a high level of accuracy using their self-created Bangla numeral data set, the speech data of numerals used to train the model had limited variations.} This paper can be improved by collecting more variational Bangla speech numeral data, developing and evaluating the dataset on upgraded and suitable deep learning models.

\subsubsection{Machine Learning Approaches}

In 2017, Ahammad et al.\cite{ahammad2016connected} proposed a connected digit recognition system based on neural network for the Bangla language. For training and testing purposes, the authors proposed two datasets of connected digits comprising fifteen male and fifteen female speakers. Connected digits were subjected to segmentation to acquire isolated words, and the proposed system used MFCC analysis to elicit meaningful features for recognition.  Backpropagation neural network was used for training purposes that used 352 neurons in the input layer, keeping it according to the number of feature parameters in the MFCC file. Again, in this proposed system, for adjusting weight, the gradient descent method had been used in BPNN. The average accuracy found for recognizing the digits in this research were: Shunno (0) - 92.31\%, EK (1) - 98.72\%, DUI (2) - 98.72\%, TIN (3) - 98.72\%, CHAR (4) - 82.05\%, PACH (5) - 76.92\%, CHOY (6) - 75.64\%, SHAT (7) - 92.30\%, AAT (8) - 93.59\%, NOY (9) - 89.74\%. \hl{The primary goal of this paper is to create an automatic recognition system for connected Bangla digits. Although the authors attained a high level of accuracy using their self-created Bangla numeral data set, they trained the model with a smaller dataset.} This paper can be improved by enlarging the dataset for training and testing purposes, employing hybrid classifiers, or incorporating robust features.
\par 
Nahid et al.\cite{nahid2017bengali} proposed a technique for creating a Bangla Speech Recognition system using a double layered LSTM-RNN approach. The authors tested the system on a dataset containing 2,000 words and split the dataset into 75:12.5:12.5 for training, validation, and testing purposes, respectively. Noise reduction and phoneme mapping were made on the raw speech data. The authors divided each word into several frames and selected the first thirteen MFCC for extracting different types of features from the raw speech data. The authors imposed random disparity in the number of frames three times for each training example. The proposed model was a deep recurrent neural network with two layers of 100 LSTM cells each. the proposed model achieved 28.7\% phon detection error rate \cite{ghannay2016acoustic} and 13.2\% word detection error rate \cite{sarma2004context}. \hl{The main drawbacks of this paper are that the dataset is tiny, and the experimental methods for labeling and threshold value selection are unclear.} This paper can be improved by enlarging the dataset for training, validating, and testing purposes and using GRU cells in the place of LSTM cells.
\begin{table*}
\centering
\caption{Summary Table for Speech Processing and Recognition System.}
\label{tab:speechtab}
\resizebox{\linewidth}{!}{%
\begin{tabu}{>{\hspace{0pt}}p{0.069\linewidth}>{\hspace{0pt}}p{0.559\linewidth}>{\hspace{0pt}}p{0.209\linewidth}>{\hspace{0pt}}p{0.104\linewidth}} 
\toprule
\multicolumn{1}{>{\hspace{0pt}}m{0.069\linewidth}}{} & \multicolumn{1}{>{\hspace{0pt}}m{0.559\linewidth}}{} & \multicolumn{1}{>{\hspace{0pt}}m{0.209\linewidth}}{} & \multicolumn{1}{>{\hspace{0pt}}m{0.104\linewidth}}{} \\
\multicolumn{1}{>{\centering\hspace{0pt}}p{0.069\linewidth}}{\vcell{\textbf{Article}}} & \multicolumn{1}{>{\centering\hspace{0pt}}p{0.559\linewidth}}{\vcell{\textbf{Dataset and Results}}} & \multicolumn{1}{>{\centering\hspace{0pt}}p{0.209\linewidth}}{\vcell{\textbf{Preprocessing Steps and Methods}}} & \multicolumn{1}{>{\centering\arraybackslash\hspace{0pt}}p{0.104\linewidth}}{\vcell{\textbf{Task}}} \\[-\rowheight]
\multicolumn{1}{>{\centering\hspace{0pt}}m{0.069\linewidth}}{\printcelltop} & \multicolumn{1}{>{\centering\hspace{0pt}}m{0.559\linewidth}}{\printcelltop} & \multicolumn{1}{>{\centering\hspace{0pt}}m{0.209\linewidth}}{\printcelltop} & \multicolumn{1}{>{\centering\arraybackslash\hspace{0pt}}m{0.104\linewidth}}{\printcelltop} \\ 
\\
\hline\hline
\multicolumn{1}{>{\hspace{0pt}}m{0.065\linewidth}}{} & \multicolumn{1}{>{\hspace{0pt}}m{0.559\linewidth}}{} & \multicolumn{1}{>{\hspace{0pt}}m{0.209\linewidth}}{} & \multicolumn{1}{>{\hspace{0pt}}m{0.104\linewidth}}{} \\
\vcell{Sultana et al. \cite{sultana2012bangla}\newline} & \vcell{396 Bangla words, and 270 of them were distinct. 78.54\% accuracy for repeated words and 74.81\% accuracy for distinct words.} & \vcell{Generating an XML grammar and speech application program interface.} & \vcell{Speech to text conversion.} \\[-\rowheight]
\printcelltop & \printcelltop & \printcelltop & \printcelltop \\ \\
\vcell{Ali et al. \cite{ali2013automatic}} & \vcell{100 words were recorded 1,000 times. MFCC and DTW achieved 78\% accuracy, LPC and DTW achieved 60\% accuracy, MFCC and GMM yield 84\% accuracy, and MFCC, LPC, and DTW achieved 50\% accuracy.} & \vcell{Analog to digital conversion, pre-emphasis filter, hamming window, FFT, MFCC, LPC, GMM and DTW methods. \newline} & \vcell{Noisy isolated speech recognition.} \\[-\rowheight]
\printcelltop & \printcelltop & \printcelltop & \printcelltop \\ \\
\vcell{Hasnat et al. \cite{hasnat2007isolated}} & \vcell{100 words, 5 to 10 samples for each word, and 1,814 phonemes. The isolated speech recognition model yielded 70\% speaker-dependent and 90\% speaker-independent accuracy, and the continuous speech recognition model achieved 80\% speaker-dependent and 60\% speaker-independent accuracy.\newline} & \vcell{Noise elimination, end point detection and Cambridge HMM toolkit.\par{}} & \vcell{Noisy isolated and continuous speech recognition.} \\[-\rowheight]
\printcelltop & \printcelltop & \printcelltop & \printcelltop \\ \\
\vcell{Aadit et al. \cite{aadit2016pitch}\newline} & \vcell{Speech data were collected from 25 males and 25 females. Similar pitch frequencies for the same vowels and consonants were found in both time and frequency domain. Female speakers had higher pitch and formant frequencies than male speakers.} & \vcell{Downsampling the speech signal, autocorrelation cepstrum and LPC.} & \vcell{Pitch and formant estimation of noise-free speech data.} \\[-\rowheight]
\printcelltop & \printcelltop & \printcelltop & \printcelltop \\ \\
\vcell{Ahmed et al. \cite{ahmed2018implementation}\newline} & \vcell{500 sentences containing 2,733 Bangla words. 86.7\% word recognition accuracy.} & \vcell{CMU Sphinx and Android Text-to-Speech API.} & \vcell{Noisy numeral speech recognition.} \\[-\rowheight]
\printcelltop & \printcelltop & \printcelltop & \printcelltop \\ \\
\vcell{Paul et al. \cite{paul2021bengali}\newline} & \vcell{1,000 audio samples for (0-9) classes. 91.7\% accuracy.} & \vcell{MFCC and GMM.} & \vcell{Numeral speech recognition.} \\[-\rowheight]
\printcelltop & \printcelltop & \printcelltop & \printcelltop \\ \\
\vcell{Ahammad et al. \cite{ahammad2016connected}\newline} & \vcell{Two datasets of connected speech in Bangla. Accuracy for recognizing: Shunno (0) - 92.31\%, EK (1) - 98.72\%, DUI (2) - 98.72\%, TIN (3) - 98.72\%, CHAR (4) - 82.05\%, PACH (5) - 76.92\%, CHOY (6) - 75.64\%, SHAT (7) - 92.30\%, AAT (8) - 93.59\%, NOY (9) - 89.74\%.\par{}} & \vcell{Segmenting the raw speech data and Backpropagation neural network.} & \vcell{Noise free numeral speech recognition.} \\[-\rowheight]
\printcelltop & \printcelltop & \printcelltop & \printcelltop \\ \\
\vcell{Nahid et al. \cite{nahid2017bengali}\newline} & \vcell{Two thousand words and 114 words of them are unique. 28.7\% phon detection error rate and 13.2\% word detection error rate.} & \vcell{Noise reduction, phoneme mapping and LSTM.} & \vcell{Noisy real numbers speech recognition.} \\[-\rowheight]
\printcelltop & \printcelltop & \printcelltop & \printcelltop \\ \\
\vcell{Sumit et al. \cite{sumit2018noise}\newline} & \vcell{350 and 111 hours of Bangla speech. 9.15\% and 34.83\% character error rate on Socian and Babel noisy speech, respectively and 12.31\% and 10.65\% character error rate on Socian and CRBLP read speech, respectively.} & \vcell{Aligning and segmenting the audio clips, multi-layer neural network, CNN, RNN, GRU and FC layers.} & \vcell{Augmented noisy speech recognition.} \\[-\rowheight]
\printcelltop & \printcelltop & \printcelltop & \printcelltop \\ \\
\vcell{Sumon et al. \cite{sumon2018bangla}\newline} & \vcell{Ten classes of data and 100 samples per class. MFCC model yielded an accuracy of 74\%, the proposed raw model yielded 71\% accuracy, and the proposed transfer model achieves 73\% accuracy.} & \vcell{CNN and Transfer learning.} & \vcell{Short speech command recognition.} \\[-\rowheight]
\printcelltop & \printcelltop & \printcelltop & \printcelltop \\ \\
\vcell{Islam et al. \cite{islam2019speech}\newline} & \vcell{200,000 wav audio files. 86.058\% accuracy in the CNN model.} & \vcell{Sampling the raw speech data, CNN and RNN.} & \vcell{General public speech recognition.} \\[-\rowheight]
\printcelltop & \printcelltop & \printcelltop & \printcelltop \\ \\
\vcell{Shuvo et al. \cite{shuvo2019bangla}\newline} & \vcell{6,000 utterances. 93.65\% test set recognition accuracy.} & \vcell{CNN} & \vcell{Noise-free numeral speech recognition.} \\[-\rowheight]
\printcelltop & \printcelltop & \printcelltop & \printcelltop \\ \\
\vcell{Sharmina et al. \cite{sharmin2020bengali}\newline} & \vcell{1,230 audio files. 98.37\% accuracy was found with 98\% precision, 98\% recall, and 98\% F1-score.} & \vcell{CNN} & \vcell{Numeral speech recognition.} \\[-\rowheight]
\printcelltop & \printcelltop & \printcelltop & \printcelltop \\ 
\vcell{Ovishake et al. \cite{sen2021convolutional}\newline} & \vcell{4,000 audio files. 97.1\% accuracy in test set} & \vcell{CNN} & \vcell{Numeral speech recognition.} \\[-\rowheight]
\printcelltop & \printcelltop & \printcelltop & \printcelltop \\ 
\vcell{Paul et al. \cite{paul2009bangla}\newline} & \vcell{60 training examples for each digit. High accuracy of recognizing individual digits.} & \vcell{Pre-emphasis filtering and speech coding, LPC and ANN.} & \vcell{General public speech recognition.} \\[-\rowheight]
\printcelltop & \printcelltop & \printcelltop & \printcelltop \\ \\
\vcell{Sultan et al. \cite{sultan2021adrisya}} & \vcell{Bangla human voice commands.} & \vcell{Computers, peripheral devices of the computer and home appliances were used.} & \vcell{Virtual assistant for visually impaired people.} \\[-\rowheight]
\printcelltop & \printcelltop & \printcelltop & \printcelltop \\
\bottomrule
\end{tabu}
}
\end{table*}

\par 
In 2018, Sumit et al.\cite{sumit2018noise} proposed an approach that aims at developing an end-to-end deep learning method leveraging current signs of progress in the automatic speech recognition system to recognize continuous Bangla speech for noisy environments. Here the authors used data augmentation and deep learning model architecture to improve the robustness of their proposed model. The authors used the Kaldi toolkit \cite{povey2011kaldi} for aligning and segmenting tasks of the raw clips in a maximum of ten seconds long with their corresponding transcripts. Here, a multi-layer neural network was used, and the authors evaluated the model architecture up to nine layers, including convolutional neural networks, gated recurrent units, and fully connected layers. The authors used a private dataset, Babel, in both the training and testing phase and CRBLP speech corpus \cite{murtoza2011phonetically} in the test phase only. The authors used connectionist temporal classification \cite{salazar2019self} for aligning input and output sequence and ignoring blanks and repeated characters without penalty. The proposed model yielded a 9.15\% character error rate on Socian noisy speech and 34.83\%  character error rate \cite{starner1994line} on Babel noisy speech. The proposed model also achieved 12.31\% and 10.65\% character error rate on Socian and CRBLP reading speech, respectively. \hl{The authors mainly concentrated on developing an automatic speech recognition system to recognize continuous Bangla speech for noisy environments. They tested their proposed model on a moderately large dataset, but the feature extraction and preprocessing steps are not defined clearly in this paper.} This paper can be improved by developing an additional language model to improve word or character level performance.
\par 
Sumon et al.\cite{sumon2018bangla} proposed three CNN architectures to recognize Bangla short speech commands. The authors proposed three CNN model approaches, including MFCC based CNN model, raw CNN model, and pre-trained CNN model using transfer learning for performing the proposed task. The authors extracted MFCC features from the audio files, and after normalizing them, the authors fed the features to convolutional neural network architecture. The MFCC model had one convolutional layer with five filters, and each of the filters has a stride of 1. The raw model had a similar architecture with one convolutional layer and a softmax layer. In the pre-trained model, MFCC features were extracted, and the proposed model had three convolutional layers, which were associated with max-pooling and batch normalization layers. The proposed MFCC model yielded accuracy of 74\%, the proposed raw model yielded 71\% accuracy, and the proposed transfer model gave 73\% accuracy. The authors described all types of model performance briefly with diagrams and performance tables. \hl{The authors mainly concentrated on building a speech recognition system to recognize Bangla short speech commands. The pre-trained English speech dataset comprises 65,000 samples; however, their self-made dataset has just 10,000 samples with a slight variation.} This paper can be improved by collecting more variational speech data of Bangla short speech commands.
\par 

In 2019, Islam et al.\cite{islam2019speech} proposed a speech recognition system in the Bengali language using CNN and developed RNN based method to find the Bengali character level probabilities. The probabilities were improved by using the CTC loss function \cite{lee2021intermediate} and a language model. For building the architecture of the CNN system, the authors extracted the features from the raw speech data by using mel frequency cepstral coefficients. The authors implemented a five hidden layer-based neural network where the first layer was the input layer, and the last layer was the output layer. For training the proposed CNN model, the authors took 30 samples of each word to train the proposed CNN model, and for testing purposes, they took a vocabulary of 100 words. The recognition system with feature extraction and spectrum had an average accuracy rate of 86.058\%. To build the RNN system's architecture, the authors used a dataset of 33,000 Bangla audio files spoken by 500 speakers. Here, the input files were fed into three fully connected layers, followed by a bidirectional RNN layer \cite{graves2013speech}, and finally, another fully connected layer. The output consisted of the softmax function that resulted from character probabilities for each character in the alphabet. The authors claimed that the model could not memorize the outcomes as they started discarding more parameters using dropout. \hl{The authors obtained a significant accuracy in the CNN model however there needed a powerful tool to train the RNN model with thousands of data.} This paper can be improved by collecting and training the proposed CNN and RNN model with more data.
\par
Shuvo et al.\cite{shuvo2019bangla} proposed a Bangla numeral recognition system from the speech signal utilizing CNN. A speech dataset of ten isolated Bangla digits of a total of 6,000 utterances was recorded in a noise-free environment. MFCC analysis was used to produce meaningful MFCC features using the Librosa library \cite{tjandra2017listening}, where 16 MFCC coefficients were extracted from the numeral speech signals. Convolutional neural network was used for the recognition task. The proposed CNN model's development steps for classifying the spoken Bangla digits were described briefly in this paper. The authors claimed that Bangla digits of six were misclassified with nine, and seven were misclassified with eight because they are phonetically very close to their counterparts when pronounced. The proposed system achieved an average of 93.65\% test set recognition accuracy. \hl{The authors mainly concentrated on building a Bangla numeral recognition system from the speech signal. The authors yielded a notable accuracy for recognizing Bangla numerals and gathered a pretty good variation in their self-built numeral dataset. This paper can be improved by collecting more speech data of Bangla digits.} \par

In 2020, Sharmina et al.\cite{sharmin2020bengali} proposed an in-depth learning approach for classifying the Bengali spoken digits. A dataset containing 1,230 audio files was created for the experiment from five males and five females. MFCC features were extracted from the audio files using the python Librosa library. The Python Scikit-learn library \cite{geron2019hands} was used to divide 80\% of the data into training set and the rest of the 20\% data into test set. Keras API written in python \cite{de2018neural} was used for developing a convolutional neural network for feature learning and classification purposes. The authors claimed that when 1,230 data were fed into the model, an accuracy of 98.37\% was found with 98\% precision, 98\% recall, and 98\% F1-score. The authors performed three experiments based on different amounts of the dataset and compared them showing the accuracy, precision, recall, and F1-score. They also showed the importance of gender, dialects, age groups by performing experiments on the proposed model. \hl{The authors mainly concentrated on constructing a Bangla spoken digit classification system using the voice signal. Although the authors acquired a fair accuracy for detecting Bangla spoken digits, the dataset size for evaluating the suggested models is tiny for training and testing purposes.} This paper can be improved by collecting more variational Bangla speech data of the digits. 

In 2021, Ovishake et al. \cite{sen2021convolutional} proposed a CNN based approach to recognize Bangla spoken digits from speech signal.  This research used audio recordings of Bangladeshis of diverse genders, ages, and accents to construct a massive speech collection of spoken '0-9' Bangla digits. The dataset was created by recording of 400 noisy and noise-free samples per digit. For extracting significant features from raw voice data, MFCCs were used. CNN was used to recognize Bangla numeral digits. Across the entire dataset, the proposed technique correctly distinguishes '0-9' Bangla spoken digits with 97.1\% accuracy. The model's efficiency was further evaluated using 10-fold cross-validation, which resulted in a 96.7\% accuracy. \hl{This paper's main purpose is to develop an automatic recognition system for connected Bangla digits. The authors achieved a good level of accuracy using their self-created Bangla numeral data set, but they trained the model using a smaller dataset. This paper could be improved by using hybrid classifiers and enlarging the dataset for training and testing.}

\subsubsection{Combination of Classical and ML Approaches}

Paul et al.\cite{paul2009bangla} proposed a Bangla speech recognition system using LPC and ANN. The authors divided the proposed system into two major parts. The speech signal processing part consisted of speech starting and endpoint detection, windowing, filtering, calculating the LPC and cepstral coefficients, and finally constructing the codebook by vector quantization \cite{manchanda2018hybrid}. The speech pattern recognition part consisted of recognizing patterns using an ANN. Pre-emphasis filter and speech coding were applied to the digitized speech. The recognizer was designed to identify the ten digits, and for each digit, the input to the recognizer was the feature vectors. The authors claimed that, they used four different Bangla words uttered by four different persons, and a satisfactory level of accuracy was obtained for recognizing individual digits. \hl{The authors mainly concentrated on creating a Bangla Speech Recognition System employing both classical and machine learning methodologies. The dataset size and actual accuracy were not provided in this paper.} This paper can be improved by enlarging the dataset size for training and testing purposes and by developing a suitable and upgraded version of deep learning models for the recognition task.\par

In 2021, Sultan et al.\cite{sultan2021adrisya} proposed a Bangla virtual assistant \cite{doumbouya2021using}, `Adrisya Sahayak (Invisible Helper)' for visually impaired people. `Adrisya Sahayak' worked as a desktop application specially created for Bangla-speaking visually impaired persons to reduce their difficulties in their daily activities. The authors created this virtual assistant by utilizing a computer and home appliances' peripheral devices. This virtual assistant will help visually impaired persons because the existing virtual assistants are mostly in the English language. \hl{The researchers mainly concentrated on establishing a Bangla virtual assistant for visually impaired persons. However, the proposed virtual assistant was a user-independent technology that can only do minimal computer operations in the home environment using Bangla human voice commands.} \par
Table \ref{tab:speechtab} shows the short description of the articles of Bangla speech processing and recognition systems.

\subsection{Methods Used in BNLP} 
\begin{figure}[h!]
\centering
\begin{tikzpicture}[scale = 0.8]
 
\pie[
    text = legend,
]
{44/Classical Methods, 6/CNN, 6.25/ANN, 7.5/Naive Bayes, 8.75/SVM, 
    16.25/Other ML Methods,
    11.25/Hybrid Methods
   }
 
\end{tikzpicture}
\caption{Methods used in Bangla Natural Language Processing.}
\label{fig:BNLPmethods}
\end{figure}
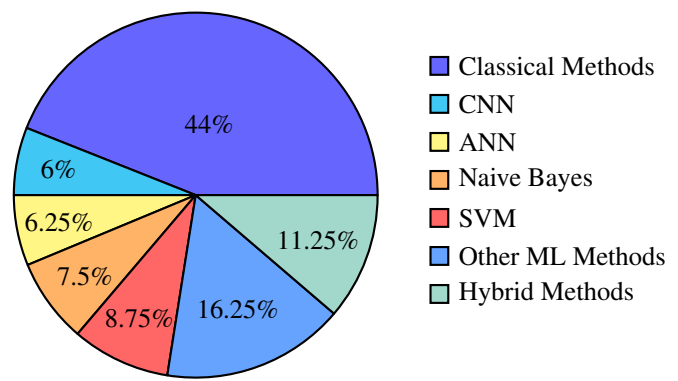

Figure \ref{fig:BNLPmethods} shows the overall summary of the methods used in BNLP.
\subsubsection{Classical} 
Table \ref{tab:classicalab} shows the short description of classical methods used in BNLP till now.

\subsubsection{ML and DL} 

Table \ref{tab:ML-BNLP} shows the short description of machine learning and deep learning methods used in Bangla natural language processing till now.
\par

\section{Current Challenges and Future Trends in BNLP}

\label{sec:challenges}
\subsection{Challenges in Preprocessing}

Bangla text data preprocessing's most common challenges are character-set dependence, corpus dependence, application dependence, writing system dependence, tokenizing of punctuations, tokenizing of space-delimiters, preprocessing of Bangla grammatically wrong words, preprocessing of ambiguous words, preprocessing of erroneous words, preprocessing multi-label Bangla data. Moreover, the automatic computerized services for Bangla text preprocessing and relevant research works are hardly available for facilitating researches on Bangla language text data  \cite{indurkhya2010handbook,haque2020approaches}. \par

The most common challenges of Bangla speech data preprocessing are: preprocessing of falsely interpreted Bangla speeches, preprocessing of very noisy or loud Bangla speeches, preprocessing of reverberations and overlapping Bangla speeches, preprocessing of Bangla speech signal that has an unnecessary delay at the end of speech frames, preprocessing of sparsely spoken speech, preprocessing ambiguous speech, preprocessing of erroneous speech data, and preprocessing of different sampling rates speeches in a corpus \cite{kuligowska2018speech,sun2018novel}. \par

\subsection{Challenges in Text Format}

The most common challenges of developing Bangla question answering systems are: extracting answers to questions that have ambiguous answers, answering tricky questions, answering procedural, temporal, and spatial questions, unavailability of datasets for creating open domain-based question answering system, extracting answers for questions that are synonyms of Bangla words, lexical gaps between the knowledge corpus and questions, difficulties in developing multiple choices questions, and lack of significant researches on Bangla question answering systems \cite{hoffner2017survey,kafle2017visual}. \par
At the time of developing Bangla spam and fake detection systems, the researchers faced many difficulties, including timeliness and oddity can change the classification of fake and authentic information, the classifying of spam and ham (non-spam) Bangla documents is a difficult task, unavailability of public datasets, as well as lack of significant researches on Bangla spam and fake detection \cite{zhou2019fake}. \par

Researchers faced some difficulties at the time of developing Bangla word sense disambiguation systems, including variation in structure and formatting of Bangla words that are used in various sentences, shortage of significant resources and knowledge for research in this area, unavailability of sufficient tools for developing such systems for Bangla language, and the existing approaches are only targeted at nouns and verbs \cite{moon2015challenges}. 

\par 

Sentiment 
analysis is a well-nourished topic of Bangla NLP. Most of the challenges in sentiment analysis come from a lack of a proper dataset. The sentiment labels have limited classes, so, working with a wide range of sentiments is hard due to the lack of dataset availability. \hl{Furthermore, }detecting ambiguous sentiment is also a big challenge in sentiment analysis. Sentiment analysis in Bangla code-mixed sentences is another challenging task that also requires a large dataset.

\par
Machine translation in Bangla is also an important topic to be working on. Idioms and phrases are one of the vital parts of a language. Idioms and phrases differ in every language based on their culture and regional environment. One of the biggest challenges of machine translation is handling idioms and phrases. \hl{Moreover, }limited corpus for creating a machine translation dictionary is also an issue. 

\par
Parts of speech has a big impact on developing other domains of Bangla NLP. Bangla language has very diverse parts of speech tags. The tags are very different than we can see in the English language. It is still a challenge to tag all the parts of speech properly as the Bangla language structure has different parts of speech compare to English. 

\par
Text summarization in Bangla is quite challenging in abstractive text summarization task. abstractive text summarization has a big challenge in summarizing long sentences.  In terms of mixed Bangla sentences, this challenge gets harder. The complexity of summarization is less in extractive methods than abstractive. \hl{Additionally, }taking care of complexity on abstractive summarization is also a challenging task. 

\par
Parsing in Bangla is quite difficult due to the structure of the language. Also, complexity is an issue.

\par
In the case of information extraction, the researchers face the limitation of proper tools for extracting the information to create a corpus for their model. Also, Bangla does not have rich linguistic tools to handle the different aspects of the information retrieval tasks. The lack of quality corpus to train a model is another significant step down in the field of Bangla NLP. Additionally, information retrieval from a Bangla document imposes a great challenge as Bangla has almost a set of 350 characters (full set of letters, letter combinations, and numbering) \cite{B_Alphabet}. Bangla also has no capitalization, which becomes a problem for the information extraction from various tasks. Also, the modern machine learning model's parameters have not been tweaked for Bangla.

\par
The first complexity the researchers face in constructing a NER system for Bangla is that it has no capitalization. Unlike English, all the words are written in capital blocks. So, there is no easy way to detect the name entity in the middle of a sentence by observing the capitalization. Also, in Bangla, an adjective can be used as a proper noun. Like a person's name could be "SUNDAR" (beautiful), which is an adjective. These ambiguities add up to the challenges of NER tasks. Also, there is a lack of a machine-readable dictionary and properly annotated corpus for Bangla. To build a NER system, researchers often create their own corpus sets to make their model work. \hl{Particularly, }these constraints make it difficult for the researchers to create a good NER system.

\par 
Parsing is one of the essential factors in the field of natural language processing. In constructing Bangla parsers, researchers face many obstacles. Bangla is a morphologically rich language, and there is still no rigorous morphological analysis of the Bangla language. Also, Bangla has no comprehensive corpus to work with. The lack of resources and complexity of Bangla grammar and structure becomes the main constraints for the Bangla language parsing. Additionally, the lack of good linguistic tools makes it harder to create a satisfactory model. So, most of the researchers do the intermediate works for Bangla parsers.

\par 
Text summarization in Bangla also suffers from the lack of good linguistic tools available in Bangla. Also, the scarcity of well-built corpus has an impact, especially for summarization, a lot of linguistic tools are required to make a good model. To make a summary sentiment of articles plays a vital role in various tasks and Bangla still does not have a gold standard sentiment analysis model. \hl{Additionally, }most of the works have been done using statistical models, and the evaluation of these models is also problematic as there is no concrete metric for evaluating the results. 

\subsection{Challenges in Speech Format}

The development of a speech processing and recognition system that can efficiently work in natural, freestyle, loud, and noisy environments and preparing a plan that effectively works on all possible practical grounds are some of the most crucial challenges \cite{deng2004challenges}.
The most commonly faced challenges of Bangla speech processing and recognition are: imprecision and false interpretations of Bangla speech words, accents and local differences of Bangla speech data, a vast number of non-native Bangla speakers and their variations on speaking attitudes, code-switching, and code-mixing incidences \cite{ayeomoni2006code}, developed methods have inefficiency in the processing of noisy and loud Bangla speech data, unique phonological systems of speeches \cite{boets2011preschool}, word segmentation problems, fuzzy grammatical structures, lack of publicly available Bangla speech corpus and lack of significant researches on Bangla speech data. Additionally, building a recognition system that can effectively process and identify all types of Bangla speech data is a very complicated task. Consequently, the accuracy remains inadequate for recognizing the Bangla speeches efficiently \cite{besacier2014automatic,sumon2018bangla}.
\begin{table*}
\centering
\caption{Classical Methods used in BNLP.}
\label{tab:classicalab}
\label{tab:CL-BNLP}
\resizebox{\linewidth}{!}{%
\begin{tabu}{>{\hspace{0pt}}p{0.238\linewidth}>{\hspace{0pt}}p{0.703\linewidth}}
\hline
\toprule

\multicolumn{1}{>{\hspace{0pt}}m{0.238\linewidth}}{} & \multicolumn{1}{>{\hspace{0pt}}m{0.703\linewidth}}{} \\
\multicolumn{1}{>{\centering\hspace{0pt}}p{0.238\linewidth}}{\vcell{\textbf{Classical Methods}}} & \multicolumn{1}{>{\centering\arraybackslash\hspace{0pt}}p{0.703\linewidth}}{\vcell{\textbf{Use Cases \hl{and Best Results}}}} \\[-\rowheight]
\multicolumn{1}{>{\centering\hspace{0pt}}m{0.238\linewidth}}{\printcelltop} & \multicolumn{1}{>{\centering\arraybackslash\hspace{0pt}}m{0.703\linewidth}}{\printcelltop} \\ 
\\
\hline\hline
\multicolumn{1}{>{\hspace{0pt}}m{0.238\linewidth}}{} & \multicolumn{1}{>{\hspace{0pt}}m{0.703\linewidth}}{} \\
\vcell{Rule based Methods} & \vcell{Information Extraction (Chandra  \etal \cite{chandra2013hunting}, \hl{Accuracy:95.96\%}), Machine Translation (Chowdhury et al. \cite{chowdhury2013developing} and Anwar \etal\cite{anwar2009syntax}\hl{, Accuracy:93.33\%}), Named Entity Recognition (Chaudhuri  \etal \cite{chaudhuri2008experiment}, \hl{Precision:94.24\%}) and Parts of Speech Tagging (Hoque et al. \cite{hoque2015bangla}\hl{, Accuracy:93.7\%}), Sentiment Analysis (Mandal \etal\cite{mandal2018preparing}\hl{, Accuracy:81\%} and Chakrabarti et al. \cite{chakrabarti2011layered}), and Word Sense Disambiguation (Alam \etal \cite{alam2008text}, \hl{Accuracy:71\%}). \newline} \\[-\rowheight]
\printcelltop & \printcelltop \\
\vcell{Verb based Methods\par{}} & \vcell{Machine Translation (Rabbani \etal\cite{rabbani2014new}).\newline} \\[-\rowheight]
\printcelltop & \printcelltop \\
\vcell{CFG\par{}} & \vcell{Parsing (Mehedy \etal \cite{mehedy2003bangla}).\newline} \\[-\rowheight]
\printcelltop & \printcelltop \\
\vcell{Language Independent\par{} Statistical Models.\newline} & \vcell{Information Extraction (Chandra \etal \cite{chandra2013hunting}, \hl{Accuracy:95.96\%}).\newline} \\[-\rowheight]
\printcelltop & \printcelltop \\
\vcell{Morphological Parsing Techniques\par{}} & \vcell{Parsing (Dasgupta \etal\cite{dasgupta2005morphological}\hl{, Accuracy:100\%}).\newline}\\[-\rowheight]
\printcelltop & \printcelltop \\
\vcell{HMM} & \vcell{Parts of Speech Tagging (Hammad Ali \cite{ali2010unsupervised}) and Speech Processing and recognition (Hasnat et al. \cite{hasnat2007isolated}, \hl{Accuracy:90\%}).\newline} \\[-\rowheight]
\printcelltop & \printcelltop \\
\vcell{Hash Mapping} & \vcell{Parts of Speech Tagging (Ismail \etal\cite{ismail2014developing}).\newline} \\[-\rowheight]
\printcelltop & \printcelltop \\
\vcell{Anaphora-Cataphora\par{} Resolution Methods\newline} & \vcell{Question Answering System (Khan et al. \cite{khan2018improving}, \hl{ Accuracy:74\%}).} \\[-\rowheight]
\printcelltop & \printcelltop \\
\vcell{Basic Classical Methods} & \vcell{Spam and Fake Detection (Hossain et al. \cite{hossain2020banfakenews}, \hl{Accuracy:91\%}).\newline}\\[-\rowheight]
\printcelltop & \printcelltop \\
\vcell{Heuristic Methods} & \vcell{Text Summarization (Abujar et al. \cite{abujar2017heuristic}\hl{, Accuracy:86\%}).\newline}  \\[-\rowheight]
\printcelltop & \printcelltop \\
\vcell{Sentence Ranking Methods} & \vcell{Text Summarization (Efat \etal \cite{efat2013automated}\hl{, F1 score:83.57}).\newline} \\[-\rowheight]
\printcelltop & \printcelltop \\
\vcell{Apriori Methods} & \vcell{Word Sense Disambiguation (Kaysar et al. \cite{kaysar2018word}, \hl{Accuracy:46\%}).\newline}\\[-\rowheight]
\printcelltop & \printcelltop \\
\vcell{Parse Tree Methods} & \vcell{Word Sense Disambiguation (Haque \etal \cite{haque2016bangla}, \hl{Accuracy:82.40\%}).\newline} \\[-\rowheight]
\printcelltop & \printcelltop \\
\vcell{Linear Predictive Coding} & \vcell{Speech Processing and Recognition (Aadit et al.  \cite{aadit2016pitch}).\newline}  \\[-\rowheight]
\printcelltop & \printcelltop \\
\vcell{Gaussian Mixture Model} & \vcell{Speech Processing and Recognition (Ali et al. \cite{ali2013automatic}, \hl{Accuracy:84\%}).\newline} \\[-\rowheight]
\printcelltop & \printcelltop \\
\vcell{Dynamic Time Warping} & \vcell{Speech Processing and Recognition (Ali et al. \cite{ali2013automatic}, \hl{Accuracy:78\%}).\newline}  \\[-\rowheight]
\printcelltop & \printcelltop \\

\hline
\end{tabu}
}
\end{table*}
\begin{table*}
\centering
\caption{Machine Learning Methods used in BNLP.}
\label{tab:ML-BNLP}
\resizebox{\linewidth}{!}{%
\begin{tabu}{>{\hspace{0pt}}p{0.238\linewidth}>{\hspace{0pt}}p{0.703\linewidth}}
\hline
\toprule

\multicolumn{1}{>{\hspace{0pt}}m{0.238\linewidth}}{} & \multicolumn{1}{>{\hspace{0pt}}m{0.703\linewidth}}{} \\
\multicolumn{1}{>{\centering\hspace{0pt}}p{0.238\linewidth}}{\vcell{\textbf{Machine Leaning and Deep Learning Methods}}} & \multicolumn{1}{>{\centering\arraybackslash\hspace{0pt}}p{0.703\linewidth}}{\vcell{\textbf{Use Cases \hl{and Best Results}}}} \\[-\rowheight]
\multicolumn{1}{>{\centering\hspace{0pt}}m{0.238\linewidth}}{\printcelltop} & \multicolumn{1}{>{\centering\arraybackslash\hspace{0pt}}m{0.703\linewidth}}{\printcelltop} \\ 
\\
\hline\hline
\multicolumn{1}{>{\hspace{0pt}}m{0.238\linewidth}}{} & \multicolumn{1}{>{\hspace{0pt}}m{0.703\linewidth}}{} \\
 \vcell{CNN} &
  \vcell{Information Extraction (Sharif et al. \cite{7853957}, \hl{ Accuracy:78\%}), Speech Processing and Recognition (Sharmina et al.\cite{sharmin2020bengali}, \hl{ Accuracy:98.37\%} and Shuvo et al.\cite{shuvo2019bangla}, \hl{ Accuracy:93.65\%}), Question Answering System (Sarker et al. \cite{sarker2019bengali}, \hl{ Accuracy:90.6\%}), and Sentiment Analysis (Tripto et al. \cite{tripto2018detecting}\,hl{ Accuracy:60.89\%} and Sarkar \cite{sarkar2019sentiment},\hl{ Accuracy:60.89\%}). \newline} \\[-\rowheight]
\printcelltop & \printcelltop \\

  \vcell{Backpropagation Neural\newline Network (BPNN)\newline} &
  \vcell{Speech Processing and Recognition (Ahammad et al. \cite{ahammad2016connected}, \hl{ Accuracy:98.72\%}). \newline } \\[-\rowheight]
\printcelltop & \printcelltop \\

  \vcell{Transfer Learning} &
  \vcell{Speech Processing and Recognition (Sumon et al. \cite{sumon2018bangla}, \hl{ Accuracy:73\%}).\newline } \\[-\rowheight]
\printcelltop & \printcelltop \\

  \vcell{Gated Recurrent Unit (GRU) \newline based RNN\newline} &
  \vcell{Speech Processing and Recognition (Sumit et al. \cite{sumit2018noise}, \hl{ Character error rate:9.15\%}).\newline}  \\[-\rowheight]
\printcelltop & \printcelltop \\

  \vcell{RNN} &
  \vcell{Speech Processing and Recognition (Islam et al. \cite{islam2019speech}, \hl{ Accuracy:86.058\%}) and Named Entity Recognition (Banik et al. \cite{banik2018gru}, \hl{ F1-score:69\%}).\newline}  \\[-\rowheight]
\printcelltop & \printcelltop \\

  \vcell{ANN} &
  \vcell{Speech Processing and Recognition (Sarma et al. \cite{sarma2015acoustic}, ), Question Answering System, Spam and Fake Detection, and Information Extraction (Rahman et al. \cite{rahman2009real}, \hl{ Accuracy:91.5\%}).\newline}  \\[-\rowheight]
\printcelltop & \printcelltop \\

  \vcell{Deep Generative Models\newline based methods\newline} &
  \vcell{Speech Processing and Recognition (Ahmed et al. \cite{ahmed2015acoustic}, \hl{ Accuracy:94.05\%}).\newline}  \\[-\rowheight]
\printcelltop & \printcelltop \\

  \vcell{DNN} &
  \vcell{Speech Processing and Recognition (Ahmed et al. \cite{ahmed2015acoustic}, \hl{ Accuracy:94.05\%}).\newline}  \\[-\rowheight]
\printcelltop & \printcelltop \\

  \vcell{Naive Bayes} &
  \vcell{Question Answering System (Kowsher et al. \cite{kowsher2019bangla}, \hl{ Accuracy:91.34\%}), Word Sense Disambiguation (Nazah et al. \cite{nazah2017word}, \hl{ Accuracy:77.5\%}), and Sentiment Analysis (Tripto et al. \cite{tripto2018detecting},  \cite{islam2019design},\hl{ Accuracy:60\%} \cite{islam2016supervised}, Mahtab et al. \cite{mahtab2018sentiment},\hl{ Precision:77\%} \cite{mahtab2018sentiment},\hl{ Accuracy:70.47\%}, and Ghosal et al.
  \cite{ghosal2015sentiment},\hl{ Accuracy:91.67\%}).\newline}  \\[-\rowheight]
\printcelltop & \printcelltop \\

  \vcell{SVM} &
  \vcell{Question Answering System (Kowsher et al. \cite{kowsher2019bangla}, \hl{ Accuracy:91.34\%}), Spam and Fake Detection (Hussain et al. \cite{hussain2020detection}, \hl{ Accuracy:96.64\%}), and Sentiment Analysis (Chowdhury et al. \cite{chowdhury2014performing},\hl{ F-score:0.93}, Tripto et al. \cite{tripto2018detecting},\hl{ Accuracy:60\%}, Mahtab et al. \cite{mahtab2018sentiment},\hl{ Accuracy:73.49\%}, Sarkar et al. \cite{sarkar2017sentiment},\hl{ Accuracy:45\%}, and Ghosal et al. \cite{ghosal2015sentiment},\hl{ Accuracy:98.7\%}).\newline}  \\[-\rowheight]
\printcelltop & \printcelltop \\

  \vcell{Stocastic Gradient Descent} &
  \vcell{Question Answering System (Islam et al. \cite{islam2016word}, \hl{ Precision:0.95562\%}) and Sentiment Analysis (Mandal et al. \cite{mandal2018preparing},\hl{ F-score:78.70\%}).\newline}  \\[-\rowheight]
\printcelltop & \printcelltop \\

  \vcell{Decision Tree} &
  \vcell{Question Answering System (Sarker et al. \cite{sarker2019bengali}, \hl{ Accuracy:75.3\%}) and Sentiment Analysis (Mahtab et al. \cite{mahtab2018sentiment}, \hl{ Accuracy:64.765\%} and Ghosal et al. \cite{ghosal2015sentiment},\hl{ Accuracy:95.167\%}).\newline}  \\[-\rowheight]
\printcelltop & \printcelltop \\

  \vcell{N-gram Model} &
  \vcell{Question Answering System (Islam et al. \cite{islam2019design}, \hl{ Precision:0.35\%}), Machine Translation (Islam et al. \cite{islam2010english},\hl{ BLEU:11.70, NIST:4.27 and TER:0.76}), and Parts of Speech Tagging (Hasan et al. \cite{hasan2007comparison},\hl{ Accuracy:90\%}).\newline}  \\[-\rowheight]
\printcelltop & \printcelltop \\

  \vcell{Logistic Regression} &
  \vcell{Spam and Fake Detection (Ahmed et al. \cite{ahmed2018detecting}, \hl{ Accuracy:89\%}).\newline}  \\[-\rowheight]
\printcelltop & \printcelltop \\

  \vcell{Lagrangian Support Vector\newline Machine\newline} &
  \vcell{Spam and Fake Detection (Ahmed et al. \cite{ahmed2018detecting}, \hl{ Accuracy:92\%}).\newline}  \\[-\rowheight]
\printcelltop & \printcelltop \\

  \vcell{Maximum Entropy} &
  \vcell{Sentiment Analysis (Chowdhury et al. \cite{chowdhury2014performing},\hl{ F-score:0.85}).\newline}  \\
  [-\rowheight]
\printcelltop & \printcelltop \\

  \vcell{LSTM based RNN} &
  \vcell{Sentiment Analysis (Hassan et al. \cite{hassan2016sentiment},\hl{ Accuracy:70\%}, Tripto et al.\cite{tripto2018detecting},\hl{ Accuracy:65.96\%}, and Uddin et al. \cite{uddin2019extracting},\hl{ Accuracy:77.1\%}) and Speech Processing and Recognition (Nahid et al. \cite{nahid2017bengali}, \hl{ Word detection error rate:13.2\%}).\newline}  \\
  [-\rowheight]
\printcelltop & \printcelltop \\

 \vcell{Multidimensional Naïve Bayes} &
  \vcell{Spam and Fake Detection (Islam et al. \cite{islam2019using}, \hl{ Accuracy:82.44\%}) and Sentiment Analysis (Sarkar et al. \cite{sarkar2017sentiment},\hl{ Accuracy:44.20\%}).\newline}  \\
  [-\rowheight]
\printcelltop & \printcelltop \\

  \vcell{KNN} &
  \vcell{Information Extraction (Mandal et al. \cite{mandal2011handwritten}, \hl{ Accuracy:88.95\%}), Word Sense Disambiguation (Pal et al. \cite{pal2018word}, \hl{ Precision: 84\%}, and Pandit et al. \cite{pandit2015memory}, \hl{ Accuracy:71\%}). \newline} \\
  [-\rowheight]
\printcelltop & \printcelltop \\

  \vcell{Random Forest} &
  \vcell{Sentiment Analysis (Ghosal et al. \cite{ghosal2015sentiment},\hl{ Accuracy:98.65\%}).\newline}  \\
  [-\rowheight]
\printcelltop & \printcelltop \\

  \vcell{K-Means Clustering} &
  \vcell{Text Summarization (Akter et al. \cite{akter2017extractive}).\newline}  \\
  [-\rowheight]
\printcelltop & \printcelltop \\

  \vcell{Unsupervised Method\newline using ULMFiT Model\newline}  &
  \vcell{Text Summarization (Chowdhury et  al. \cite{rayan2021unsupervised},\hl{ Rouge-1:12.17, Rogue-2:1.92, and Rouge-L:11.35}) and Parsing (Das et al. \cite{das2010morphological},\hl{ Accuracy:74.6\%}).}
\\
[-\rowheight]
\printcelltop & \printcelltop \\
\hline
\end{tabu}
}
\end{table*}
\begin{table*}
\centering
\caption{Summary Table for Sentiment Analysis and Machine Translation Datasets.}
\label{tab:text-datasets-fuad}
\begin{tabular}{>{\hspace{0pt}}p{0.071\linewidth}>{\hspace{0pt}}p{0.111\linewidth}>{\hspace{0pt}}p{0.653\linewidth}>{\hspace{0pt}}p{0.065\linewidth}} 
\hline
\toprule \\
\multicolumn{1}{>{\centering\hspace{0pt}}m{0.071\linewidth}}{\textbf{Name}} & \multicolumn{1}{>{\centering\hspace{0pt}}m{0.111\linewidth}}{\textbf{Article}} & \multicolumn{1}{>{\centering\hspace{0pt}}m{0.653\linewidth}}{\textbf{Description}} & \multicolumn{1}{>{\centering\arraybackslash\hspace{0pt}}m{0.065\linewidth}}{\textbf{Task}} \\ 
\\
\hline\hline
\vcell{Bengali ABSA dataset} & \vcell{Mahtab \etal \cite{mahtab2018sentiment}} & \vcell{Aspect based sentiment analysis dataset. ABSA Bengali dataset containing 2,979 data and manually collected data contained 1,601 test samples. The manual dataset was collected from the Prothom-Alo newspaper. The dataset was labelled as praise, criticism, and sadness. There were 513, 604, and 484 labelled data on praise,~ criticism, and sadness, respectively.~} & \vcell{Sentiment analysis.} \\[-\rowheight]
\\
\printcelltop & \printcelltop & \printcelltop & \printcelltop \\
\vcell{SAIL tweeter dataset} & \vcell{Kamal Sarkar \cite{sarkar2019sentiment}} & \vcell{SAIL dataset of Bengali tweets (1,500 data). The dataset contains three labelled classes (positive, neutral, and negative). The dataset consists of 356 negative, 368 positives, and 276 neutral examples.~} & \vcell{Sentiment analysis.} \\[-\rowheight]
\\
\printcelltop & \printcelltop & \printcelltop & \printcelltop \\
\vcell{EMILLE corpus} & \vcell{Islam \etal \cite{islam2010english}} & \vcell{This corpus contains three components: monolingual, parallel, and annotated corpora. Fourteen are~ monolingual corpora, including both written and spoken data for fourteen South Asian languages. The corpora contains approximately 92,799,000~ words (2,627,000 words of transcribed spoken data for various South Asian languages).} & \vcell{Machine translation.} \\[-\rowheight]
\\
\printcelltop & \printcelltop & \printcelltop & \printcelltop \\
\vcell{KDE4 corpus} & \vcell{Islam \etal \cite{islam2010english}} & \vcell{A parallel corpus containing 92 languages,~60.75M tokens, 8.89M sentence fragments.} & \vcell{Machine translation.} \\[-\rowheight]
\\
\printcelltop & \printcelltop & \printcelltop & \printcelltop \\
\bottomrule
\end{tabular}
\end{table*}

\begin{table*}
\centering
\caption{Summary Table for Bangla Spam and Fake Detection System Datasets.}
\label{tab:spamdatasettab}
\resizebox{\linewidth}{!}{%
\begin{tabu}{>{\hspace{0pt}}p{0.071\linewidth}>{\hspace{0pt}}p{0.726\linewidth}>{\hspace{0pt}}p{0.142\linewidth}} 
\toprule
\multicolumn{1}{>{\hspace{0pt}}m{0.071\linewidth}}{} & \multicolumn{1}{>{\hspace{0pt}}m{0.726\linewidth}}{} & \multicolumn{1}{>{\hspace{0pt}}m{0.142\linewidth}}{} \\
\multicolumn{1}{>{\centering\hspace{0pt}}p{0.071\linewidth}}{\vcell{\textbf{Article}}} & \multicolumn{1}{>{\centering\hspace{0pt}}p{0.726\linewidth}}{\vcell{\textbf{Dataset Description}}} & \multicolumn{1}{>{\centering\arraybackslash\hspace{0pt}}p{0.142\linewidth}}{\vcell{\textbf{Task}}} \\[-\rowheight]
\multicolumn{1}{>{\centering\hspace{0pt}}m{0.071\linewidth}}{\printcelltop} & \multicolumn{1}{>{\centering\hspace{0pt}}m{0.726\linewidth}}{\printcelltop} & \multicolumn{1}{>{\centering\arraybackslash\hspace{0pt}}m{0.142\linewidth}}{\printcelltop} \\ 
\\
\hline\hline
\multicolumn{1}{>{\hspace{0pt}}m{0.071\linewidth}}{} & \multicolumn{1}{>{\hspace{0pt}}m{0.726\linewidth}}{} & \multicolumn{1}{>{\hspace{0pt}}m{0.142\linewidth}}{} \\
\vcell{Islam et al. \cite{islam2019using}\newline} & \vcell{The proposed dataset has 1,965 variational instances, including 646 spam and 1,319 ham data.} & \vcell{Spam detection from malicious Bangla text data.} \\[-\rowheight]
\printcelltop & \printcelltop & \printcelltop \\ \\
\vcell{Hussain et al. \cite{hussain2020detection}} & \vcell{The authors collected around 2,500 articles for building the dataset from public articles. 1,548 articles from the popular website were considered as real news, and 993 articles from satire news articles were considered as fake news.\newline} & \vcell{Fake Bangla news detection.} \\[-\rowheight]
\printcelltop & \printcelltop & \printcelltop \\
\vcell{Hossain et al. \cite{hossain2020banfakenews}} & \vcell{The Dataset contains 50K news, and 8.5K news of them were manually annonated. Misleading, clickbait, and satire: these three news types were collected from 22 popular news portals of Bangladesh.} & \vcell{Noisy isolated and continuous speech recognition.} \\[-\rowheight]
\printcelltop & \printcelltop & \printcelltop \\
\bottomrule
\end{tabu}
}
\end{table*}

\par 
\begin{table*}
\centering
\caption{Summary Table for Bangla Question Answering System Datasets.}
\label{tab:questiondatasettab}
\resizebox{\linewidth}{!}{%
\begin{tabu}{>{\hspace{0pt}}p{0.083\linewidth}>{\hspace{0pt}}p{0.628\linewidth}>{\hspace{0pt}}p{0.228\linewidth}} 
\toprule
\multicolumn{1}{>{\hspace{0pt}}m{0.083\linewidth}}{} & \multicolumn{1}{>{\hspace{0pt}}m{0.628\linewidth}}{} & \multicolumn{1}{>{\hspace{0pt}}m{0.228\linewidth}}{} \\
\multicolumn{1}{>{\centering\hspace{0pt}}p{0.083\linewidth}}{\vcell{\textbf{Article}}} & \multicolumn{1}{>{\centering\hspace{0pt}}p{0.628\linewidth}}{\vcell{\textbf{Dataset Description}}} & \multicolumn{1}{>{\centering\arraybackslash\hspace{0pt}}p{0.228\linewidth}}{\vcell{\textbf{Task}}} \\[-\rowheight]
\multicolumn{1}{>{\centering\hspace{0pt}}m{0.083\linewidth}}{\printcelltop} & \multicolumn{1}{>{\centering\hspace{0pt}}m{0.628\linewidth}}{\printcelltop} & \multicolumn{1}{>{\centering\arraybackslash\hspace{0pt}}m{0.228\linewidth}}{\printcelltop} \\ 
\\
\hline\hline
\multicolumn{1}{>{\hspace{0pt}}m{0.083\linewidth}}{} & \multicolumn{1}{>{\hspace{0pt}}m{0.628\linewidth}}{} & \multicolumn{1}{>{\hspace{0pt}}m{0.228\linewidth}}{} \\
\vcell{Khan et al. \cite{khan2018improving}} & \vcell{50 affirmative sentences of 50 documents were selected where one possesses nouns, and another possesses pronouns of corresponding nouns. Ten question queries for each document were selected.\newline} & \vcell{Improving answer extraction technique In Bangali Q/A system.} \\[-\rowheight]
\printcelltop & \printcelltop & \printcelltop \\
\vcell{Islam et al. \cite{islam2016word}\newline} & \vcell{3,334 questions were collected from random documents.\par{}} & \vcell{Word or phrase based answer type classification for Bangla Q/A system.} \\[-\rowheight]
\printcelltop & \printcelltop & \printcelltop \\
\vcell{Islam et al. \cite{islam2019design}\newline} & \vcell{Several documents from Wikipedia were collected, and 500 questions were selected from these proposed documents. \par{}} & \vcell{Bangla Q/A system for multiple documents.} \\[-\rowheight]
\printcelltop & \printcelltop & \printcelltop \\
\vcell{Uddin et al. \cite{uddin2020end}\newline} & \vcell{11,500 question-answer pairs based on history domain related questions and single supporting line were created from Bangla wiki2 corpus.} & \vcell{Bangla paraphrased Q/A system.} \\[-\rowheight]
\printcelltop & \printcelltop & \printcelltop \\
\vcell{Sarker et al. \cite{sarker2019bengali}\newline} & \vcell{15,355 questions and 220 documents were selected from a closed domain.} & \vcell{Bengali Q/A System for Factoid Questions.\par{}\textbf{}} \\[-\rowheight]
\printcelltop & \printcelltop & \printcelltop \\
\vcell{Kowsher et al. \cite{kowsher2019bangla}} & \vcell{74 topics and 3,127 questions from the topics were selected for training, and 2,852 questions from the relevant topics were selected for testng.} & \vcell{Bangla intelligence-based Q/A system.} \\[-\rowheight]
\printcelltop & \printcelltop & \printcelltop \\
\bottomrule
\end{tabu}
}
\end{table*}

\begin{table*}
\centering
\caption{Summary Table for Bangla Speech Processing and Recognition System Datasets.}
\label{tab:speechdatasettab}
\resizebox{\linewidth}{!}{%
\begin{tabu}{>{\hspace{0pt}}p{0.078\linewidth}>{\hspace{0pt}}p{0.428\linewidth}>{\hspace{0pt}}p{0.28\linewidth}>{\hspace{0pt}}p{0.165\linewidth}} 
\toprule
\multicolumn{1}{>{\hspace{0pt}}m{0.078\linewidth}}{} & \multicolumn{1}{>{\hspace{0pt}}m{0.428\linewidth}}{} & \multicolumn{1}{>{\hspace{0pt}}m{0.28\linewidth}}{} & \multicolumn{1}{>{\hspace{0pt}}m{0.165\linewidth}}{} \\
\multicolumn{1}{>{\centering\hspace{0pt}}p{0.078\linewidth}}{\vcell{\textbf{Article}}} & \multicolumn{1}{>{\centering\hspace{0pt}}p{0.428\linewidth}}{\vcell{\textbf{Dataset Description}}} & \multicolumn{1}{>{\centering\hspace{0pt}}p{0.28\linewidth}}{\vcell{\textbf{Speakers Specification}}} & \multicolumn{1}{>{\centering\arraybackslash\hspace{0pt}}p{0.165\linewidth}}{\vcell{\textbf{Speech Environment and Task}}} \\[-\rowheight]
\multicolumn{1}{>{\centering\hspace{0pt}}m{0.078\linewidth}}{\printcelltop} & \multicolumn{1}{>{\centering\hspace{0pt}}m{0.428\linewidth}}{\printcelltop} & \multicolumn{1}{>{\centering\hspace{0pt}}m{0.28\linewidth}}{\printcelltop} & \multicolumn{1}{>{\centering\arraybackslash\hspace{0pt}}m{0.165\linewidth}}{\printcelltop} \\
\\
\hline\hline
\multicolumn{1}{>{\hspace{0pt}}m{0.06\linewidth}}{} & \multicolumn{1}{>{\hspace{0pt}}m{0.441\linewidth}}{} & \multicolumn{1}{>{\hspace{0pt}}m{0.299\linewidth}}{} & \multicolumn{1}{>{\hspace{0pt}}m{0.136\linewidth}}{} \\
\vcell{Sultana et al. \cite{sultana2012bangla}\newline} & \vcell{396 Bangla words with five paragraphs and 270 words of them are distinct.} & \vcell{Not mentioned.} & \vcell{Speech to text conversion of newspapers' data.\par{}} \\[-\rowheight]
\printcelltop & \printcelltop & \printcelltop & \printcelltop \\
\vcell{Ali et al. \cite{ali2013automatic}\newline} & \vcell{100 words were recorded 1,000 times. Each word was recorded with a 3-second duration and ten repetitions.} & \vcell{Not mentioned.} & \vcell{Noisy isolated speech recognition from random articles.\newline} \\[-\rowheight]
\printcelltop & \printcelltop & \printcelltop & \printcelltop \\
\vcell{Hasnat et al. \cite{hasnat2007isolated}\newline} & \vcell{100 words were used for testing purposes. 5 to 10 samples for each word were taken for the training purposes of isolated speech recognition, and 1,814 phonemes were used for training purposes of continuous speech recognition.\par{}} & \vcell{Five speakers.} & \vcell{Noisy isolated and continuous speech recognition from random articles.\newline} \\[-\rowheight]
\printcelltop & \printcelltop & \printcelltop & \printcelltop \\
\vcell{Aadit et al. \cite{aadit2016pitch}\newline} & \vcell{Not mentioned.} & \vcell{Twenty-five males and 25 females, and their age range is from 20 to 25 years.} & \vcell{Pitch and formant estimation of noise-free speech data.} \\[-\rowheight]
\printcelltop & \printcelltop & \printcelltop & \printcelltop \\ \\
\vcell{Ahmed et al. \cite{ahmed2018implementation}\newline} & \vcell{500 sentences containing 2,733 Bangla word speeches for training the proposed model. One hundred audio files from 20 more speakers were used for testing the proposed model.} & \vcell{Seven speakers consisting of five males and two females, all aged between 19 and 23 years, were from five various locations of Bangladesh.\newline} & \vcell{Noisy numeral speech recognition.} \\[-\rowheight]
\printcelltop & \printcelltop & \printcelltop & \printcelltop \\
\vcell{Paul et al. \cite{paul2021bengali}\newline} & \vcell{1,000 audio samples for (0-9) classes.} & \vcell{Not mentioned.} & \vcell{Numeral speech recognition.} \\[-\rowheight]
\printcelltop & \printcelltop & \printcelltop & \printcelltop \\
\vcell{Ahammad et al. \cite{ahammad2016connected}\newline} & \vcell{Two datasets of connected speech in Bangla for training and testing.} & \vcell{15 male and 15 female speakers.} & \vcell{Noise free numeral speech recognition.} \\[-\rowheight]
\printcelltop & \printcelltop & \printcelltop & \printcelltop \\
\vcell{Nahid et al. \cite{nahid2017bengali}\newline} & \vcell{2,000 words and 114 words of them are unique.} & \vcell{15 male speakers of age between 20 to 24.} & \vcell{Noisy real numbers speech recognition.} \\[-\rowheight]
\printcelltop & \printcelltop & \printcelltop & \printcelltop \\
\vcell{Sumit et al. \cite{sumit2018noise}} & \vcell{A self-constructed (Babel) dataset was used in both the training and test phases, and CRBLP speech corpora was used only in the test phase. Training data contained about 350 hours of Bangla speech, and Testing data included approximately 111 hours of Bangla speech.\newline} & \vcell{Not mentioned.} & \vcell{Augmented noisy speech recognition from bus, cafe, street and pedestrian areas.} \\[-\rowheight]
\printcelltop & \printcelltop & \printcelltop & \printcelltop \\
\vcell{Sumon et al. \cite{sumon2018bangla}\newline} & \vcell{The developed dataset has ten classes of data and 100 samples per class, and each utterance duration was less than 2 seconds. The pre-trained English speech dataset has 65,000 samples, and each utterance duration was not more than 1 second.} & \vcell{Not mentioned.} & \vcell{Short speech command recognition.} \\[-\rowheight]
\printcelltop & \printcelltop & \printcelltop & \printcelltop \\ \\
\vcell{Islam et al. \cite{islam2019speech}\newline} & \vcell{The dataset contains 200,000 wav audio files, and 33,000 audio files were used for a RNN model for the experiment.} & \vcell{500 speakers.} & \vcell{General public speech recognition.} \\[-\rowheight]
\printcelltop & \printcelltop & \printcelltop & \printcelltop \\
\vcell{Shuvo et al. \cite{shuvo2019bangla}} & \vcell{6,000 utterances and each report was one second in length. 420 instances of each digit were used in the training set, and 180 cases of each number were used in the testing set.} & \vcell{60 male and 60 female native Bangla speakers from different regions of Bangladesh. Five trails of each digit from each speaker were recorded.\newline} & \vcell{Noise-free numeral speech recognition.} \\[-\rowheight]
\printcelltop & \printcelltop & \printcelltop & \printcelltop \\
\vcell{Sharmina et al. \cite{sharmin2020bengali}\newline} & \vcell{1,230 audio files.} & \vcell{Five males and five females. The people are from different age groups and various parts of Bangladesh.} & \vcell{Numeral speech recognition.} \\ [-\rowheight]
\printcelltop & \printcelltop & \printcelltop & \printcelltop \\ \\
\vcell{Ovishake et al. \cite{sen2021convolutional}\newline} & \vcell{4,000 audio samples.} & \vcell{Nineteen speakers and the speakers are from different age groups and various parts of Bangladesh.} & \vcell{Numeral speech recognition.} \\ [-\rowheight]
\printcelltop & \printcelltop & \printcelltop & \printcelltop \\ \\
\vcell{Paul et al. \cite{paul2009bangla}\newline} & \vcell{There were 60 training examples for each digit, and in the testing phase, there were ten samples for each number.} & \vcell{Four different Bangla words uttered by four different persons.} & \vcell{General public speech recognition.} \\[-\rowheight]
\printcelltop & \printcelltop & \printcelltop & \printcelltop \\
\vcell{Sultan et al.\cite{sultan2021adrisya}\newline} & \vcell{Bangla human voice commands.} & \vcell{Not mentioned.} & \vcell{Bangla virtual assistant.} \\ [-\rowheight]
\printcelltop & \printcelltop & \printcelltop & \printcelltop \\
\vcell{Khan et al. \cite{khan2018construction}} & \vcell{54,000 audio files were maintained as a training corpus, and 27,000 audio files were held as a testing corpus. A list of high-frequent 1,081 words were selected for recording a total of 292 hours.} & \vcell{Training data were collected from 50 male and 50 female persons, and test data were collected from another 25 male and 25 female persons. The speakers were in the range of 18 to 25 years.\newline} & \vcell{Creation of noisy isolated speech dataset.} \\[-\rowheight]
\printcelltop & \printcelltop & \printcelltop & \printcelltop \\
\vcell{Khan et al. \cite{khan2018creation} } & \vcell{26,000 audio files were maintained as a training corpus, and 15,600 audio files were kept as a testing corpus. 52 sentences were selected from 1,000 high-frequency words for recording a total of 62 hours.} & \vcell{Train data were collected from 50 male and 50 female persons, and test data were collected from another 25 male and 25 female persons. The speakers are in the range of 18 to 25 years.} & \vcell{Creation of noisy connected speech dataset.} \\[-\rowheight]
\printcelltop & \printcelltop & \printcelltop & \printcelltop \\
\bottomrule
\end{tabu}
}
\end{table*}

\subsection{FUTURE TRENDS IN Text Format}

\hl{The future trends in developing Bangla question answering systems include: developing algorithmic Bangla Information Retrieval System (BIRS)} \cite{futurequestion}\hl{, classification of Bangla questions towards a factoid question answering system} \cite{monisha2019classification}\hl{, developing both closed domain-based and open domain-based question answering systems, developing of Bangla datasets for both of the open and closed domain-based question answering systems and question answering appplications based on machine learning and deep learning methods and many more. \par
Bangla spam and fake detection system's future research trends include: developing hateful speech detection in public Facebook pages and other social sites }\cite{ishmam2019hateful}\hl{, misinformation detection on online social networks }\cite{islam2020deep}\hl{, developing of Bangla datasets for both of the spam and ham data and many more. However, most of the researches on Bangla spam and fake detection applications is now following machine learning and deep learning methods.
\par
The future trends in developing Bangla Text summarization is moving towards developing more accurate extractive and abstractive summarization techniques. Abstractive summarization is a complex task and very few works have been done on this topic. The extractive summarization has some existing works but needs more improvements for matching the state of the art quality. \par 
In recent years, a lot of works have been done on Bangla sentiment analysis. With the more availability of datasets in this section, the accuracy is increasing. But most of the sentiment analysis work was done in three (positive, negative and neutral) classes. Five or more classes of sentiment work is being introduced with a respective dataset which will bring more dimentionality in sentiment analysis task.\par
Bangla parts of speech tagging require a different approach than the approaches taken in English parts of speech tagging. The previous works were basically based on a rule-based approach. The future trend in this topic may focus on a hybrid approach that will increase efficiency. \par 
Due to the increasing availability of dataset and vocabulary Bangla machine translation is gradually improving. The recent works are more focused on using the machine learning approach as it has proved to be more efficient and accurate. The future trend of machine translation will primarily be in the hand of the machine learning approach. \par
Bangla word sense disambiguation system's future trends of research include: developing word sense disambiguation applications using supervised methodology }\cite{pal2019novel}\hl{, developing modern tools and frameworks, and creating Bangla datasets for working with Bangla word sense disambiguation applications. Now-a-days, most of the researches on Bangla word sense disambiguation applications are using machine learning and deep learning methods. \par
The future research trends in Bangla information extraction systems include: developing proper linguistic tools for extracting the information, creating datasets for information extraction applications, and many more. }\hl{Moreover, modern machine learning and deep learning techniques are commonly used for developing Bangla information extraction systems.\par
Bangla Named Entity Recognition systems future research directions include: developing proper linguistic tools and machine-readable dictionaries, creating datasets for named entity recognition applications, newest machine learning and deep learning techniques are also being used for developing Bangla named entity recognition systems and many more.}

\subsection{FUTURE TRENDS IN SPEECH FORMAT}

\hl{Bangla speech processing and recognition system's future research directions are: developing speech recognition systems for continuous and isolated speech data with modern machine learning and deep learning techniques, introducing larger datasets for Bangla natural speech for improving the performance of automatic speech recognition systems, developing modern tools and frameworks for detecting speech data, working with accents and variational local speech data. Some more future directions might include: developing speech recognition systems for non-native Bangla speakers, code-switching and code-mixing incidences, and noisy speech data. }\hl{Furthermore, creating datasets for isolated and continuous speech data and many more future works can be done in this field. Today, modern machine learning and deep learning techniques are being used in most speech processing and recognition applications and producing excellent outcomes with high accuracy. Speech recognition research will continue to produce novel methods and techniques in machine learning and deep learning field.}\par

\section{BNLP Datasets}
\label{sec:datasets}
\subsection{Text Datasets}
\par
Table \ref{tab:text-datasets-fuad} shows the short description of datasets that are used in Bangla sentiment analysis and machine translation. Short description of datasets that are used in Bangla spam and fake detection systems are shown in Table \ref{tab:spamdatasettab}. We also show the short description of datasets that are used in Bangla question answering system in
Table \ref{tab:questiondatasettab}.\par

\subsection{Speech Datasets}

In 2018, Khan et al. \cite{khan2018construction} proposed a technique of developing a speech corpus of isolated words in the Bangla language that was recorded, including high-frequency words from a text corpus "BdNC01" \cite{khan2017creation}. Here, 54,000 audio files were maintained as training corpus from 50 male and 50 female persons, and 27,000 audio files were maintained as testing corpus from another 25 male and 25 female persons. The speakers were in the range of 18 to 25 years. High frequent 1,081 words were selected and arranged alphabetically to construct a significantly large scale isolated speech database for recording a total of 292 hours. Noisy files were cleaned using various filters, and shallow speech level data’s amplitudes were magnified to a suitable level. The time length of each file was, on average, 13 seconds. \hl{The researchers concentrated mainly on building an isolated Bangla speech corpus. The authors obtained a lot of audio data for constructing the dataset, but the dataset lacked variations of data concerning gender, age, and places of the speakers; again, the connected Bangla speeches were not recorded in the suggested dataset.}\par
Khan et al. \cite{khan2018creation} proposed another technique of developing a speech corpus of connected Bangla words that was recorded from newspaper text corpus "BdNC01". In this research, the authors dealt with 1,000 high-frequency words. From three issues of daily newspapers, they picked randomly 52 sentences, including these high-frequency words for recording. Twenty-six thousand audio files were maintained as training corpus from 50 male and 50 female persons, and 15,600 audio files were maintained as testing corpus from another 25 male and 25 female persons. The speakers were in the range of 18 to 25 years. Fifty-two sentences were selected from 1,000 high-frequency words for recording a total of 62 hours. Noisy files were cleaned using various filters, and short speech level data’s amplitudes were magnified to a suitable level. The original unedited recorded files were edited for completing a sentence. \hl{The authors primarily focused on establishing a connected Bangla speech corpus. The created dataset lacked data differences for gender, age, and places of the speakers; again, the isolated Bangla speeches were not recorded in the suggested dataset.}
\par
We summarize the datasets that are used in Bangla speech processing and recognition system in Table \ref{tab:speechdatasettab}.

\section{Conclusion}
\label{sec:conclusion}
Natural language processing is an emerging field in the modern machine learning and deep learning realm. The usage of natural language processing is increasing day by day. The importance of BNLP is undeniable as Bangla is one of the languages spoken by many people around the world. We have presented a broad discussion of the BNLP field in this paper. We have covered various preprocessing techniques and discussed many classical and machine learning methods used in this field, giving us an idea about mainly used methods and new areas yet to be explored. \hl{Moreover,} we discussed challenges and future research possibilities and further reviewed the characteristics and complexity essential to understanding modern challenges in this field. The necessity and needs of BNLP in the modern world were also discussed. \hl{Furthermore, }we present various methods and approaches of different categories in the respective sections, giving some brief idea about implemented approaches. We also addressed the challenges faced during the development of BNLP systems. Additionally, we have summarized various datasets which are commonly used. In conclusion, we have presented a comprehensive review of BNLP methods with datasets and results while addressing the limitations, suggesting improvement ideas and discussing current and future trends.

\section*{Acknowledgment}
The authors would like to thank for the support from Taif University Researchers Supporting Project number (TURSP-2020/10), Taif University, Taif, Saudi Arabia.

\bibliographystyle{ieeetr}
\bibliography{bibiliography}

\begin{IEEEbiography}[{\includegraphics[width=1in,height=1.25in,clip,keepaspectratio]{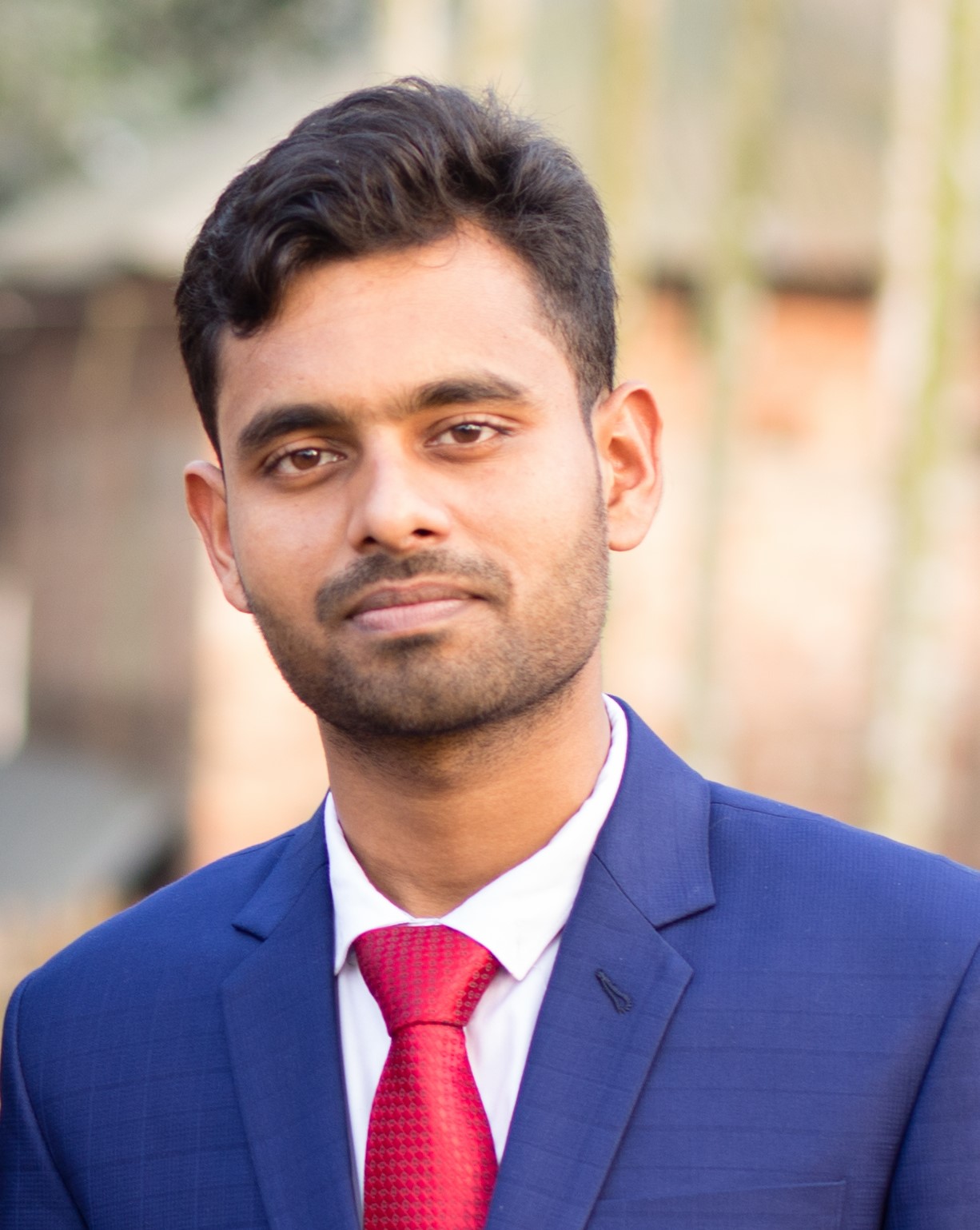}}]{OVISHAKE SEN} was born in Thakurgaon, Bangladesh. Currently, he is pursuing a B.Sc. degree in computer science and engineering (CSE) with the Khulna University of Engineering \& Technology (KUET), Khulna, Bangladesh. He has developed some exciting projects using C, C++, Python, Java, HTML, CSS, ASP.net, SQL, Android, and iOS. His research interests include natural language processing, computer vision, speech processing, machine learning, deep learning, competitive programming, and data science.
\end{IEEEbiography}
\vskip -1\baselineskip plus -1fil
\begin{IEEEbiography}[{\includegraphics[width=1in,height=1.25in,clip,keepaspectratio]{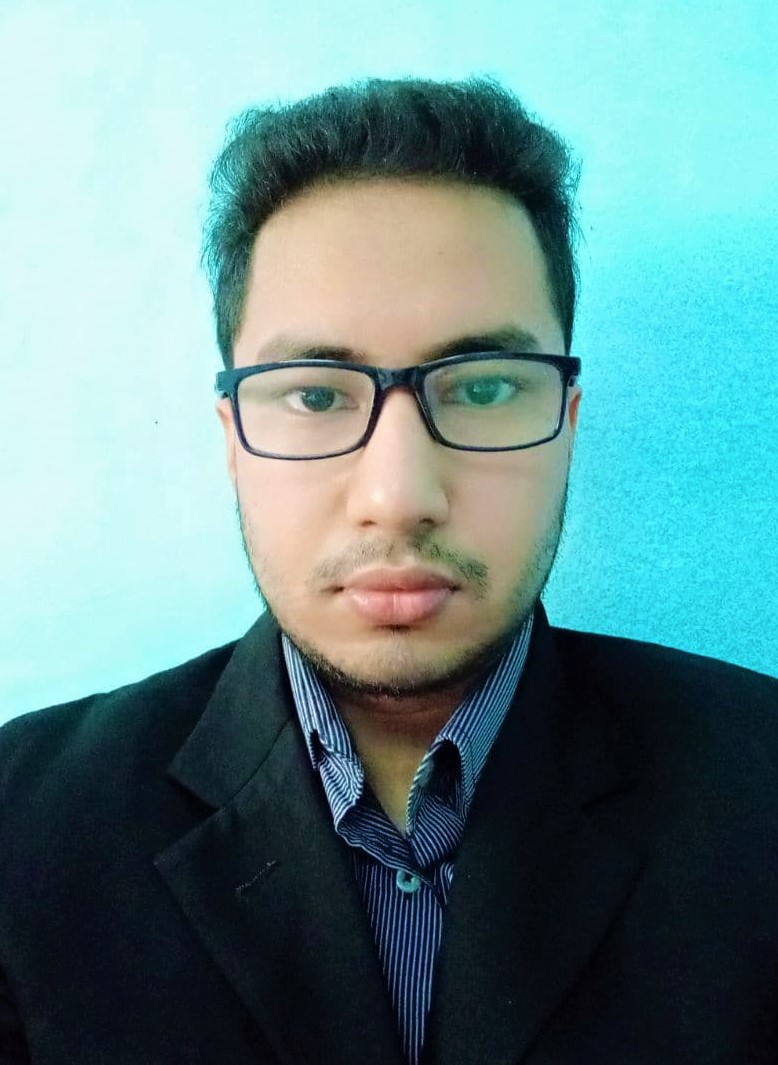}}]{Mohtasim Fuad} was born in Chattrogram, Bangladesh. He is currently pursuing a B.Sc. degree in computer science and engineering (CSE) with the Khulna University of Engineering \& Technology (KUET), Khulna, Bangladesh. His research interests include computer vision, natural language processing, data science, machine learning and deep learning. He is currently working on deep learning projects.
\end{IEEEbiography}

\vskip -1\baselineskip plus -1fil
\begin{IEEEbiography}[{\includegraphics[width=1in,height=1.25in,clip,keepaspectratio]{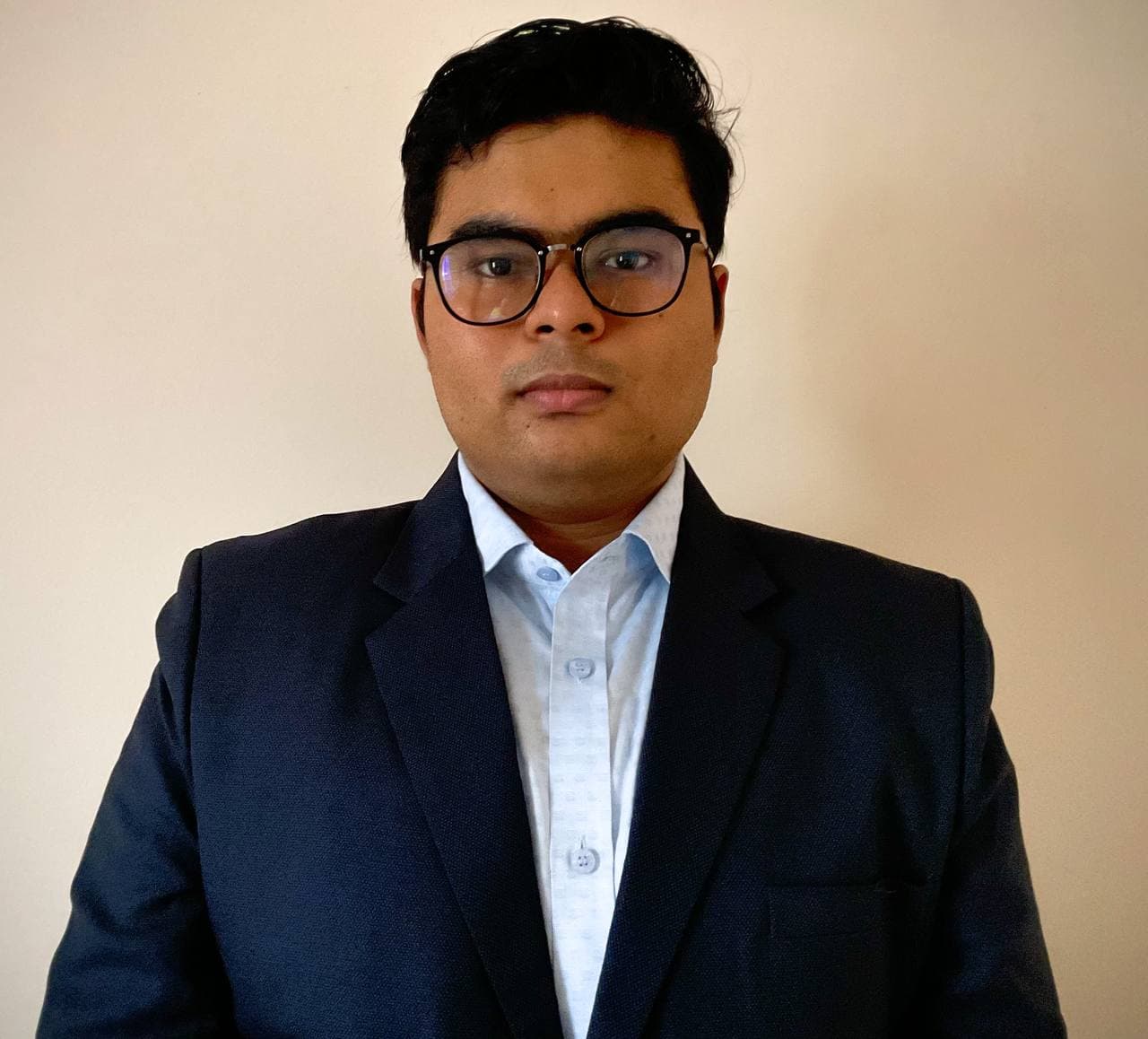}}]{MD. NAZRUL ISLAM} was born in Chandpur, Bangladesh. He is currently pursuing a B.Sc. degree in computer science and engineering (CSE) with the Khulna University of Engineering \& Technology (KUET), Khulna, Bangladesh. He has worked on some exciting projects and some collaborative works throughout his study. He has worked sincerely at one of the data science projects of OneBlood Blood Centers in a  concerted effort with success. His research interests include machine learning, arm architecture, data science, deep learning, computer vision, RISC architecture, and natural language processing.
\end{IEEEbiography}

\vskip -1\baselineskip plus -1fil
\begin{IEEEbiography}[{\includegraphics[width=1in,height=1.25in,clip,keepaspectratio]{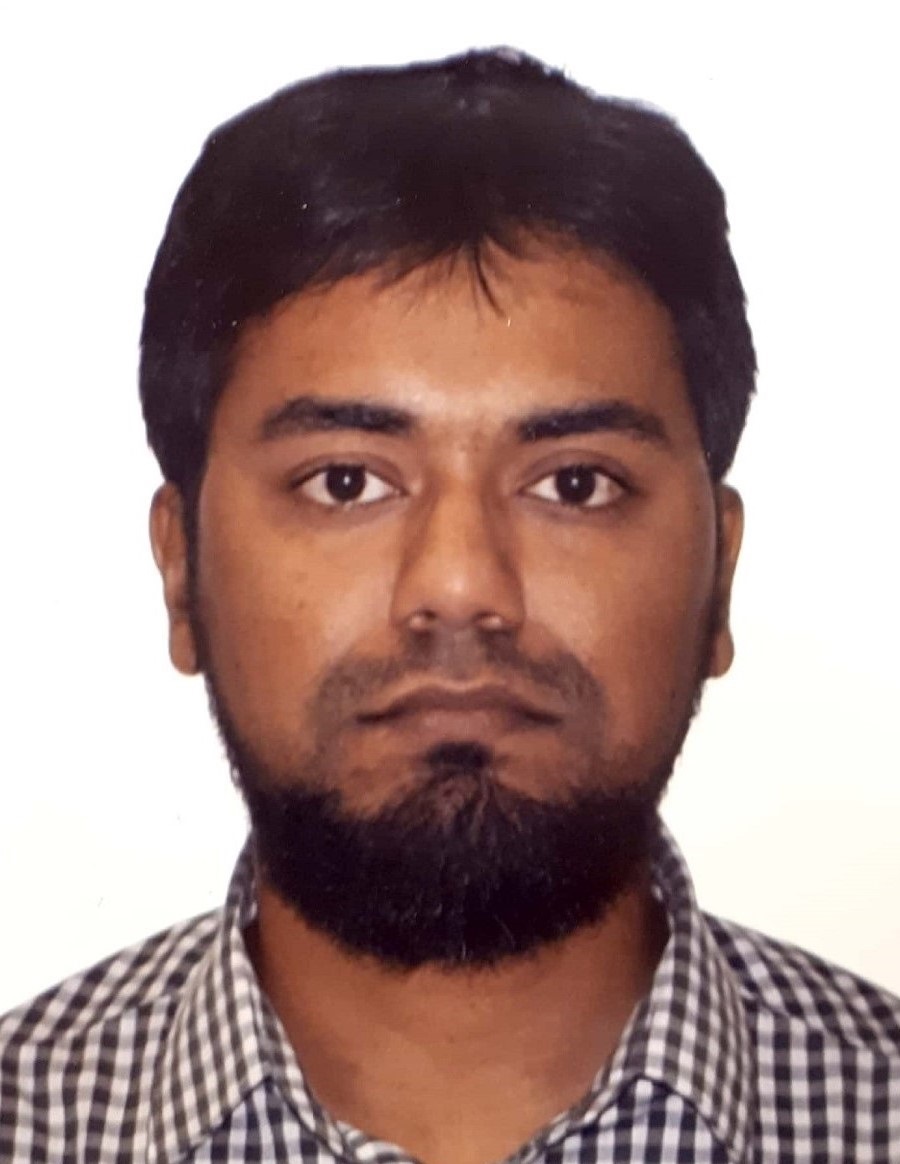}}]{Jakaria Rabbi} received a master's degree in Computing Science from University of Alberta, Edmonton, Canada. He is currently working as an Assistant Professor at the department of Computer Science and Engineering (CSE), Khulna University of Engineering \& Technology (KUET), Khulna, Bangladesh. His research interests include  Machine Learning, Deep Learning, Computer Vision Artificial Intelligence, Data Science and Remote Sensing. He has authored and coauthored several articles in peer-reviewed Remote Sensing journal and IEEE conferences.
\end{IEEEbiography}
\vskip -1\baselineskip plus -1fil
\begin{IEEEbiography}[{\includegraphics[width=1in,height=1.25in,clip,keepaspectratio]{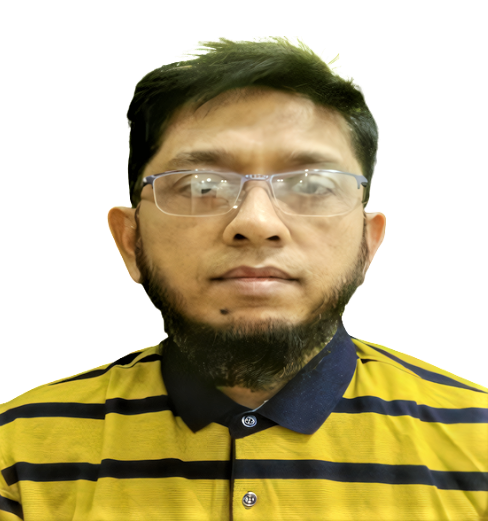}}]{Mehedi Masud} (SM’11) is a Full Professor in the Department of Computer Science at the Taif University, Taif, KSA. Dr. Mehedi Masud received his Ph.D. in Computer Science from the University of Ottawa, Canada. His research interests include cloud computing, distributed algorithms, data security, data interoperability, formal methods, cloud and multimedia for healthcare. He has authored and coauthored around 50 publications including refereed IEEE/ACM/Springer/Elsevier journals, conference papers, books, and book chapters. He has served as a technical program committee member in different international conferences. He is a recipient of a number of awards including, the Research in Excellence Award from Taif University. He is on the Associate Editorial Board of IEEE Access, International Journal of Knowledge Society Research (IJKSR), and editorial board member of Journal of Software. He also served as a guest editor of ComSIS Journal and Journal of Universal Computer Science (JUCS). Dr. Mehedi is a Senior Member of IEEE, a member of ACM.
\end{IEEEbiography}
\vskip -1\baselineskip plus -1fil
\begin{IEEEbiography}[{\includegraphics[width=1.05in,height=1.2in]{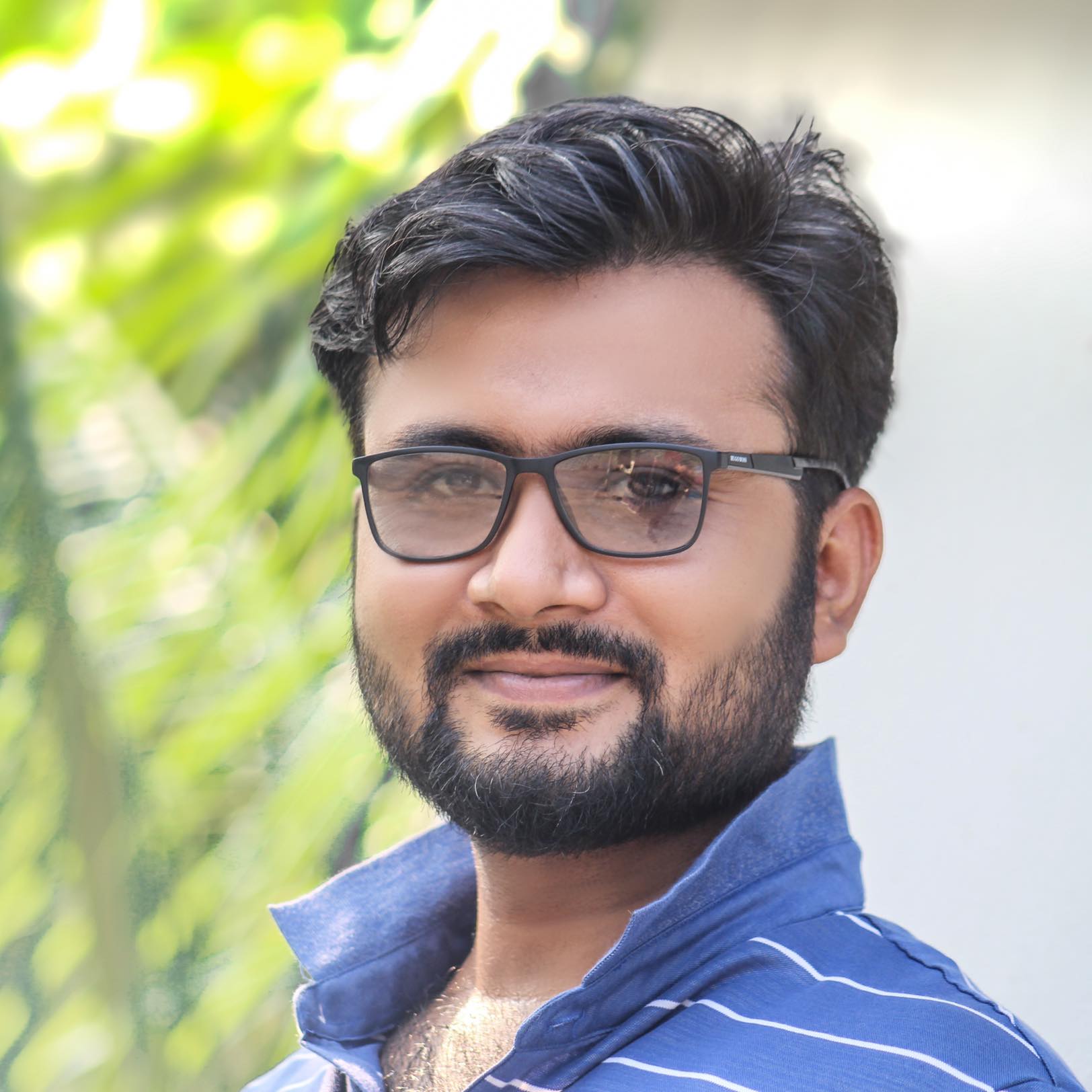}}]{Md. Kamrul Hasan} received B. Sc. and M. Sc. engineering degrees in Electrical and Electronic Engineering (EEE) from Khulna University of Engineering \& Technology (KUET) in 2014 and 2017, respectively. He received another M. Sc. in Medical Imaging and Application (MAIA) from France (University of Burgundy), Italy (the University of Cassino and Southern Lazio), and Spain (University of Girona) as an Erasmus scholar in 2019. Currently, Mr. Hasan is serving as an Assistant Professor at KUET in the EEE department. He analyzed different medical image modalities and machine learning during the MAIA study to build a generic computer-aided diagnosis system. His research interest includes medical image and data analysis, machine learning, deep convolutional neural network, medical image reconstruction, and surgical robotics in minimally invasive surgery. Mr. Hasan is currently a supervisor of several undergraduate students on the classification, segmentation, and registration of medical images with different modalities. He has already published many research articles on medical image and signal processing in different international journals and conferences.
\end{IEEEbiography}
\vskip -1\baselineskip plus -1fil
\begin{IEEEbiography}[{\includegraphics[width=1in,height=1.25in,clip,keepaspectratio]{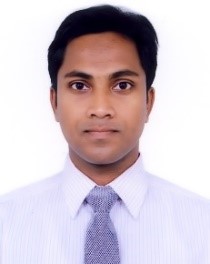}}]{Md. Abdul Awal} has completed his B.Sc. in Electronics and Communication Engineering (ECE) from ECE Discipline, Khulna University in 2009. Later on, he has finished his M.Sc. in Biomedical Engineering from Khulna University of Engineering and Technology in 2011. He completed his PhD in Biomedical Engineering from The University of Queensland, Australia, in 2018. His research interests are Signal Processing, especially Biomedical Signal Processing, Big Data Analysis, Image Processing, Time-Frequency Analysis, Machine Learning Algorithms, Deep Learning, Optimization, and Computational Intelligence Biomedical Engineering. He has more than 35 papers published in internationally accredited journals and conferences. He is currently working as an Associate Professor at ECE Discipline, Khulna University, Khulna, Bangladesh. He is now investigating some projects as principal investigator and co-investigator and supervising several undergraduate and post-graduate students.
\end{IEEEbiography}
\vskip -1\baselineskip plus -1fil
\begin{IEEEbiography}[{\includegraphics[width=1in,height=1.25in,clip,keepaspectratio]{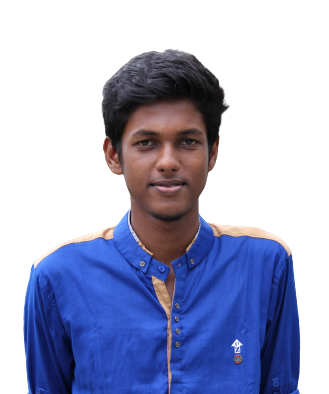}}]{Awal Ahmed Fime}
was born in Jashore,
Bangladesh. He is currently pursuing the B.Sc.
degree in Computer Science and Engineering (CSE)
 with the Khulna University of Engineering \&
Technology (KUET), Khulna, Bangladesh. His
research interests includes Computer vision, Artificial Intelligence, Signal processing,
Machine learning, and Deep learning.
He has already worked on some Web and Mobile Application using ASP.NET, CSS, JavaScript, Android
throughout his study using latest technology.
\end{IEEEbiography}
\vskip -1\baselineskip plus -1fil
\begin{IEEEbiography}[{\includegraphics[width=1in,height=1.25in,clip,keepaspectratio]{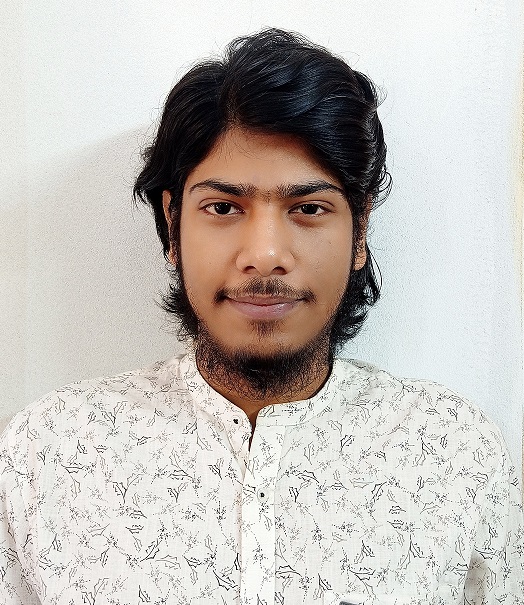}}]{Md. Tahmid Hasan Fuad} was born in Rajshahi, Bangladesh. He is currently pursuing a B.Sc. degree in Computer Science and Engineering (CSE) with the Khulna University of Engineering \& Technology (KUET), Khulna, Bangladesh. His research interests include Image Processing, Computer Vision, Artificial Intelligence, Machine Learning, and Deep Learning. He has already worked on some exciting Android Development and Machine Learning based mini-projects. He has also done some mini-projects using C++, Python, Java etc.
\end{IEEEbiography}
\vskip -1\baselineskip plus -1fil
\begin{IEEEbiography}[{\includegraphics[width=1in,height=1.25in,clip,keepaspectratio]{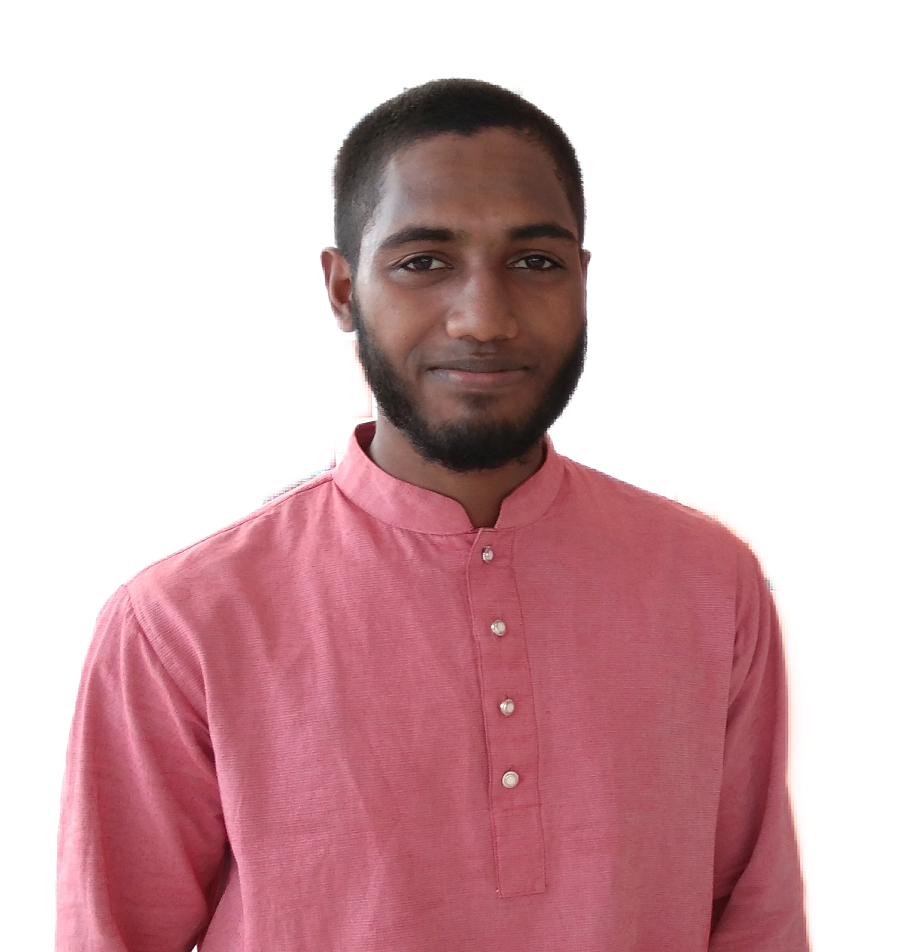}}]{Delowar Sikder}
was born in Patuakhali,
Bangladesh. He is currently pursuing the B.Sc.
degree in Computer Science and Engineering (CSE)
 with the Khulna University of Engineering \&
Technology (KUET), Khulna, Bangladesh. His
research interests include Computer vision, Artificial Intelligence,
Machine learning, and Deep learning.
He has also keen interested to Automated system design, Web and Mobile Application Development. 
He has already worked on some interesting projects
throughout his study using latest technology.
\end{IEEEbiography}
\vskip -1\baselineskip plus -1fil
\begin{IEEEbiography}[{\includegraphics[width=1in,height=1.25in,clip,keepaspectratio]{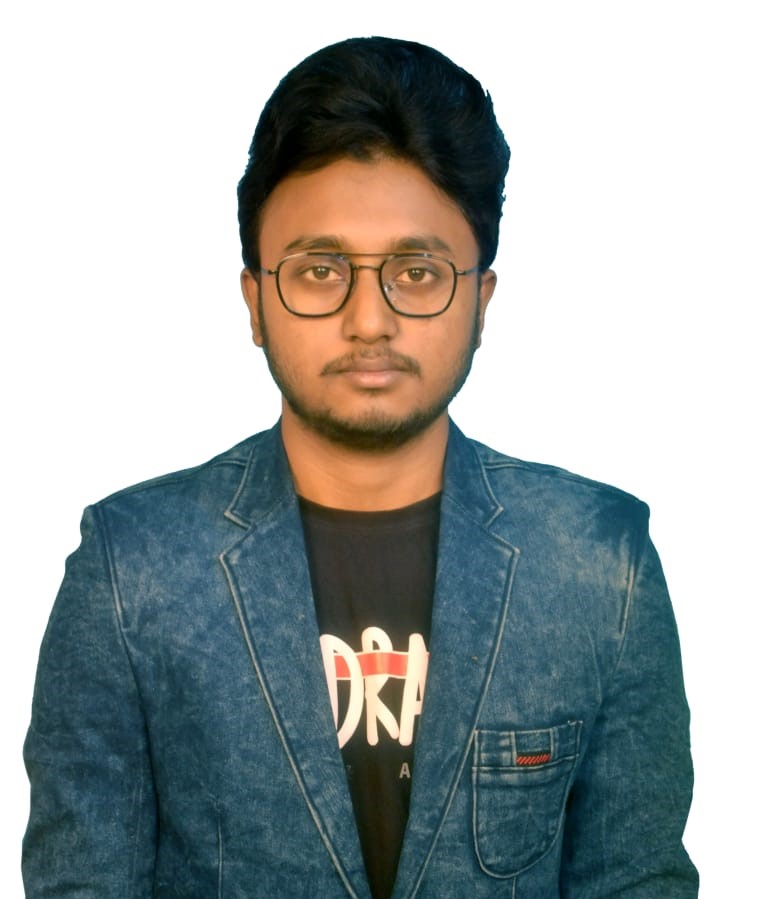}}]{MD.AKIL RAIHAN IFTEE}
was born in  
Joypurhat,  Bangladesh. He is currently pursuing a B.Sc. degree in Computer Science and Engineering (CSE) with the Khulna University of Engineering \& Technology (KUET), Khulna, Bangladesh. His research interests include  Deep  Learning, Data Science, Artificial Intelligence, Machine Learning, and Natural Language Processing. He has already developed some projects using C, C++, Python, Java, HTML, SQL, etc. He is a regular participant in machine learning and data science competitions on an online platform such as Kaggle, Hacker-Earth, etc.
\end{IEEEbiography}

\EOD

\end{document}